\documentclass{article}

\usepackage{microtype}
\usepackage{graphicx}
\usepackage{subfig}
\usepackage{booktabs} 
\usepackage{caption}

\usepackage{hyperref}
\usepackage{caption}
\usepackage{adjustbox}
\usepackage{textcomp}
\usepackage{array} 

\usepackage[dvipsnames]{xcolor}

\definecolor{my_green}{RGB}{51,102,0}
\definecolor{my_yellow}{RGB}{255,165,0}
\definecolor{my_red}{RGB}{204, 0, 0}
\definecolor{my_purple}{HTML}{B54DF5}

\newcommand{\red}[1]{\textcolor{red}{#1}}

\newcommand{\purple}[1]{\textcolor{my_purple}{#1}}
\newcommand{\green}[1]{\textcolor{my_green}{#1}}

\usepackage{tcolorbox}

\definecolor{myboxcolor}{RGB}{245,245,245} 
\definecolor{myframe}{RGB}{0,0,128} 

\newtcolorbox{mybanner}{
  colback=myboxcolor,
  colframe=myframe,
  boxrule=1pt, 
  left=1pt,
  right=1pt,
  top=1pt,
  bottom=1pt,
}

\newtcolorbox{mybody}{
  colback=myboxcolor,
  colframe=myframe,
  boxrule=1pt, 
  left=1pt,
  right=1pt,
  top=1pt,
  bottom=1pt,
}

\usepackage{wrapfig}
\usepackage{xspace}
\usepackage[utf8]{inputenc} 
\usepackage[T1]{fontenc}    

\usepackage{url}         
\usepackage{nicefrac}       
\usepackage{microtype}      
\usepackage{xcolor}         
\usepackage{subfig}
\usepackage{pifont}
\usepackage[normalem]{ulem}

\definecolor{darkgreen}{rgb}{0,0.5,0}
\newcommand{\cmark}{\textcolor{darkgreen}{\ding{51}}}
\newcommand{\xmark}{\textcolor{red}{\ding{55}}}

\newcommand{\name}{\textsc{ConTextual}\xspace}

\usepackage[accepted]{icml2024}

\usepackage{amsmath}
\usepackage{amssymb}
\usepackage{mathtools}
\usepackage{amsthm}

\usepackage[capitalize,noabbrev]{cleveref}

\theoremstyle{plain}

\theoremstyle{definition}

\theoremstyle{remark}

\usepackage[textsize=tiny]{todonotes}

\icmltitlerunning{\name:  Evaluating Context-Sensitive Text-Rich Visual Reasoning in Large Multimodal Models}

\begin{document}

\twocolumn[
\icmltitle{\Large{\name}: Evaluating Context-Sensitive Text-Rich Visual Reasoning in Large Multimodal Models}



\icmlsetsymbol{equal}{*}

\begin{icmlauthorlist}
\icmlauthor{Rohan Wadhawan}{equal,sch}
\icmlauthor{Hritik Bansal}{equal,sch}
\icmlauthor{Kai-Wei Chang}{sch}
\icmlauthor{Nanyun Peng}{sch}
\end{icmlauthorlist}

\icmlaffiliation{sch}{Department of Computer Science, University of California Los Angeles, USA}

\icmlcorrespondingauthor{Rohan Wadhawan}{rwadhawan7@g.ucla.edu}
\icmlcorrespondingauthor{Hritik Bansal}{hbansal@g.ucla.edu}

\icmlkeywords{Machine Learning, ICML}

\vskip 0.3in
]



\printAffiliationsAndNotice{\icmlEqualContribution} 



\begin{abstract}
Many real-world tasks require an agent to reason jointly over text and visual objects, (e.g., navigating in public spaces), which we refer to as context-sensitive text-rich visual reasoning. Specifically, these tasks require an understanding of the context in which the text interacts with visual elements within an image. 
However, there is a lack of existing datasets to benchmark the state-of-the-art multimodal models' capability on context-sensitive text-rich visual reasoning. 
In this paper, we introduce \name, a novel dataset featuring human-crafted instructions that require context-sensitive reasoning for text-rich images. We conduct experiments to assess the performance of 14 foundation models (GPT-4V, Gemini-Pro-Vision, LLaVA-Next) and establish a human performance baseline. Further, we perform human evaluations of the model responses and observe a significant performance gap of 30.8\% between  GPT-4V (the current best-performing Large Multimodal Model) and human performance. Our fine-grained analysis reveals that GPT-4V encounters difficulties interpreting time-related data and infographics. However, it demonstrates proficiency in comprehending abstract visual contexts such as memes and quotes. Finally, our qualitative analysis uncovers various factors contributing to poor performance including lack of precise visual perception and hallucinations. Our dataset, code, and leaderboard can be found on the project page \href{https://con-textual.github.io/}{https://con-textual.github.io/}. 
\end{abstract}


\begin{figure}[!h]
\centering
    \subfloat[\small{Comparing an instance of \textbf{\name} to existing datasets (e.g., ESTVQA). \textbf{\name} requires contextualized understanding of the interactions between the textual and visual elements in the image while ESTVQA can be solved solely through text-based reasoning combined with accurate OCR detection.}]{\includegraphics[width=0.95\linewidth]{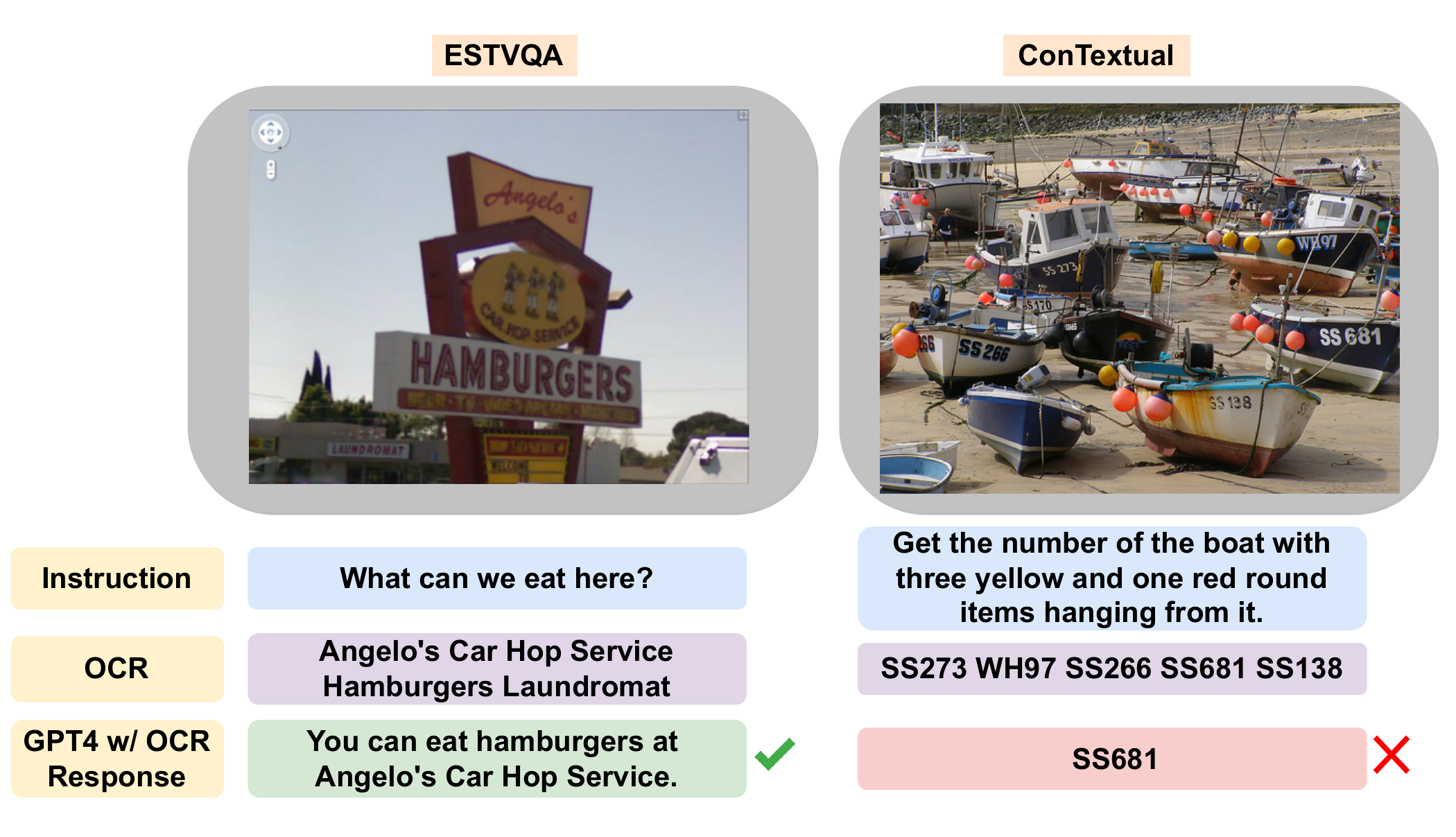}
    \label{fig:compare_ex}}\par
    \captionsetup{width=\linewidth}
    \subfloat[\small{Performance of GPT-4V, Gemini-Pro-Vision, ShareGPT-4V-7B, and humans on the \textbf{\name} dataset, with Human Evaluation (left sub-graph) and GPT4 Evaluation (right sub-graph).}]{\includegraphics[scale=0.25]{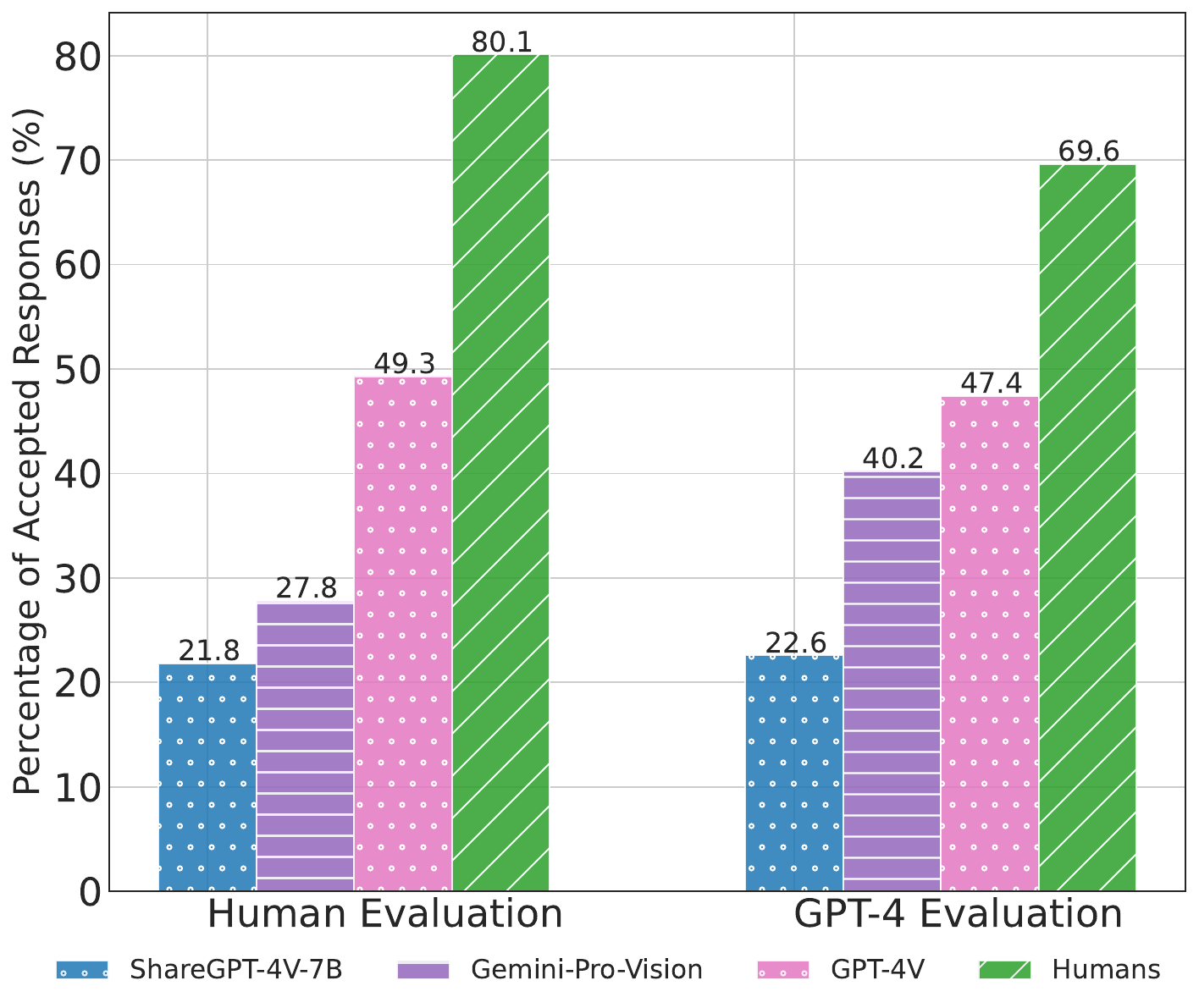}
    \label{fig:contextual_performance}}
\caption{\small{Comparisons between our dataset \textbf{\name} and prior work ESTVQA, along with benchmark performances of large multimodal models on \textbf{\name}.} }
\label{fig:teaser_fig}
\end{figure}

\begin{figure*}[!h]
    \centering
    \resizebox{0.9\textwidth}{!}{{\includegraphics[scale=0.33]{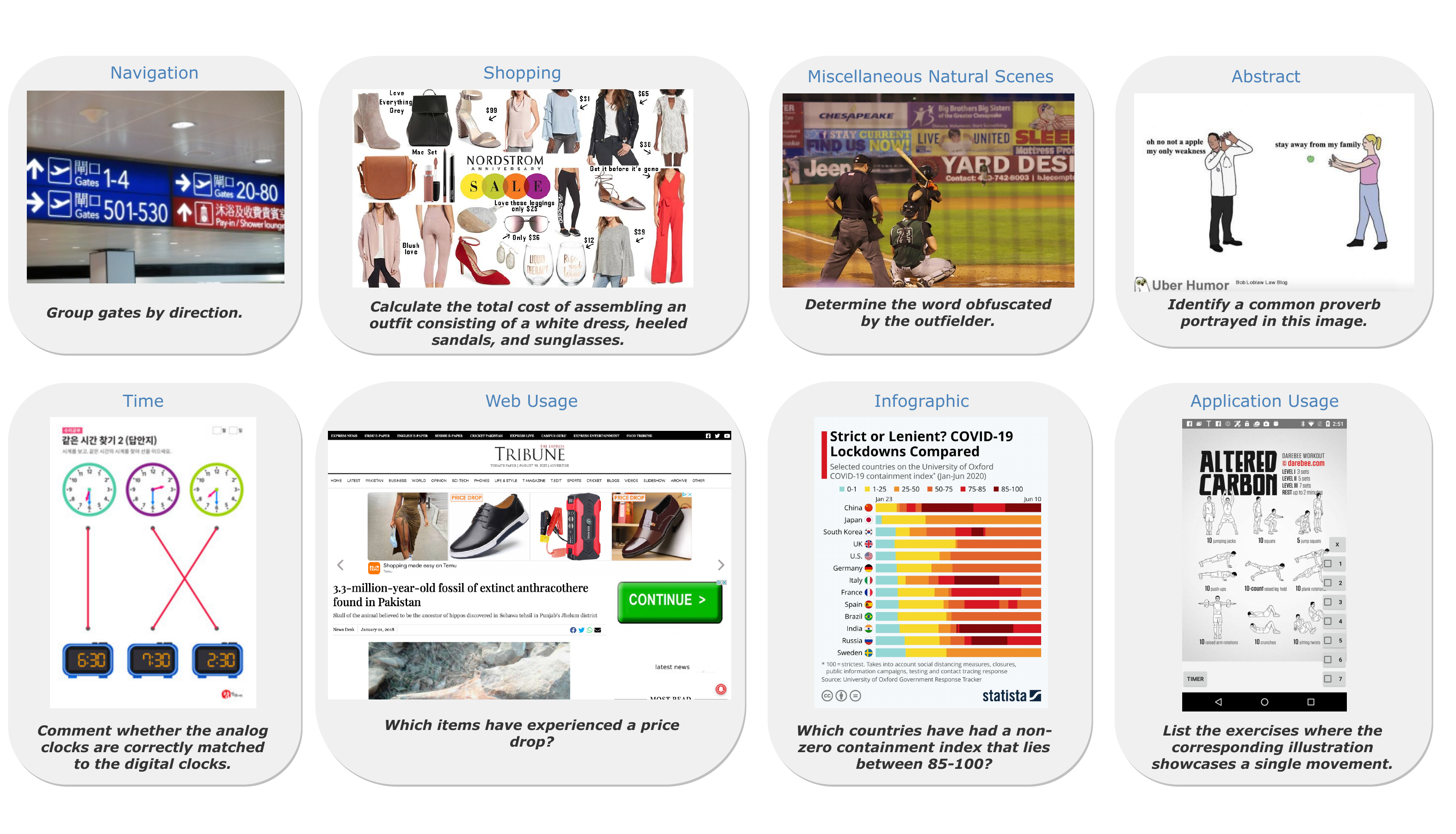}}}
    \caption{A sample (image, instruction) from each of the 8 visual scenarios in \textbf{\name} dataset. The categories organized in a left-to-right, top-to-bottom reading order include Navigation, Shopping, Miscellaneous Natural Scenes, Abstract, Time, Web Usage, Infographic, and Application Usage.}
    \label{fig:visual_examples_ins}
\end{figure*}

\section{Introduction}
The recent development of large multimodal models (LMMs) has resulted in models capable of responding to human instructions, posed as questions or imperative tasks, over images \cite{llava,chen2023sharegpt4v,gpt4v,team2023gemini,instructblip,qwen,idefics,mplugowl}. Many real-world images contain texts within them which provides cues for comprehensively understanding them. The ability to reason about the interactions between the text and visual context in the images powers many real-world applications. For example, interpreting text-rich scenes (e.g., navigating maps in public spaces) for assisting the visually impaired, and creative understanding of abstract text-rich images (e.g., memes).

Previous datasets such as TextVQA \cite{textvqa}, STVQA \cite{textvqa}, ESTVQA \cite{estvqa} have been proposed to assess the visual reasoning ability of multi-modal models over text-rich images. However, these datasets focused on accessing the OCR capability of the models to \textit{read} the text in the image, and they usually do not require the model to capture the visual context in the image to answer the question. For example, in Figure \ref{fig:compare_ex}, we highlight an example from the ESTVQA dataset. Here, we show that a high accuracy OCR of the images (e.g, `Angelo's Car Hop Service Hamburgers Laundromat') has sufficient signal to answer the question (e.g., `What can we eat here?'). Though accessing the OCR capability is important, these examples do not test the unique potential of the LMMs to jointly reason over the embedded text and visual context in the image. 

\begin{table*}[t!]
\centering
\caption{\small{Comparison with related works for evaluating large multimodal models. We abbreviate Context-sensitive as Consens., Generation as Gen. We compare our work with LLaVA \cite{llava}, VisIT \cite{visit-bench}, \cite{textvqa}, STVQA \cite{stvqa}, DUDE \cite{van2023document}, InfoVQA \cite{infovqa}, and SEED Bench \cite{li2023seed}.}}
\label{tab:related_work}
\setlength{\tabcolsep}{0.8mm}
\resizebox{0.7\linewidth}{!}{
\begin{tabular}{lcccccccc}
\hline
 & \textbf{Ours} & LLaVA & VisIT & TextVQA & STVQA   & DUDE    & InfoVQA    & SEED \\
\hline
Consens. Text-Rich Visual Reasoning & \cmark          & \xmark           & \xmark           & \xmark       & \xmark       & \xmark       & \xmark                    & \xmark          \\
Text in Images                        & \cmark          & \xmark           & \xmark           & \cmark       & \cmark       & \cmark       & \cmark                  & \xmark          \\\hline

Number of LLM/LMM Models                            & 13         & 3           & 10          & -       & -       & 9       & -          & 15         \\
Number of Images                                & 506        & 24          & 574         & 28.4K   & 23K     & 5K      & 5.4K         & 19K        \\\hline
Diverse Image Sources                        & \cmark    & \xmark      & \cmark     & \xmark  & \cmark & \cmark & \xmark     & \xmark     \\\hline
Question Instructions                                         & \cmark          & \cmark           & \cmark           & \cmark       & \cmark       & \cmark       & \cmark                & \cmark          \\
Imperative Instructions                                  & \cmark          & \xmark           & \cmark           & \xmark       & \xmark       & \xmark       & \xmark                  & \xmark          \\\hline
Instruction Gen. by Humans                         & \cmark          & \cmark           & \cmark           & \cmark       & \cmark       & \cmark       & \cmark                  & \xmark          \\
Reference Response Gen. by Humans                         & \cmark          & \cmark           & \xmark           & \cmark       & \cmark       & \cmark       & \cmark                & \xmark          \\\hline
Human Evaluation                                   & \cmark          & \xmark           & \cmark           & \cmark       & \xmark       & \cmark       & \cmark                  & \xmark          \\
Automatic Evaluation                                    & \cmark          & \cmark           & \cmark           & \cmark       & \cmark       & \cmark       & \cmark                  & \cmark          \\
Human-Auto Eval. Correlation         & \cmark          & \xmark           & \cmark           & \xmark       & \xmark       & \xmark       & \xmark                    & \xmark          \\\hline
Human performance                            & \cmark          & \xmark           & \xmark           & \cmark       & \xmark       & \cmark       & \cmark               & \xmark          \\
Absolute Score to Models                      & \cmark          & \cmark           & \cmark           & \cmark       & \cmark       & \cmark       & \cmark                 & \cmark          \\
Fine-grained Analysis                        & \cmark          & \xmark           & \cmark           & \xmark       & \xmark       & \cmark       & \xmark                    & \cmark      \\\hline   
\end{tabular}
}
\end{table*}

To evaluate multimodal models' capability of jointly reasoning over embedded text and visual context in text-rich images, we propose \textbf{\name{}}, a \textbf{Con}text-sensitive \textbf{Text}-rich vis\textbf{ual} reasoning dataset consisting of \textbf{506} challenging instructions for LMMs evaluation. \name covers \textbf{eight} real-world scenarios with text-rich images: \textit{time reading}, \textit{shopping}, \textit{navigation}, \textit{abstract scenes}, \textit{mobile application}, \textit{webpages}, \textit{infographics}, and \textit{miscellaneous natural scenes} (Figure \ref{fig:visual_examples_ins}). The diverse visual nature of these categories enables us to conduct a detailed, nuanced evaluation of the model's capabilities. 

Each instance of \name contains a human-written instruction (question or imperative task) and a human-written ground-truth response (\S\ref{sec:dataset_collection}), with the constraint that to respond to an instruction accurately, a model must require context-sensitive joint reasoning over the textual and visual cues in the image. Figure \ref{fig:compare_ex} shows an example from our dataset. The instruction (`Get the number of the boat with three yellow and one red round items hanging from it.') cannot be answered even by perfectly capturing the OCR of the text content within the image (e.g., `SS273 WH97 SS266 SS681 SS138'). We summarize our work compared to the related works in Table \ref{tab:related_work}.

We conduct extensive experiments using \name{} to assess the reasoning abilities of 14 foundation models over context-sensitive text-rich images (\S \ref{sec:models}). This includes three augmented LLMs setups (e.g., GPT-4 \cite{gpt4} prompted with combinations of image OCR, image layouts, and image captions), two proprietary LMMs (e.g., GPT-4V\cite{gpt4v}, Gemini-Pro-Vision \cite{team2023gemini}), and nine open LMMs (e.g., LLaVA-Next \cite{llavanext}, LLaVA-1.5 \cite{llavav15}, ShareGPT-4V\cite{chen2023sharegpt4v}, Idefics \cite{idefics}). In addition, we perform few-shot experiments for a selected set of models (e.g., Gemini-Pro-Vision, Idefics) to analyze the effect of in-context examples on the model's performance. Further, we establish a human baseline by asking human annotators to write responses to the dataset instructions. Finally, we perform human and automatic evaluations to assess the correctness of the predicted responses with respect to the ground-truth responses in the dataset (\S \ref{sec:evaluation}, \S \ref{sec:results}). 

Through human evaluations on randomly selected 280 instances, we find that GPT-4V(ision) is the best performing LMM on the \name{} dataset where it achieves \textbf{$49.3\%$} acceptance rating to its generated responses (Figure \ref{fig:contextual_performance}). Despite this, the performance lags way behind the human baseline of \textbf{$80.1\%$} which indicates a large gap in the capabilities of the GPT-4V. In addition, we find that the best performing open-model, ShareGPT-4V-7B, only achieves $21.8\%$ rating which indicates that the capabilities of open models are way behind the proprietary models on context-sensitive text-rich visual reasoning (\S \ref{sec:results}). Our results highlight that the \name{} is a challenging dataset for modern LMMs while humans excel on it. 

Since human evaluations are hard to scale and expensive, we also perform automatic evaluation (e.g., GPT-4, GPT-4V, BLEURT \cite{bleurt}) on the complete dataset for all the models (\S \ref{sec:correlation}). Further, we perform fine-grained experiments to assess the model's performance across visual contexts (\S \ref{sec:fine_grained}). We observe that GPT-4V, the best-performing LMM, struggles with time reading and infographic visual contexts, except for abstract contexts like memes and quotes, where it outperforms humans. On the other hand, open models lag behind proprietary ones across most visual tasks, showing moderate proficiency only in abstract and natural scenes, owing to the need for more diversity of visual context in training data. However, we observe significant improvement in model performance with enhancement in image encoders, as seen with LLaVA-Next over LLaVA-v1.5. Lastly, we conduct a qualitative analysis (\S \ref{sec:qualitative_examples}) of model responses for the different visual contexts in \name, revealing that both proprietary and open models exhibit a limited capacity for fine-grained visual perception, with open models performing worse.

\section{The \name Dataset}
\label{sec:dataset_collection}

\subsection{Collection Guidelines}

We note that there is a notable gap in the existing benchmarks for text-rich images, which primarily evaluate text reading capabilities of LMMs. Our dataset bridges this gap and offers an evaluation framework to test the joint reasoning capabilities of the LMMs over the embedded text and the visual features in the image (Figure \ref{fig:teaser_fig} (b)). Our dataset encompasses a variety of tasks across diverse natural and digital text-rich visual scenarios, enabling robust testing.

Broadly, our benchmark follows these key collection guidelines: (a) Each sample consists of an \(<image,\ instruction,\ response>\) triplet, such that the instructions require the models to perform context-sensitive reasoning over the text and visual elements in the image. Specifically, we would avoid creating instructions that could be answered by text-based reasoning (e.g., using LLM) over the detected OCR. (b) We aim to cover diverse instructions, including questions and imperative tasks. This ensures that the resulting dataset demonstrates a rich variety of instructions. (c) We aim to create instructions of varied complexity. Specifically, they can make extractive instructions that involve extraction of a specific textual or visual elements (e.g., `Name the drink with banana flavor.'). In addition, they can make instructions that involve reasoning over the embedded information (e.g., `count the number of words in the rendered in the blue color.').

\begin{table}[h]
    \centering
    \resizebox{0.6\linewidth}{!}{%
    \begin{tabular}{l|c}
    \hline
    \textbf{Statistic} & \textbf{Number} \\
    \hline
    Total number of samples & 506 \\
    Root verbs in instructions & 79\\
    \hline
    \# Visual Contexts & 8 \\
    - Time & 50\\
    - Shopping & 50\\
    - Navigation & 50\\
    - Abstract & 50\\
    - Application Usage & 50\\
    - Web Usage & 50\\
    - Infographic & 50\\
    - Miscellaneous Natural Scenes & 156\\
    \hline
    Average Instruction Length & 65\\
    Average Response Length & 117\\
    \hline
    \end{tabular}
    }
    \caption{Key Statistics of \name.}
    \label{tab:contextual_stats}
\end{table}

\begin{figure}
    \centering
    \includegraphics[scale=0.3]{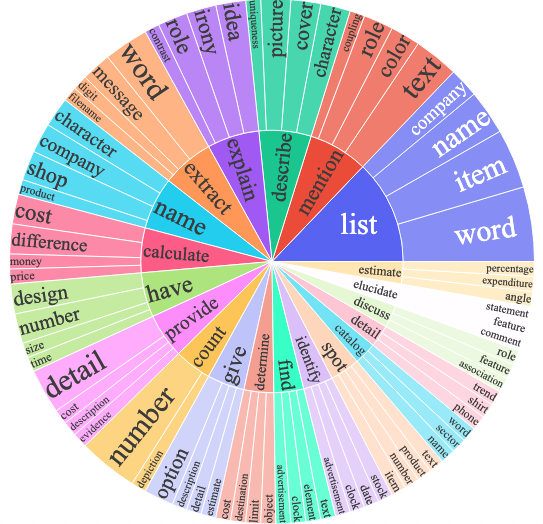}
    \caption{Top 20 Most frequently occurring verbs (inner circle) and their top 4 direct nouns (outer circle) in the instructions.}
    \label{fig:root_verb_noun}
\end{figure}

In this work, we establish a taxonomy by categorizing a dataset into eight distinct visual scenarios, encompassing real-world and digital human interactions. These scenarios include shopping, navigation, time, mobile and web usage, infographics, abstract scenes, and miscellaneous natural scenes. More details are available in Appendix \S \ref{app:data_details}.

\subsection{Data Sources}

\textbf{\name} comprises images sourced from six different sources. Firstly, we obtain images for the \textit{Time}, \textit{Shopping}, \textit{Navigation} \& \textit{Abstract} categories from the LAION-5B \cite{laion}. Specifically, we use keyword search using CLIP-retrieval UI \cite{clip-retrieval}. A keyword consists of \colorbox{lightgray!30}{category-specific word(s) + "text"} (e.g., clothes text for shopping, airports text for navigation). Some category-specific words we used are: shopping (e.g., grocery, furniture, gadgets, cosmetics, services, clothes), navigation (e.g., street signs, cars, buses, trains, metro, airport, stations, highways, roads), time (e.g., clocks, multiple clocks, digital clocks, calendars), and Abstract (e.g., memes, quotes, comic strips, science jokes, math jokes).

Secondly, we source images for the \textit{Application Usage} category from the Rico Dataset \cite{rico}, which includes 66,000 distinct UI screens originating from 9,300 Android apps. Thirdly, we scrape the website links made available by the Open WebText Initiative \cite{open-web-text} and collect screenshots for the \textit{Web Usage} category. Lastly, we acquire images from the test sets of existing VQA datasets, and proceed to annotate them with novel instruction-response pairs. Originally, these datasets consists question-and-answer pairs that primarily focus on text recognition capabilities. However, they offer an opportunity to formulate context-sensitive instructions for the images. Specifically, we reannotate these image instances, categorizing them into two groups: \textit{Infographic}, sourced from the InfographicVQA \cite{infovqa} dataset, and \textit{Miscellaneous Natural Scenes}, sourced from the STVQA \cite{stvqa} and ESTVQA \cite{estvqa} datasets.

\subsection{Data Annotation}


\textbf{Stage 1:} In this stage, we shortlist images for instruction-response pair annotation. The images that are categorized under \textit{Time}, \textit{Shopping}, \textit{Navigation}, and \textit{Abstract} undergo manual filtering to guarantee their suitability for annotation. However, for \textit{Application Usage}, \textit{Web Usage}, \textit{Infographic}, and \textit{Miscellaneous Natural Scenes}, we perform heuristic-based filtering. Specifically, we employ a PaddleOCR \cite{paddleocr} to detect the text in the image. Subsequently, we select the top 500 images with the highest number of words, a subset of which get annotated in our dataset.

\textbf{Stage 2:} We divide the authors into two groups, namely Group 1 and Group 2, each responsible for annotating four specific categories. The authors strictly adhered to the provided annotation guidelines throughout the annotation. \footnote{We observe that MTurk workers found this task time-consuming, leading to annotations that would be hard to accomplish within a limited budget.}

\textbf{Stage 3:} In this final stage, we perform a verification process for each sample annotated in Stage 2. We asked MTurk workers (mutually exclusive from the ones used for human performance baseline and human evaluation) to verify the correctness of each sample \(<image,\ instruction,\ response>\) and found that 96\% of the samples were annotated correctly. Filtering out the incorrect samples, we tasked each author group to review the samples created by the other group. This ensured adherence to guidelines and filtered out low-quality samples. Finally, we end up with a dataset of \textbf{506} instances.

We provide statistics for the \textbf{\name} benchmark, as shown in Table \ref{tab:contextual_stats}. We visualize each instruction based on its root verb and the direct noun, as shown in Figure \ref{fig:root_verb_noun}. We also annotate each sample to determine whether it requires information extraction, and mathematical reasoning (Appendix \S \ref{app:task}). We provide details on data release in \S \ref{sec:data_release}.

\begin{table*}[h]
\centering
\caption{\small{Comparison of the performance of various foundation models (augmented LLM and LMMs) and humans on a subset of \name dataset (280 samples). We report the response acceptance rating using human evaluation, automatic GPT-4 and GPT-4V based evaluation. In addition, we report standard text generation quality assessment metrics including BLEURT, Rouge-L, and BERTScore. The best performance in a column is highlighted in \textbf{BLACK} while the second best performance is highlighted in \underline{UNDERLINE}.}}
\label{tab:main_results}
\small
\begin{tabular}{lcccccc}
\hline
                                         & \textbf{Humans} & \textbf{GPT-4} & \textbf{GPT-4V}& \textbf{BLEURT} & \textbf{Rouge-L} & \textbf{BERTScore} \\\hline
GPT-4 w/ Layout-aware OCR + Caption  & 17.2  & 22.2  & 17.6  & 41.3   & 22.5    & 53.9      \\
GPT-4V  \cite{gpt4v}          & \underline{49.3}  & \underline{47.4}   & \underline{45.0} & \underline{45.3}   & 17.3    & 52.5      \\
Gemini-Pro-Vision   \cite{team2023gemini}             & 27.8  & 40.2 & 37.1  & 42.5   & \underline{30.1}    & \underline{58.4}\\
LLaVA-1.5-13B   \cite{llavav15}                & 17.2  & 20.6  & 17.5 & 43.6   & 21.7    & 54.8      \\
ShareGPT-4V-7B   \cite{chen2023sharegpt4v} & 21.8  & 22.6  &20.6  & 44.5   & 23.3    & 55.8      \\
\hline
Humans                                     & \textbf{80.1}  & \textbf{69.6}  & \textbf{68.6} & \textbf{47.4}   & \textbf{33.6}    & \textbf{59.8}     \\\hline
\end{tabular}%
\end{table*}

\begin{table}[h]
\centering
\small
\setlength{\tabcolsep}{1mm}
\resizebox{0.9\linewidth}{!}{
\begin{tabular}{lccccc}
\hline
 & \textbf{GPT-4} & \textbf{GPT-4V} & \textbf{BLEURT} & \textbf{RougeL} & \textbf{BERTScore} \\\hline
ROC-AUC & \textbf{85.9} & 83.9 & 72.9   & 67.6   & 66.8 \\
Spearman  & \textbf{0.71} & 0.68 & 0.38   &  0.29  & 0.28\\
\hline
\end{tabular}
}
\caption{\small{Comparison of the human and automatic evaluation metric using ROC-AUC and spearman correlation on a subset of ConTextual dataset (280 samples, similar to Table 3). We find that the GPT-4 and GPT-4V based evaluation correlate the most with humans.}}
\label{tab:correlation}
\end{table}

\section{Experiments}
\label{sec:experiments}


\subsection{Setup}
\label{sec:models}

\paragraph{Augmented LLMs.} Since our dataset is focused on text-rich visual reasoning, it is imperative to understand the extent to which a strong LLM GPT-4 can perform on \name{} dataset with the OCR information and image captions \cite{lu2023chameleon,wu2023visual,suris2023vipergpt,gupta2023visual}. To this end, we study this augmented setup under three settings: GPT-4 prompted with (a) vanilla OCR, (b) layout-aware OCR, and (c) combining layout-aware OCR with image captions. We leverage the PP-OCRv4 model of PaddleOCR library \cite{paddleocr} for extracting OCR from the images, LATIN prompt \cite{latin} inspired OCR text arrangement implementation to maintain layout-awareness in the OCR, and ShareGPT-4V-7B for the dense image captions (App. \S \ref{app:aug_llm_prompt}).

\paragraph{LMMs.} We evaluate GPT-4V \cite{gpt4v} and Gemini-Pro-Vision \cite{team2023gemini} that are representative proprietary LMMs that have achieved state-of-the-art on other visual reasoning benchmarks \cite{goyal2017making}. In addition, we evaluate a wide range of open LMMs including LLaVA-Next-34B \cite{llavanext}, LLaVA-1.5-13B \cite{llavav15}, ShareGPT-4V-7B \cite{chen2023sharegpt4v}, mPLUG-Owl-v2-7B \cite{mplugowl,ye2023mplug}, Qwen-VL-7B \cite{qwen}, InstructBLIP-Vicuna-7B \cite{instructblip}, and Idefics-9B \cite{idefics}. We include LLaVAR \cite{llavar} and BLIVA-Vicuna-7B \cite{bliva} as they were introduced for text-rich visual reasoning. 

\begin{table*}[h]
\caption{\small{Fine-grained comparison in the zero-shot performance of the foundation models and humans on the \name dataset using GPT-4 evaluation. We abbreviate the average response acceptance rating as Avg., Navigation as Nav., Abstract as Abs., Application usage as App., Infographics as Info., Miscellanous natural scenes as NS. We find that the GPT-4V outperforms all the model baselines on most of the categories while Gemini-Pro-Vision is the best on Web usage and natural scenes. The best performance in a column is highlighted in \textbf{BLACK} while the second best performance is highlighted by \underline{UNDERLINE}.}}
\label{tab:fine_grained_GPT_4}
\centering
\small
\resizebox{0.95\linewidth}{!}{
\begin{tabular}{l|c|cccccccc}
\hline
\textsc{Models}& \textsc{Avg.} & \textsc{Time}  & \textsc{Shop.} & \textsc{Nav.}  & \textsc{Abs.}  & \textsc{App.}  & \textsc{Web.}  & \textsc{Info.}   & \textsc{Misc. NS.} \\\hline
\textit{Augmented Large Language Models} \\
\hline
GPT-4 w/ OCR& 15.9& 4.0  & 10.0  & 14.0    & 30.6  & 8.0& 16.0 & 28.6&  16.9   \\
GPT-4 w/ Layout-aware OCR& 18.2& 8.0  & 20.0  & 18.0    & 34.7  & 10.0    & 16.0 & 16.0& 20.7   \\
GPT-4 w/ Layout-aware OCR + Caption & 22.2& 6.0  & 16.0  & 24.0    & 57.1  & 14.0    & 18.0 & 8.0  &  27.3   \\
\hline
\textit{Large Multimodal Models} \\
\hline
GPT-4V \cite{gpt4v}  & \underline{47.4}& \underline{18.0} & \underline{54.0}  & \underline{48.0}    & \textbf{100.0} & \underline{48.0}    & 42.0 & \underline{28.0}   &  48.0   \\
Gemini-Pro-Vision  \cite{team2023gemini}   & 40.2& 16.0 & 32.7  & 28.6    & 65.3  & 44.9    & \underline{43.8} & 20.0& \underline{52.8}   \\
LLaVA-Next-34B  \cite{llavanext} & 36.8 & 10.0  & 36.0  & 30.6    & 66.0  & 36.0   & 28.0 & 12.0 & 51.3   \\
ShareGPT-4V-7B  \cite{chen2023sharegpt4v} & 22.6& 0.0  & 16.0  & 20.0    & 28.6  & 20.0    & 20.0 & 14.0 & 37.7   \\
Qwen-VL-7B  \cite{qwen}  & 21.8& 4.0  & 20.0  & 24.0    & 53.1  & 6.0& 18.0 & 14.0 & 27.3   \\
LLaVA-1.5B-13B   \cite{llavav15} & 20.8& 4.0  & 10.0  & 18.0    & 44.9  & 16.0    & 26.0 & 4.0  & 29.7   \\
mPLUG-Owl-v2-7B  \cite{mplugowl}    & 18.6& 4.0  & 8.0   & 24.0    & 32.7  & 20.0    & 10.0 & 12.0 & 26.0   \\
LLaVAR-13B  \cite{llavar}  & 14.9& 10.0 & 16.0  & 6.0& 44.9  & 8.0& 10.0 & 6.0& 16.7   \\
BLIVA-Vicuna-7B  \cite{bliva}  & 10.3& 2.0  & 4.0   & 14.0    & 24.5  & 4.0& 8.0  & 4.0 & 14.7   \\
InstructBLIP-Vicuna-7B \cite{instructblip} & 9.7 & 2.0  & 4.0   & 16.0    & 20.0  & 6.0& 12.0 & 2.1 & 12.0   \\
Idefics-9B \cite{idefics}    & 7.7 & 4.0  & 2.0   & 12.0    & 12.0  & 0.0& 6.0  & 2.0 & 13.3   \\

\hline
Humans                 & \textbf{69.6}     & \textbf{64.0}      & \textbf{64.0}          & \textbf{73.5}            & \underline{75.5}          & \textbf{64.0}            & \textbf{58.0}         & \textbf{72.0}                & \textbf{78.0}  \\\hline 
\end{tabular}%
}
\end{table*}
\paragraph{Humans.} We also benchmark the performance of humans on our dataset using Amazon Mechanical Turk. The selected annotators that pass an qualification test were asked to write accurate responses for all the instruction-image from the dataset. We provide the screenshot of our annotation interface in Appendix \S \ref{app:human_responses_ss}. We spent $\$180$ on collecting human predictions on our dataset.

\subsection{Evaluation}
\label{sec:evaluation}


\subsubsection{Human Evaluation}
\label{sec:human_eval}

To perform a faithful evaluation of the predicted responses, we ask human annotators sourced from Amazon Mechanical Turk to rate the predicted response quality given the image, instruction, and reference response from our dataset. First, we sample 280 instances from the dataset randomly from the \name{} dataset. Second, we collect the model responses for these instances from augmented LLM (GPT-4 with layout-aware OCR and image captions), GPT-4V, Gemini-Pro-Vision, LLaVA-1.5-13B, ShareGPT-4V-7B, and humans. In total, we have 1680 predicted responses from models and humans. Third, we show each model response, without revealing the model identity, to three human annotators independently. Specifically, the human annotators are asked to decide the predicted response is acceptable given the reference response, instruction and image from the dataset. Finally, we report the acceptance rating (0-100 in percentage) of the responses using the majority vote among the three annotator as the final decision. We provide the screenshot of our annotation interface in Appendix \ref{app:human_eval_ss}. We spent $\$1000$ in acquiring human judgments. 

\subsubsection{Automatic Evaluation}
\label{sec:auto_eval}

While human evaluation acts as a gold standard, it is hard to scale since it is expensive and time-taking. Since our dataset uniquely provides reference response for each instruction, we utilize test a wide range of reference-guided automatic evaluation methods. Specifically, these include (a) prompting an LLM GPT-4 with the instruction, reference response and predicted response, (b) prompting an LMM GPT-4V with the image, instruction, reference response and predicted response, (c) and other text generation methods like BLEURT \cite{bleurt}, Rouge-L \cite{rougel} and BERTScore \cite{zhang2019bertscore} that assess the similarity between the reference response and predicted response. Specifically, GPT-4 and GPT-4V are prompted to provide their judgement on the predicted response, same as human evaluation. We present the prompt for GPT-4 based evaluation in Appendix \S \ref{app:gpt_4_eval_prompt}. The other text generation methods provide a continuous score 0-1 which is scaled to 0-100. 

Through our automatic evaluation methods, we evaluate all the model responses on the entire dataset. Subsequently, we conduct a correlation analysis between human and automated methods, utilizing the same 1,680 responses from the human evaluation, to assess the efficacy of the automated approaches (\S \ref{sec:correlation}). Finally, we utilize the GPT-4 automatic evaluation, that achieves the highest correlation with human judgments, for large-scale evaluation of all the models on the complete dataset (\S \ref{sec:fine_grained}).

\subsection{Results}
\label{sec:results}

We compare the performance of augmented LLM, LMMs, and humans on \name{} using human and automatic evaluation in Table \ref{tab:main_results}. Through our human evaluations, we find that the humans perform the best on the dataset with the response acceptance rating of $80.1\%$. In addition, we observe that the GPT-4V achieves the highest acceptance rating of $49.3\%$ in comparison with all the other models. However, this rating is quite far from the human performance which indicates that our task is quite challenging for the state-of-the-art LMMs while humans are good at it. We find that the GPT-4V outperforms Gemini-Pro-Vision by $22\%$ highlighting a large gap in the models text-rich visual reasoning capabilities. Further, we find that augmented LLM approach achieves a very low rating of $17.2\%$ which indicates that the dataset instances cannot be solved without precise visual perception. Interestingly, we observe that the open-models such as LLaVA-1.5-13B and ShareGPT-4V-7B achieve poor acceptance ratings through human evaluations which indicates the presence of a large gap in their capabilities from proprietary models. This might be attributed to the differences in the model capacity, along with the scale and quality of the pretraining data.

As human evaluation is not scalable, we perform automatic evaluation of the model responses on the entire dataset. In Table \ref{tab:main_results}, we find that the ratings of the human responses outperform those from GPT-4V by $22.2\%$ and $23.6\%$ using GPT-4 and GPT-4V evaluation. Like human evaluation, automatic evaluation with GPT-4 and GPT-4V highlights that the human performance on the \name{} dataset is way higher than the best-performing LMM. Interestingly, the gap between the performance of GPT-4V and Gemini-Pro-Vision is $7.2\%$ as per GPT4 evaluation. We still observe a large gap in the performance of the proprietary models and open LMMs. We perform fine-grained evaluation to understand the gaps in model capabilities along the various quality dimensions in \S \ref{sec:fine_grained}. 

Furthermore, we find that the BLEURT scores for humans are the highest, while GPT-4V achieves the highest score among the LMMs. Interestingly, the open models (LLaVA-1.5, ShareGPT-4V) achieve a higher BLEURT score than Gemini-Pro-Vision. We observe similar counter-intuitive trends in our Rouge-L and BERTScore based automatic evaluations. For instance, Rouge-L and BERTScore rank open models better than GPT-4V despite considering the human responses to be the best. This counter-intuitive observation might be attributed to the sensitivity of these methods to the differences in lexical variations in the reference and predicted responses \cite{bleurt}.

\subsubsection{Correlation Analysis}
\label{sec:correlation}

We measure the correlation between the candidate automatic metrics and human judgments using ROC-AUC and spearman correlation in Table \ref{tab:correlation}. Specifically, the human judgments are considered as gold standard where we assign `0' to unaccepted responses to the instructions and `1' to the accepted responses. We find that GPT-4 based evaluation achieves the highest ROC-AUC of $85.9$ and spearman correlation of $0.71$ amongst all the automatic evaluation metrics. In addition, we observe that GPT-4V also achieves a high correlation with the human judgments which is close to GPT-4. Specifically, GPT-4 bases its judgments on the given instruction and the reference response, whereas GPT-4V, with access to an input image, may potentially be biased. This access might lead GPT-4V to overlook the reference response and depend on the visual cues from the input image for making judgments in some cases. Finally, we observe that standard text generation metrics achieve a poor ROC-AUC and Spearman correlation in comparison to GPT-4 metrics. This corroborates the findings from the prior research \cite{visit-bench} that shows GPT-4 evaluation outperforms standard text generation metrics. Further, the dataset size is sufficient to get reliable confidence intervals on GPT-4 evaluation. We compared model predictions pairwise using the paired t-test at a 95\% confidence interval. The comparison between LLaVA-v1.5 and GPT4v/Gemini-Pro-Vision yielded a P value < 0.0001, suggesting that the difference in the performance is statistically significant. Comparing GPT4V with Gemini-Pro-Vision resulted in a P value of 0.035, also denoting statistical significance. Therefore, the differences in model performance on ConTextual are statistically significant at the 95\% confidence level. As a result, we utilize GPT-4 for automatically evaluate the quality of the predicted responses.

\subsection{Fine-Grained Evaluation}
\label{sec:fine_grained}

We compare the fine-grained performance of a wide range of foundation models across different visual contexts using GPT-4 evaluation in Table \ref{tab:fine_grained_GPT_4}. In our experiments, we find that GPT-4V outshines the baseline models in almost all categories. We observe that the sole exceptions are web usage and miscellaneous natural scenes contexts, where Gemini-Pro-Vision holds the lead. Notably, GPT-4V outperforms humans on reasoning over the abstract category, highlighting that it may have been tuned to reason over a lot of memes and quotes data. In addition, we observe that all the models struggle the most in the time category while humans ace it, a skill which is could be hard to learn from the training data. After time reading, the proprietary LMMs underperform on the infographics category which consists reasoning over data visualizations. Prior work \cite{mathvista,masry2023unichart} has shown that the existing LMMs underperform humans in reasoning over charts.

Further, we observe that the best performing open model LLaVA-Next-34B bridges the gap between the other open source models like LLaVA-1.5-13B and ShareGPT-4V-7B and the closed source models like Gemini-Pro-Vision. It performs the best on abstract and natural scenes, while it struggles the most on time and infographics. The relative imbalance in performance across categories can be attributed to the lack of diverse visual contexts in their training data. For instance, the COCO dataset \cite{coco} used for vision-language alignment in the open models predominantly comprises natural scenes. However, comparing it to its predecessor LLaVA-1.5-13B, improvement in visual encoding, data diversity, and LLM capacity boost performance on \name. We also observe the open models specifically introduced for text-rich visual reasoning like LLaVAR and BLIVA-Vicuna-7B falter on \name{} dataset. This indicates that these models cannot reason when the instruction requires them jointly over the text content and visual context in the image. We perform additional fine-grained evaluation in Appendix \S \ref{sec:additional_fine_grained}. Overall, our fine-grained analysis aids in identifying the gaps in the existing models which would inspire the development of next generation LMMs.

\subsection{Study on Synthetically Scaling Data}
Creating synthetic data for context-sensitive text-rich visual reasoning is challenging. Automatic dataset generation using OCR and image caption data with LLMs like GPT-4, exemplified by LLaVAR \cite{llavar}, yields instructions solvable by OCR+LLM or basic object understanding but shows poor performance on the context-sensitive instructions, as reported in Table \ref{tab:fine_grained_GPT_4}. Further, finding suitable images for joint reasoning is complex, requiring heuristic OCR filtering and manual selection. Despite these challenges, a synthetic data generation pipeline has been developed, showing model performance on synthetic data similar to human-annotated data, demonstrating the potential to scale context-sensitive instructions effectively (refer to \S \ref{app:data_scaling} for more details).

\begin{figure*}[h!]
\begin{minipage}[h]{0.55\textwidth}
  \fontsize{9.0pt}{\baselineskip}\selectfont
\linespread{0.9}\selectfont
\begin{mybody}
\includegraphics[scale=1.4]{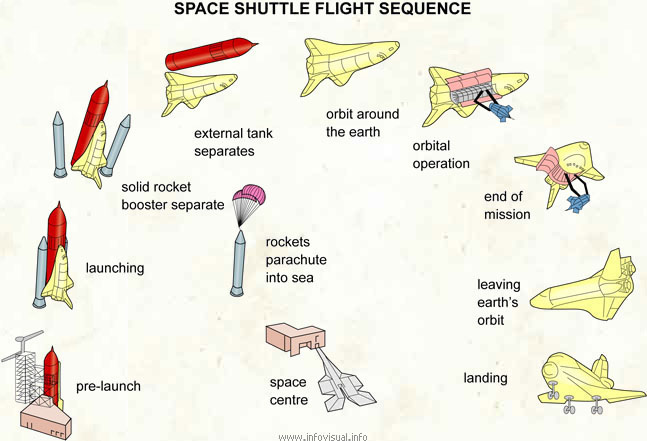}
\\
\textbf{Instruction:} Specify the colors of the rocket, space shuttle, and external tank.\\
\textbf{Reference Response:} 
The rocket is depicted in grey, space shuttle is depicted in yellow, and the external tank is depicted in red.
\end{mybody}
\begin{mybody}
\textbf{Human Response:} \textbf{\green{The rocket is gray. The space shuttle is yellow. The external tank is red.}}
\end{mybody}
\begin{mybody}
\textbf{GPT-4V Response:} 
In the image, the solid rocket boosters of the \textbf{\red{space shuttle are colored in red and light gray. The external tank is colored orange. The main body of the space shuttle is primarily white.}} \\
\textbf{Summarized Reasoning:} During the pre-launch, all components are vertical. Solid rocket boosters detach when the shuttle ascends. Once the shuttle reaches space's edge, the external tank is released. The shuttle's colors remain consistent throughout these phases.
\end{mybody}
\captionof{figure}{Here, GPT-4V responds incorrectly to the instruction despite its logical reasoning. \green{Green} denotes correct response, and \red{red} highlights response errors. \textit{Summarized Reasoning}: rationale used by GPT-4V to answer.}
\label{fig:GPT-4V_vs_humans}
\end{minipage}
  \hspace{0.05\textwidth}
\begin{minipage}[h]{0.38\textwidth}
\fontsize{9.0pt}{\baselineskip}\selectfont
\linespread{0.9}\selectfont
\begin{mybody}
\includegraphics[scale=0.36]{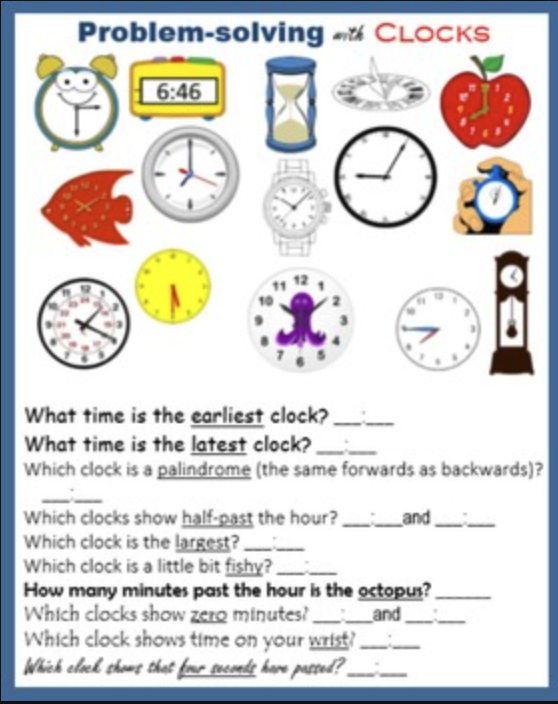}
\\
\textbf{Instruction:} Identify the object on the dial that shows the time 10:10.\\
\textbf{Reference Response:} 
The clock at 10:10 features an octopus on its dial. 
\end{mybody}
\begin{mybody}
\textbf{GPT-4V Response:} The clock with the image of \textbf{\green{an octopus}} on the face is showing the time 10:10.
\end{mybody}
\begin{mybody}
\textbf{LLaVA-Next-34B:} The object on the dial that shows the time 10:10 is the \textbf{\red{hour hand and the minute hand}}. 
\end{mybody}
\begin{mybody}
\textbf{GPT-4 w/ Layout-aware OCR + Caption Response:} \textbf{\red{The instruction does not provide enough specific information to identify the object on the dial that shows the time 10:10.}}

\end{mybody}
\captionof{figure}{GPT-4V correctly responds to the instruction. However, LLaVA-Next-34B and GPT-4 w/ Layout-aware OCR+ Caption (Augmented LLM) produce wrong responses.}
\label{fig:GPT-4V_vs_others}
\end{minipage}
\end{figure*}

\section{Qualitative Examples}
\label{sec:qualitative_examples}

\paragraph{GPT-4V vs Humans.} In Figure \ref{fig:GPT-4V_vs_humans}, we see an instance where GPT-4V provides an incorrect answer. Here, the model is asked to identify the colors of different parts of a space launch vehicle - space shuttle, external tank, and rocket thrusters. GPT-4V makes errors in color predictions but can accurately infer the diagram's information, revealing a lack of precise visual perception. We provide more examples in Appendix \S \ref{app:qualitative_examples} (Figures \ref{fig:time_wrong_1}, \ref{fig:shopping_wrong_1}, \ref{fig:navigation_wrong_1}, \ref{fig:navigation_wrong_2}, \ref{fig:infographic_wrong_2}, \ref{fig:misc_wrong_1}, \ref{fig:misc_wrong_2}), highlights that GPT-4V's core issue lies in fine-grained perception coupled with a bias for prior visual knowledge. A similar analysis was presented in the prior work \cite{guan2023hallusionbench} where GPT-4V fails on the perturbed versions of common visual illusions.

\paragraph{GPT-4V vs. Open LMMs and Augmented LLM.}
We compare the best-performing models in each category, closed-source LMM, open-source LMM, and Augmented LLM approach, that is, GPT-4V, LLaVA-Next-34B, and GPT-4 w/ Layout-aware OCR + Caption, respectively, using an example illustrated in Figure \ref{fig:GPT-4V_vs_others}. GPT-4V correctly identifies the object, showcasing superior visual perception and context-sensitive text-rich visual reasoning abilities over the LLaVA-Next and Augmented LLM approach that produces the wrong response. LLaVA-Next does not ground its response to the image due to relatively poor context-sensitive text-rich visual reasoning abilities. On the other hand, the Augmented LLM approach cannot respond to this instruction because the image caption and layout-aware OCR do not provide sufficient information to reason over embedded text and visual elements in the image. We refer to Figure \ref{fig:shopping_correct_2}, \ref{fig:navigation_correct_1}, \ref{fig:navigation_correct_2}, \ref{fig:app_usage_correct_2}, \ref{fig:web_usage_correct_2}, \ref{fig:infographic_correct_1}, \ref{fig:infographic_correct_2}, \ref{fig:misc_correct_2} for more examples demonstrating instances where open models exhibit lack of context-sensitive text-rich visual reasoning, or deficiencies in fine-grained perception.

Our analysis suggests that enhancing image encoders and increasing training data diversity can improve model perception, leading to more effective context-sensitive reasoning in text-rich visual contexts.

\section{Related Work}
\paragraph{Text-Rich Image Understanding.} Recently, there has been a growing interest in understanding the interactions between the text and visual elements in the image \cite{lee2023pix2struct,xu2020layoutlm}. To track the progress of the models in this field, several datasets were introduced like OCRVQA \cite{ocrvqa}, TextVQA \cite{textvqa}, DocVQA \cite{docvqa}, STVQA \cite{stvqa}, ESTVQA \cite{estvqa}. These datasets majorly focus on the ability of the models to accurately read the text in the documents or natural scene images. Prior work \cite{liu2023hidden} provides a benchmark to assess the ability of LMMs to perform accurate OCR. In comparison, we propose a new \name dataset, comprising a wide range of visual contexts, instruction types (questions and imperative tasks), that aims to test the LMM's ability to perform precise visual perception and complex reasoning over the visual and text elements of the image. 

\paragraph{Vision Language Reasoning Benchmarks.} Having high-quality datasets is essential to assess the progress of the fields towards building high utility models for the real-world. Traditionally, vision-language learning has focused on tasks such as visual question answering \cite{antol2015vqa,goyal2017making}, image captioning \cite{gurari2018vizwiz,coco} where the model primarily needs to understand the key objects and their relations. Later, several works were introduced to assess the commonsense reasoning, which requires the models to reason about the questions that require skills beyond recognition, including VCR \cite{zellers2019recognition,yin2021broaden}. In addition, there are several datasets and benchmarks that evaluate specific skills of the LMMs including math skills \cite{chen2022unigeo,lu2021inter,lu2021iconqa,mathvista}, world knowledge \cite{yue2023mmmu}, and grade school science diagrams \cite{kembhavi2016diagram,wang2023scibench}. Additionally, there are several datasets for meme understanding such as hateful memes \cite{kiela2020hateful}, memeifiy \cite{vyalla2020memeify}, and memecap \cite{hwang2023memecap}. Such works will require joint reasoning over text and visual content over the image. However, in our work, we broaden the scope and identify a breadth of visual domains that require context-sensitive text-rich visual reasoning. These include time reading, navigation and transportation in public spaces, meme and quote understanding, and shopping etc.


\paragraph{Large Multimodal Models.} Prior works such as LXMERT \cite{tan2019lxmert}, VisualBERT \cite{li2019visualbert}, X-decoder \cite{zou2023generalized} learn robust vision-language representations by training on image-text data such as Conceptual captions \cite{cc3m}, COCO \cite{coco}. Post-training, they will be finetuned on the specific tasks such as VQA \cite{antol2015vqa}. For document understanding, popular vision-language models include Pix2Struct \cite{lee2023pix2struct}, Donut \cite{kim2022ocr}, MatCha \cite{liu2022matcha}. However, the development of large language models \cite{gpt3,gpt4}, trained on internet-scale text corpus, shifted the paradigm towards the development of general-purpose multimodal models. Specifically, these are vision-language generative models that can solve diverse tasks in a zero-shot manner without task-specific finetuning. Notably, these are popularly known as large multimodal models (LMMs). These include proprietary models such as GPT-4V \cite{gpt4v} and Gemini-Pro-Vision \cite{team2023gemini}. These models have achieved state-of-the-art performance on the traditional vision-language models. In the open space, the models include LLaVA \cite{llava, llavav15, llavanext}, mPLUG-Owl \cite{mplugowl}, OpenFlamingo \cite{open-flamingo}, Idefics \cite{idefics}, LLaMA-Adapter \cite{llamaadapter}, Idefics \cite{idefics}. In addition, there are a class of LMMs that focus on enhanced text-rich visual reasoning capabilities including LLaVAR \cite{llavar} and BLIVA \cite{bliva}. In this work, we compare the performance of LMMs on the \name dataset. We find that the text-rich visual reasoning capabilities of the proprietary models is way superior than the open models. We also include fine-grained analysis to understand the gaps in the model performance across different visual contexts. 


\section{Conclusion}

In this work, we introduce \name, a dataset for evaluating the text-rich visual reasoning in large multimodal models. Going beyond the prior efforts that focus primarily on the testing the reading skills in the visual contexts, we create novel and challenging instructions from scratch that would require the models to capture the context in which the text is presented in an image. We ask humans to solve our dataset and also use human annotators for model response evaluation. We find that the modern LMMs (proprietary and open models) struggle to perform on our dataset while humans are good at it. In summary, our dataset paves a path for assessing the progress on reasoning over text-rich images, a domain with significant real-world applications. We make the dataset\footnote{\href{https://huggingface.co/datasets/ucla-contextual/contextual_all}{Hugging Face Dataset}} and code\footnote{\href{https://github.com/rohan598/ConTextual}{GitHub Code Repository}} available to the LMM community along with a continuously updated leaderboard \footnote{\href{https://huggingface.co/spaces/ucla-contextual/contextual_leaderboard}{HuggingFace Leaderboard}} with recent LMMs.

\section*{Acknowledgements}

This research is supported in part by the ECOLE program under Cooperative Agreement
HR00112390060 with the US Defense Advanced Research Projects Agency (DARPA), Google Research Scholar and UCLA-Amazon Science Hub for Humanity and Artificial Intelligence. Hritik Bansal is
supported in part by AFOSR MURI grant FA9550-22-1-0380.

\section*{Impact Statement}

\name is proposed to evaluate the context-sensitive text-rich visual reasoning capabilities of large multimodal models. These models are a class of generative models provide textual response to user instructions, grounded in text, for diverse images. During our data collection, we aim to ensure that the images, human-written instructions, and reference responses are not offensive to any social group. We are aware that the existing multimodal models are capable of generating harmful responses, despite the presence of safeguard filter. In addition, our qualitative analysis reveals that the model responses would hallucinate, however, we did not observe any apparent harmful and privacy sensitive information in them. 

In our experiments, we asked human annotators, mainly from the US, to provide responses to establish a human baseline. We are aware that the linguistic diversity and writing style of the human responses would change with different social groups. The extension of our work can focus on understanding the impact of different social groups on the human baseline performance on the \name dataset. A similar argument is relevant for human evaluation of the model responses. To obtain more reliable human evaluation results, future work would involve annotators from more diverse regions. 

\bibliographystyle{icml2024}
\bibliography{icml_2024}
\clearpage
\appendix
\onecolumn
\section{Dataset Details}
\label{app:data_details}


\subsection{Visual Scenarios Description}
\label{app:visual_scenarios_description}
In this section, we outline the constituent elements that make up each visual scenario, as illustrated in Table \ref{tab:contextual_details}.

\begin{table*}[h!]
    \centering
    \renewcommand{\arraystretch}{1.2}
    \resizebox{0.75\linewidth}{!}{
    \begin{tabular}{p{0.15\linewidth}p{0.85\linewidth}}
     \hline
        \textbf{Category} & \textbf{Description}\\
        \hline
         Shopping & Purchasing groceries, clothes, furniture, gadgets, cosmetics, services, and miscellaneous products. \\
         \hline
         Navigation & Different modes of transportation - passenger vehicles, trucks, buses, trains, and airplanes, and navigation signage - streets, roadways, bus stations, train stations, and airports. \\
         \hline
         Time & Items showcasing time and dates, including analog clocks, digital clocks, multi-clock setups, calendars, and other miscellaneous time-viewing setups. \\
         \hline
         Web Usage & Websites across a variety of domains, like new articles, blogs, sports, and e-commerce \\
         \hline
         App Usage & Smartphone applications on education, productivity, games, lifestyle, entertainment, news, etc. \\
         \hline
         Infographic & Infographics on local and global information spanning domains of health, sports, education, natural resources, technology, etc. \\
         \hline
         Abstract & Memes, comic strips, and other abstract concepts illustrated through text-rich images. \\
         \hline
         Miscellaneous Natural Scenes & Miscellaneous human interactions do not fall into the previous categories.\\
          \hline
    \end{tabular}
    }
    \caption{Descriptions of the eight visual scenarios in \name.}
    \label{tab:contextual_details}
\end{table*}

\subsection{Visual Scenarios Examples}
\label{app:visual_scenarios_examples}
In this section, we provide examples of each visual category in \textbf{\name}.

\begin{figure*}[h!]
    \centering
    \subfloat[\centering Single Clock]{{\includegraphics[width=0.2\linewidth]{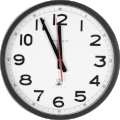}}}
    \hspace{8mm}
    \subfloat[\centering Multiple Clocks]{{\includegraphics[width=0.2\linewidth]{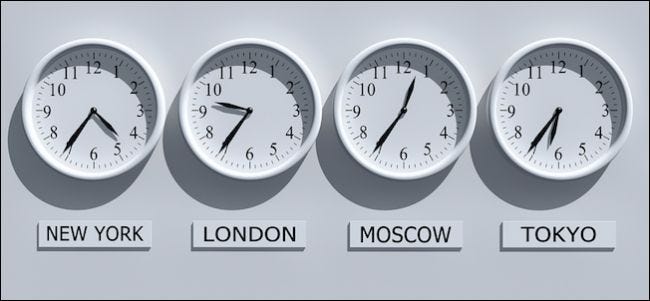}}}
    \hspace{8mm}
        \subfloat[\centering Calendar]{{\includegraphics[width=0.2\linewidth]{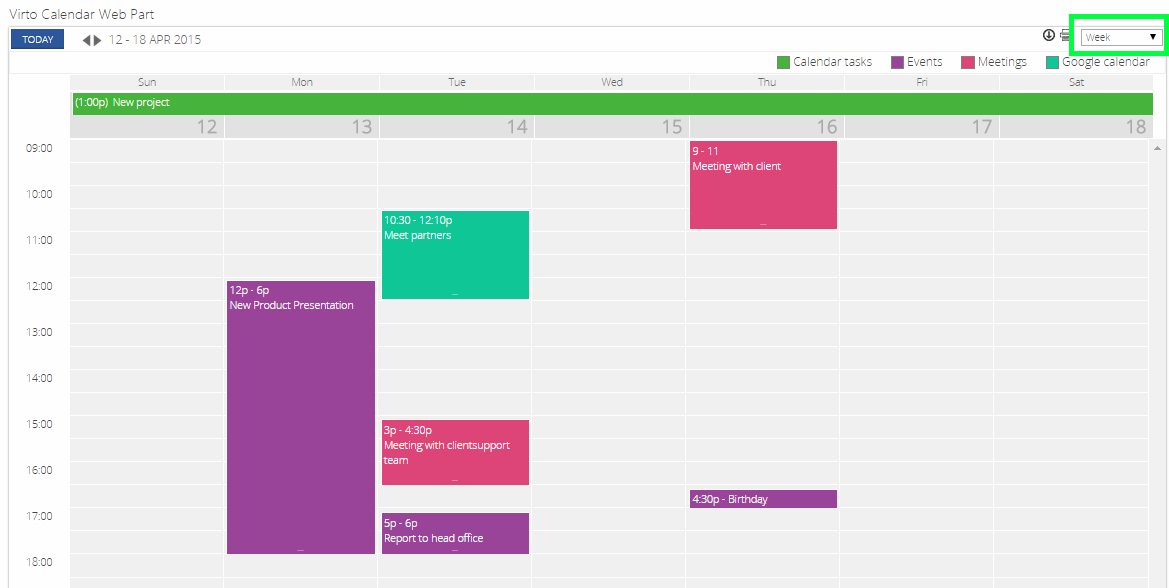}}}
        \hspace{8mm}
        \subfloat[\centering Timer]{{\includegraphics[width=0.2\linewidth]{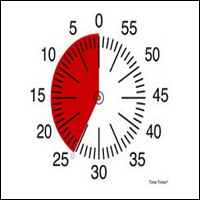}}}
    \caption{Examples of the \textit{Time} visual Scenario}
    \label{fig:time_examples}
\end{figure*}


\begin{figure*}[h!]
    \centering
    \subfloat[\centering Grocery]{{\includegraphics[width=0.15\linewidth]{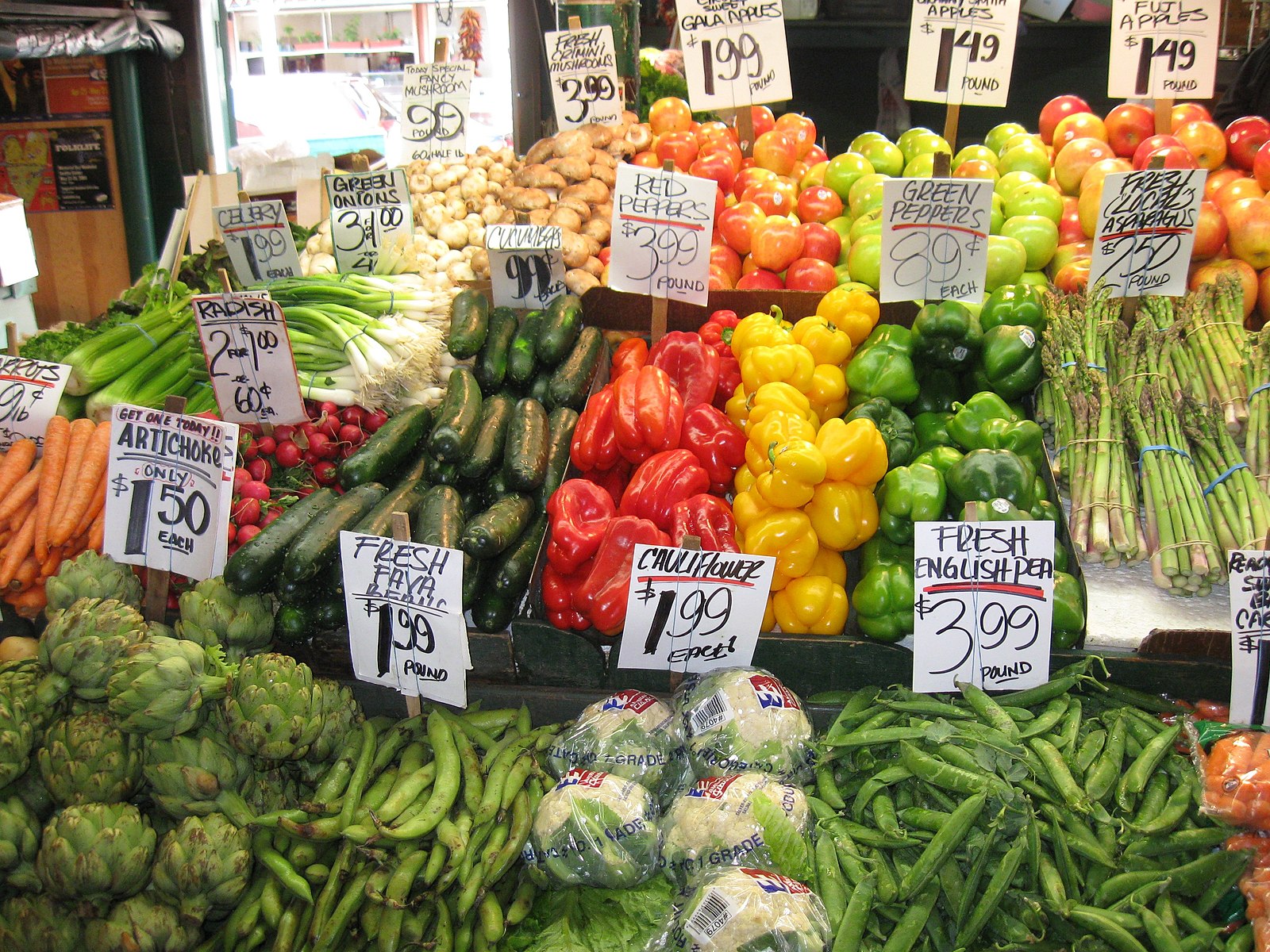}}}
    \hspace{0.01\linewidth}
    \subfloat[\centering Furniture]{{\includegraphics[width=0.15\linewidth]{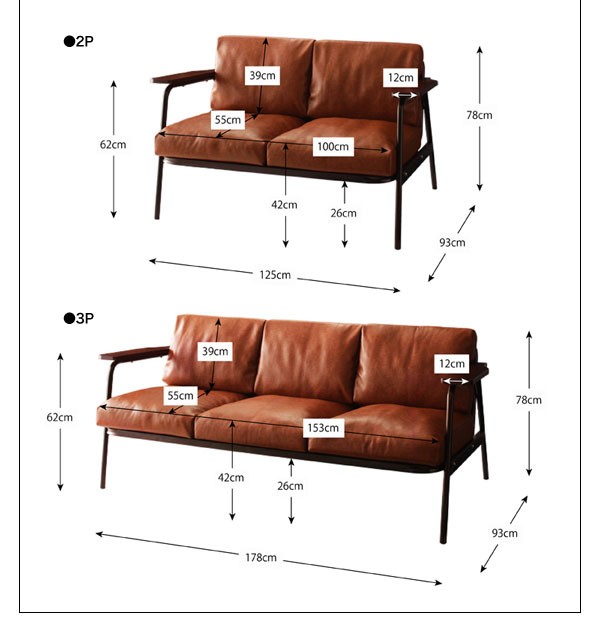}}}
    \hspace{0.01\linewidth}
        \subfloat[\centering Clothes]{{\includegraphics[width=0.15\linewidth]{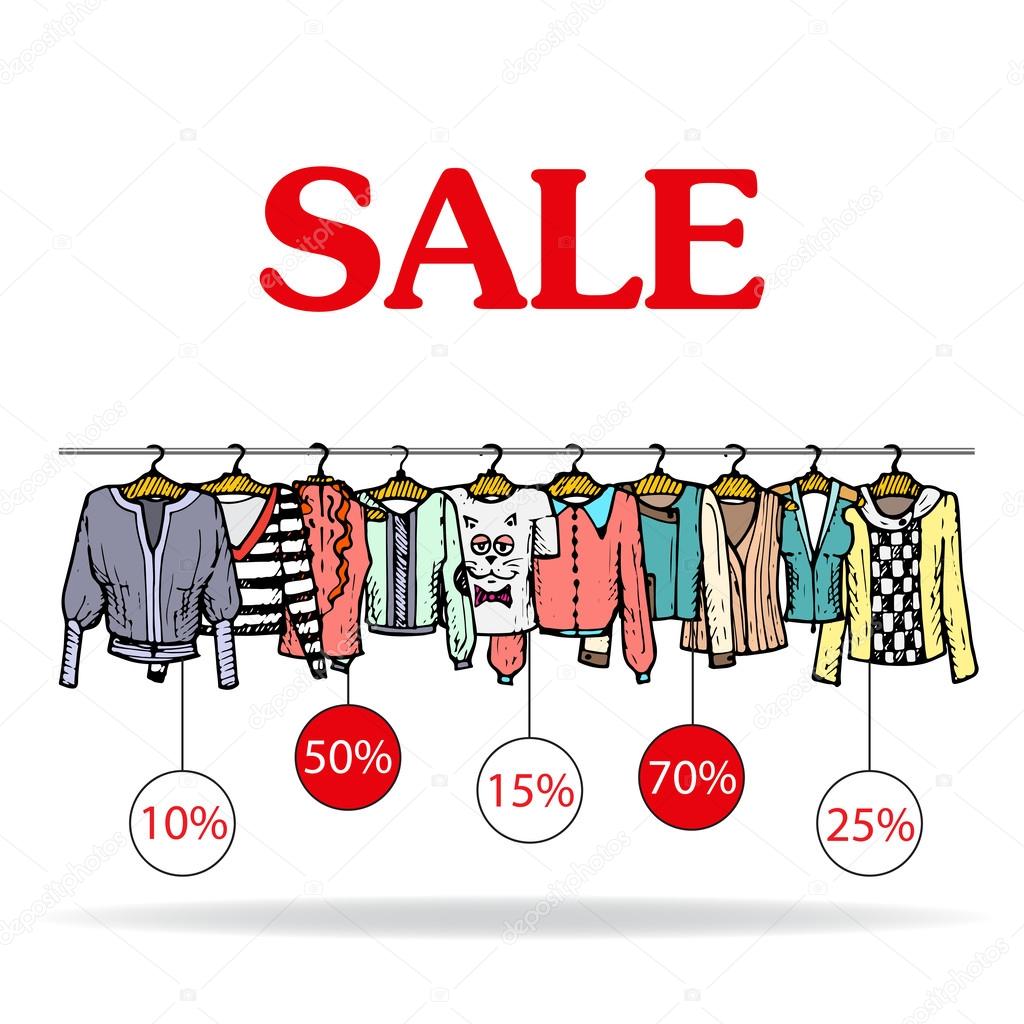}}}
        \hspace{0.01\linewidth}
        \subfloat[\centering Gadgets]{{\includegraphics[width=0.15\linewidth]{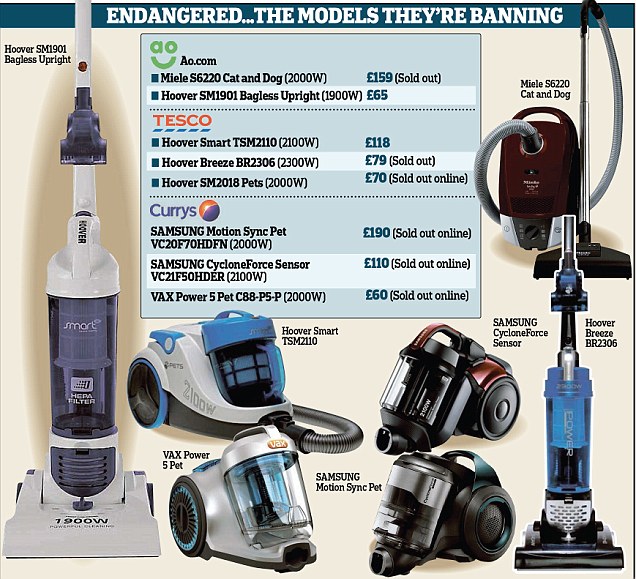}}}
            \hspace{0.01\linewidth}
        \subfloat[\centering Cosmetics]{{\includegraphics[width=0.15\linewidth]{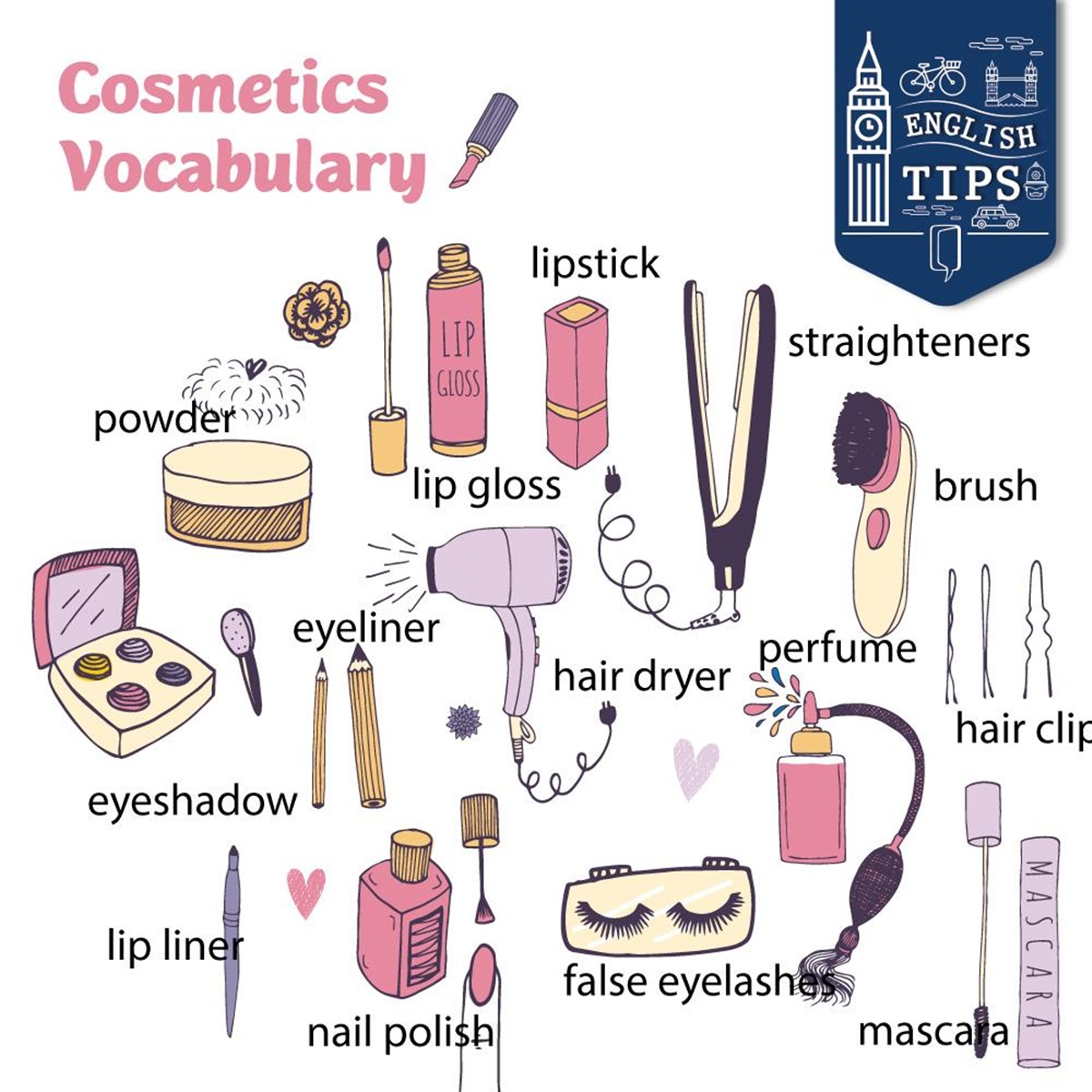}}}
            \hspace{0.01\linewidth}
        \subfloat[\centering Services]{{\includegraphics[width=0.15\linewidth]{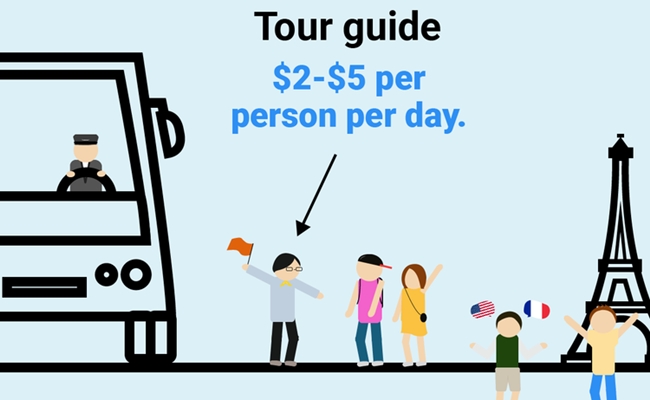}}}
    \caption{Examples of the \textit{Shopping} visual scenario}
    \label{fig:shopping_examples}
\end{figure*}


\begin{figure*}[h!]
    \centering
    \subfloat[\centering Street]{{\includegraphics[width=0.18\linewidth]{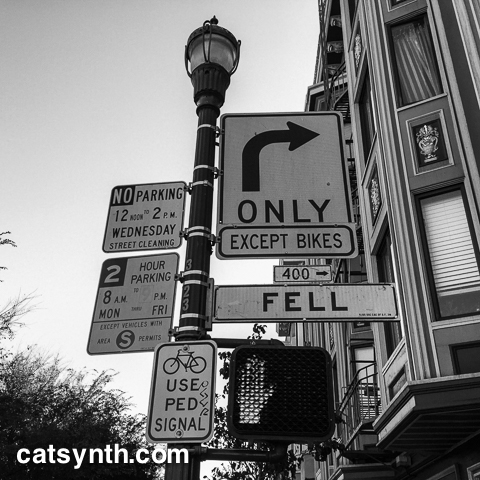}}}
    \hspace{0.01\linewidth}
    \subfloat[\centering Car]{{\includegraphics[width=0.18\linewidth]{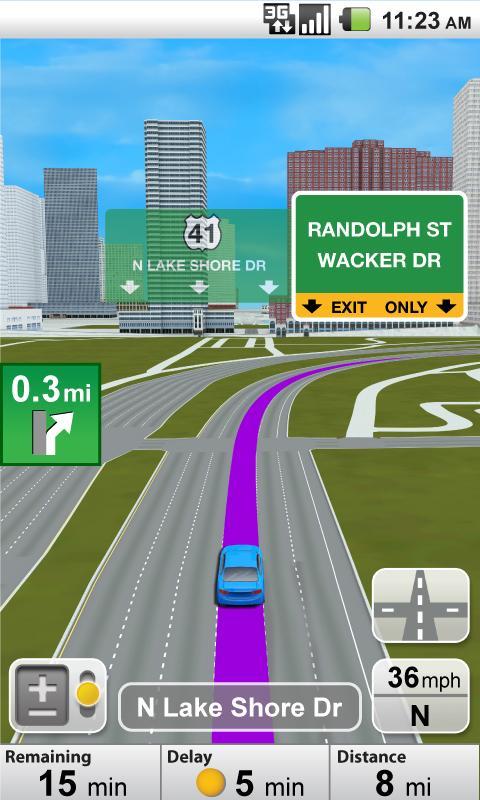}}}
    \hspace{0.01\linewidth}
        \subfloat[\centering Bus]{{\includegraphics[width=0.18\linewidth]{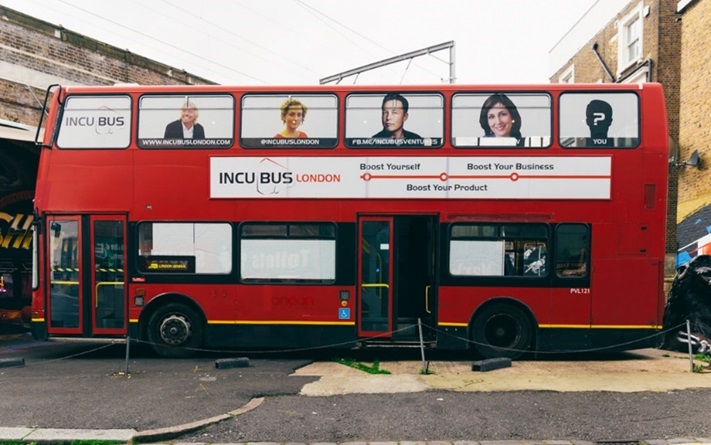}}}
        \hspace{0.01\linewidth}
        \subfloat[\centering Train]{{\includegraphics[width=0.18\linewidth]{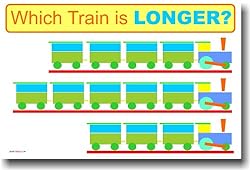}}}
            \hspace{0.01\linewidth}
        \subfloat[\centering Airport]{{\includegraphics[width=0.18\linewidth]{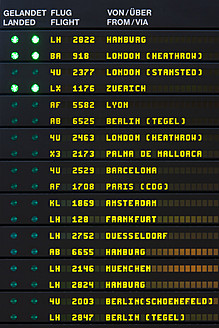}}}
    \caption{Examples of the \textit{Navigation} visual scenario}
    \label{fig:navigation_examples}
\end{figure*}


\begin{figure*}[h!]
    \centering
    \subfloat[\centering]{{\includegraphics[width=0.2\linewidth]{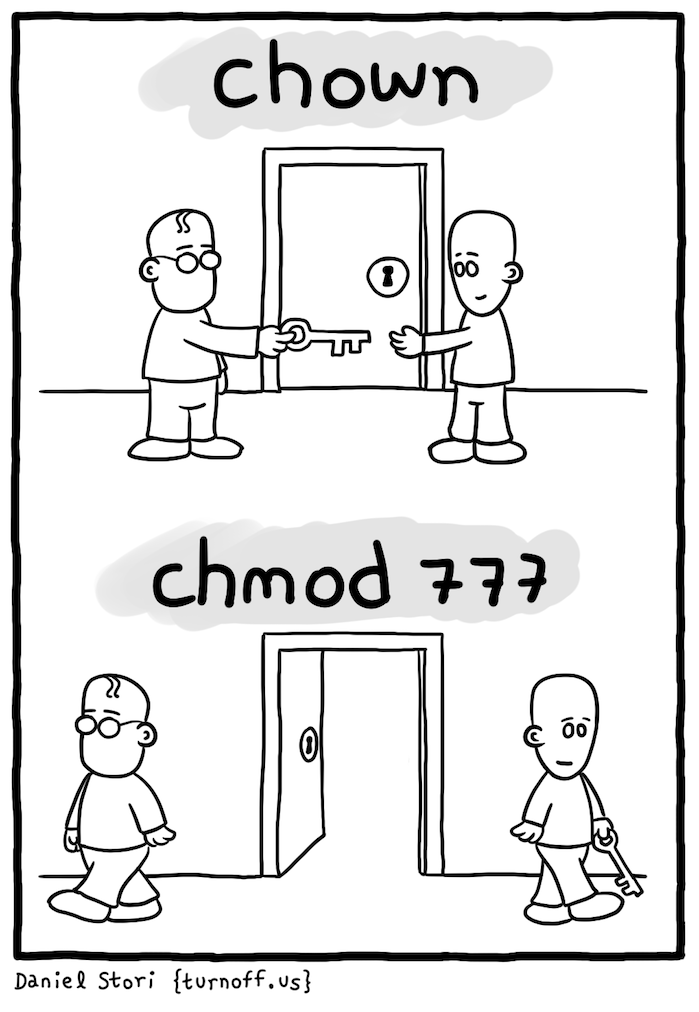}}}
    \hspace{8mm}
    \subfloat[\centering]{{\includegraphics[width=0.2\linewidth]{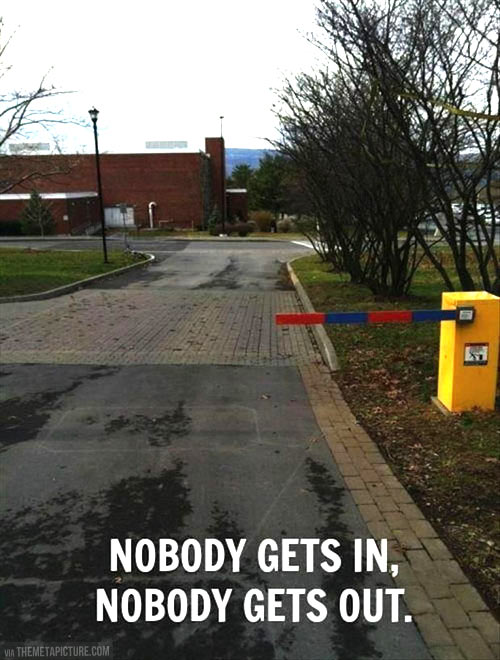}}}
    \hspace{8mm}
        \subfloat[\centering]{{\includegraphics[width=0.2\linewidth]{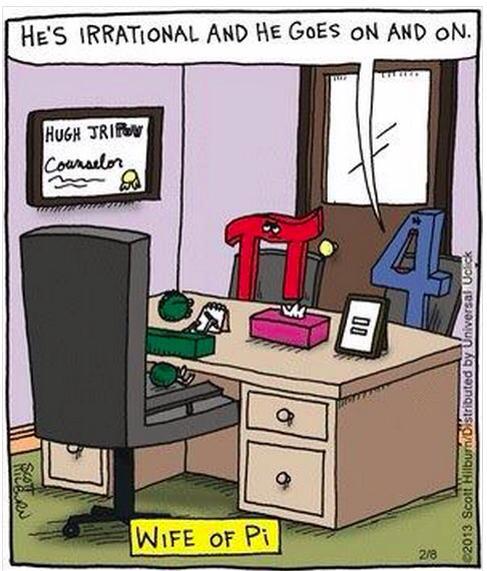}}}
        \hspace{8mm}
        \subfloat[\centering]{{\includegraphics[width=0.2\linewidth]{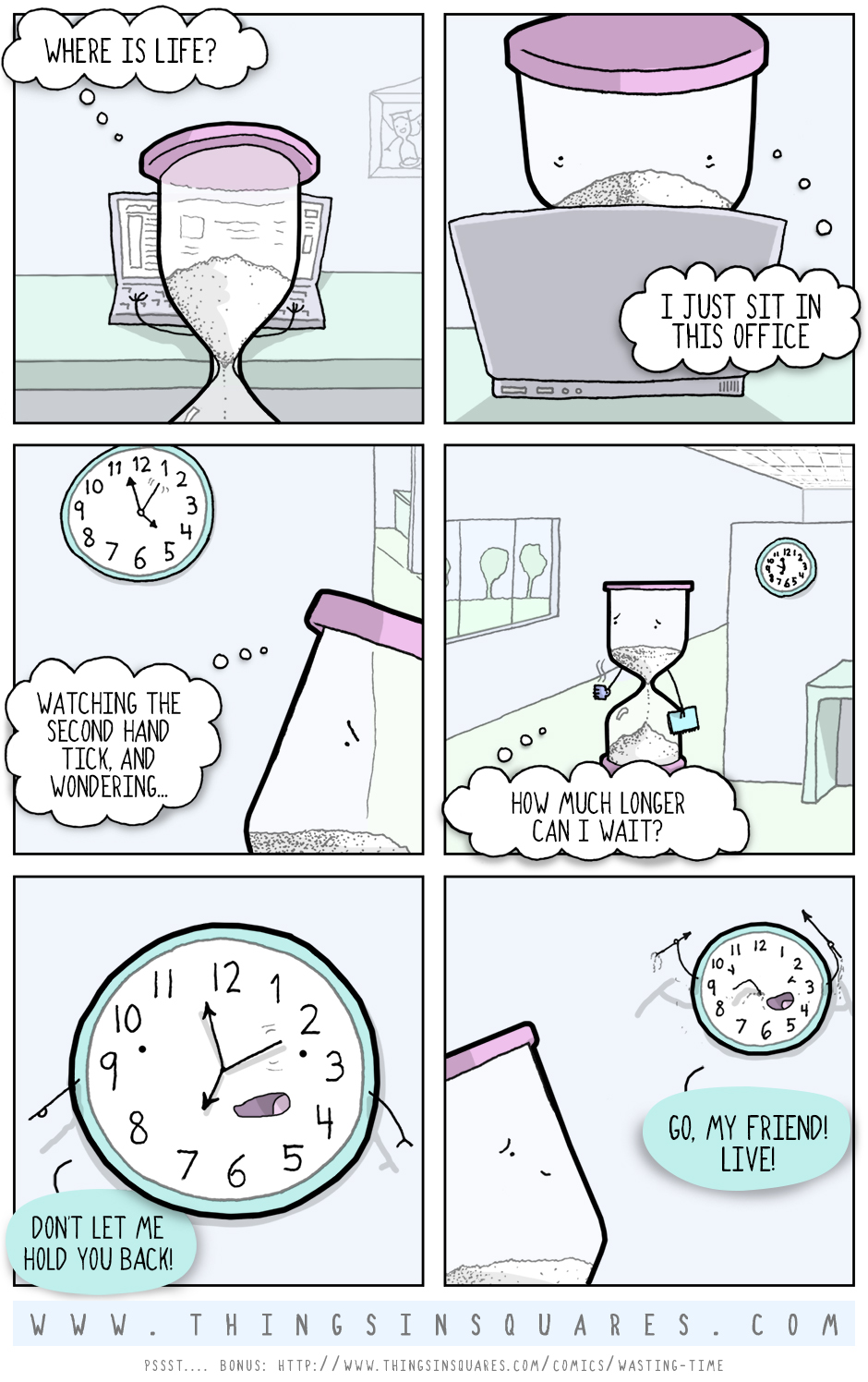}}}
    \caption{Examples of the \textit{Abstract} visual scenario}
    \label{fig:abstract_examples}
\end{figure*}


\begin{figure*}[h!]
    \centering
    \subfloat[\centering]{{\includegraphics[width=0.2\linewidth]{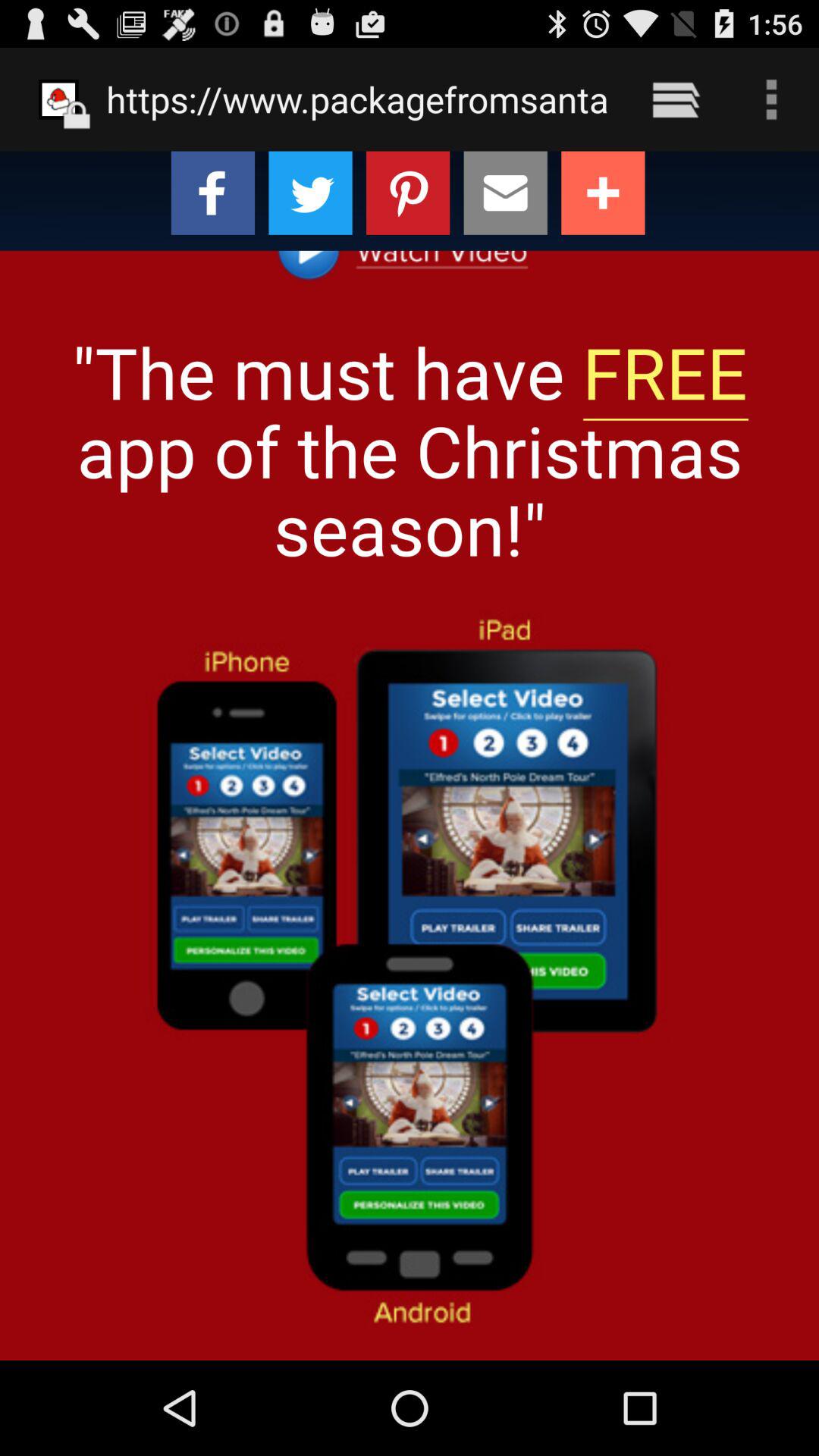}}}
    \hspace{8mm}
    \subfloat[\centering]{{\includegraphics[width=0.2\linewidth]{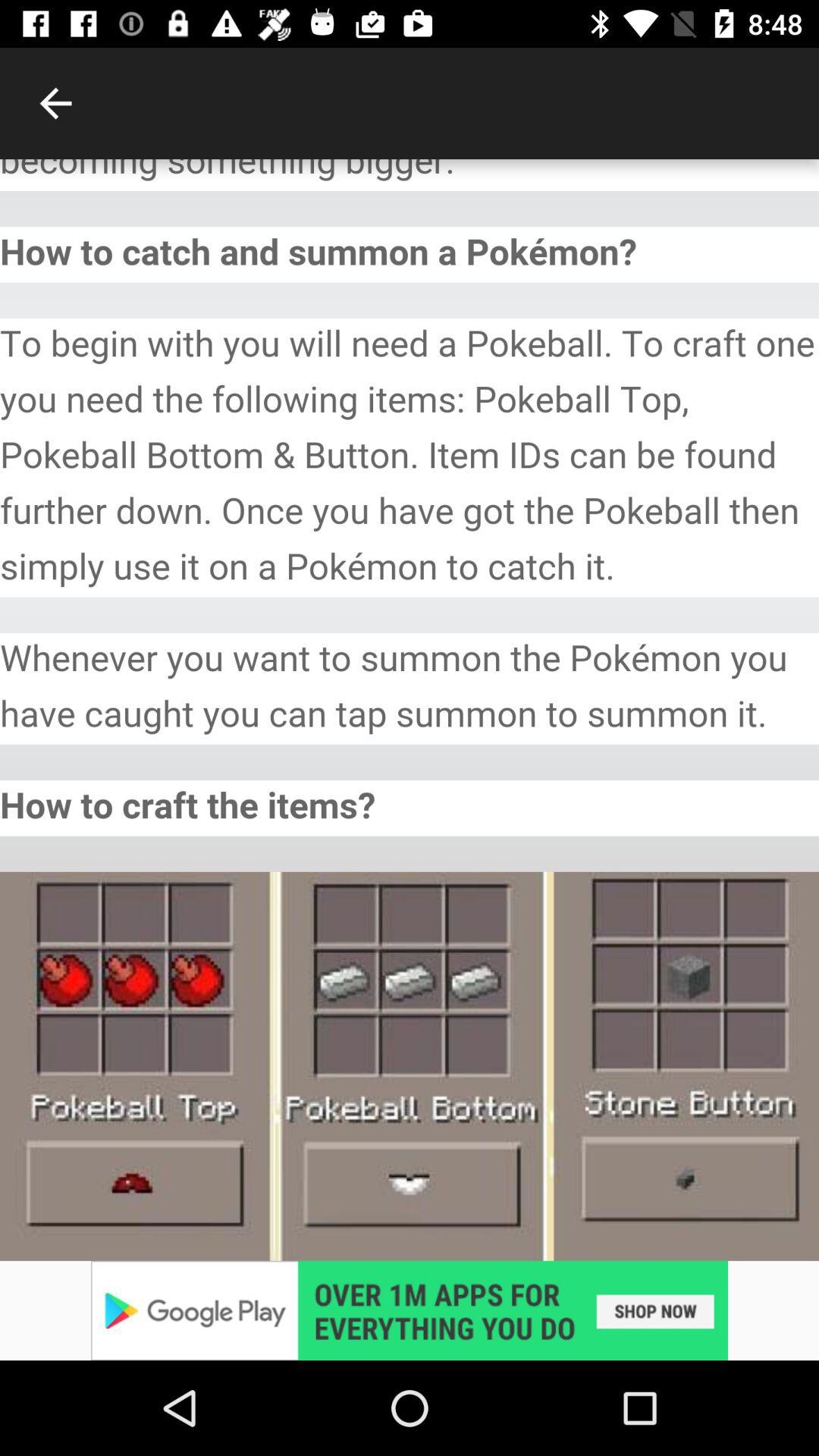}}}
    \hspace{8mm}
        \subfloat[\centering]{{\includegraphics[width=0.2\linewidth]{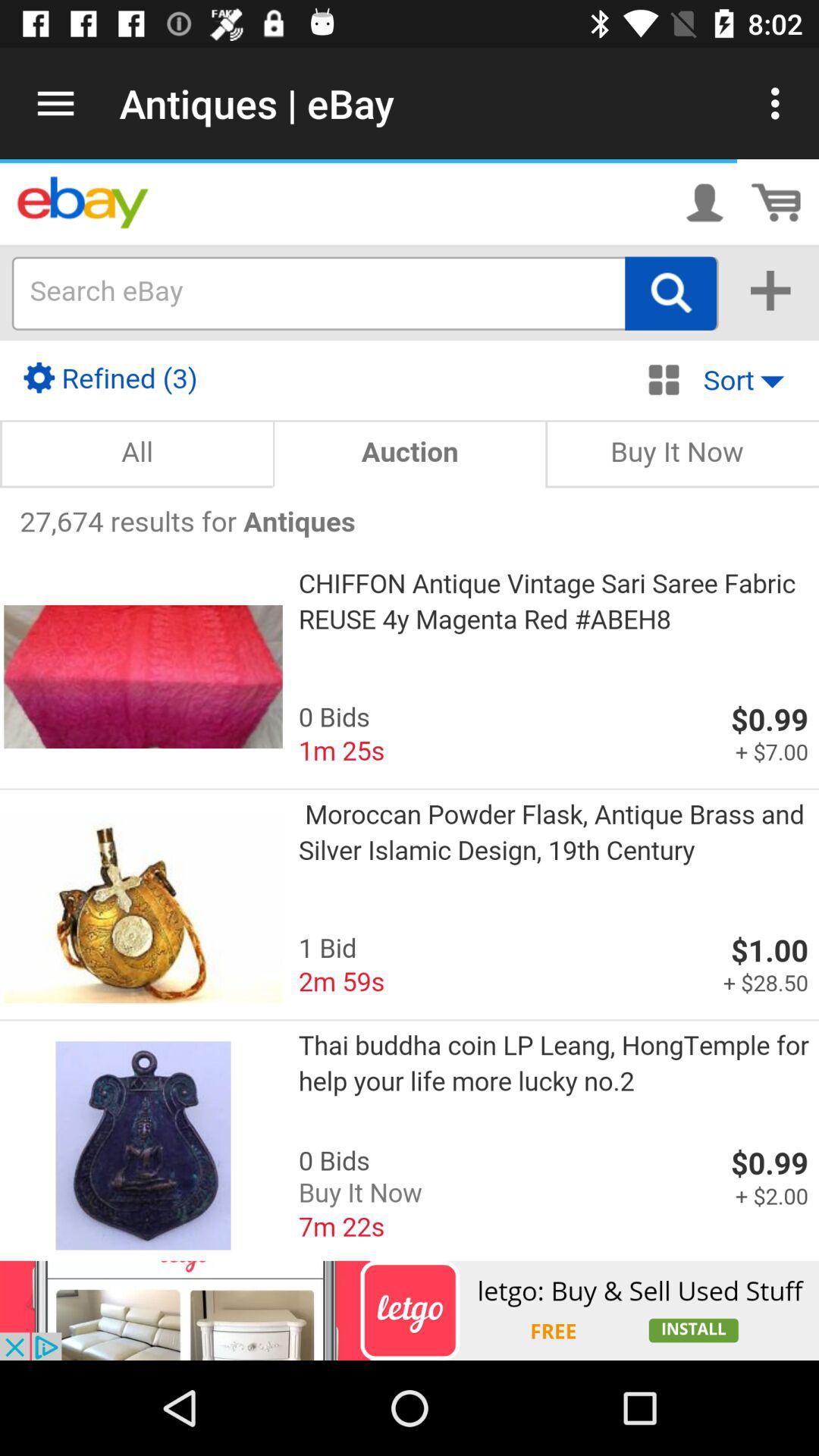}}}
        \hspace{8mm}
        \subfloat[\centering]{{\includegraphics[width=0.2\linewidth]{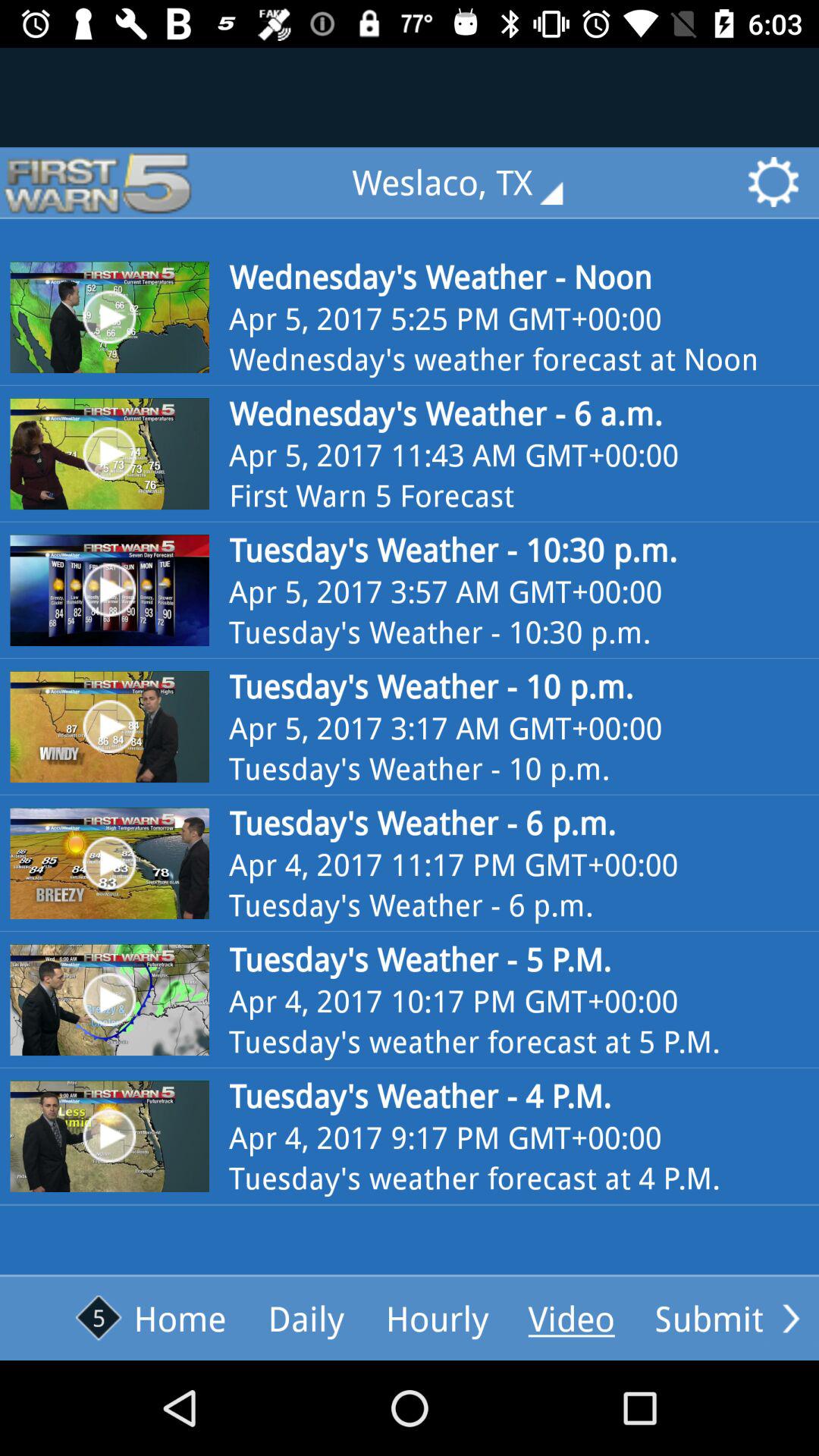}}}
    \caption{Examples of the \textit{Mobile Usage} visual scenario}
    \label{fig:app_usage_examples}
\end{figure*}


\begin{figure*}[h!]
    \centering
    \subfloat[\centering]{{\includegraphics[width=0.2\linewidth]{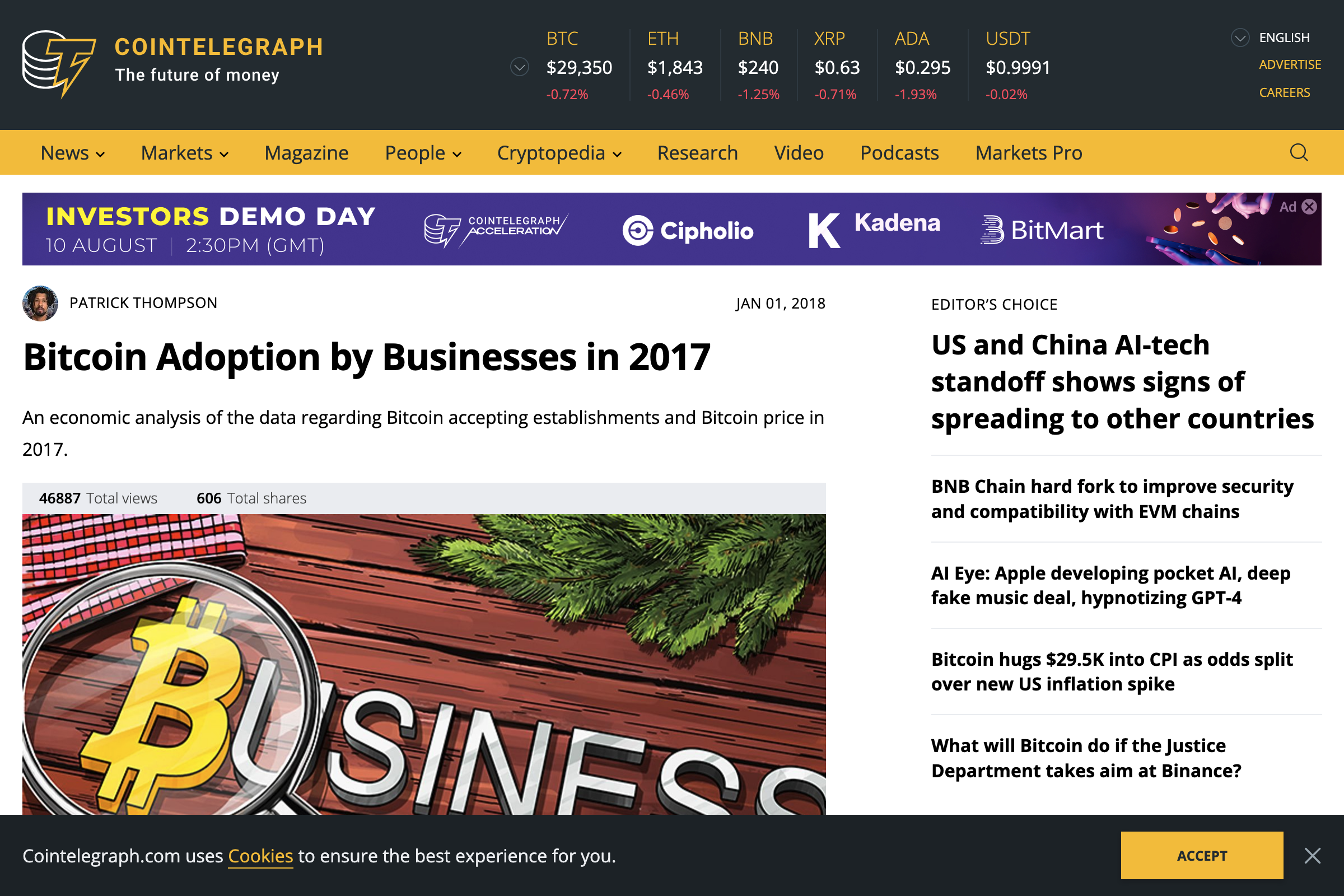}}}
    \hspace{8mm}
    \subfloat[\centering]{{\includegraphics[width=0.2\linewidth]{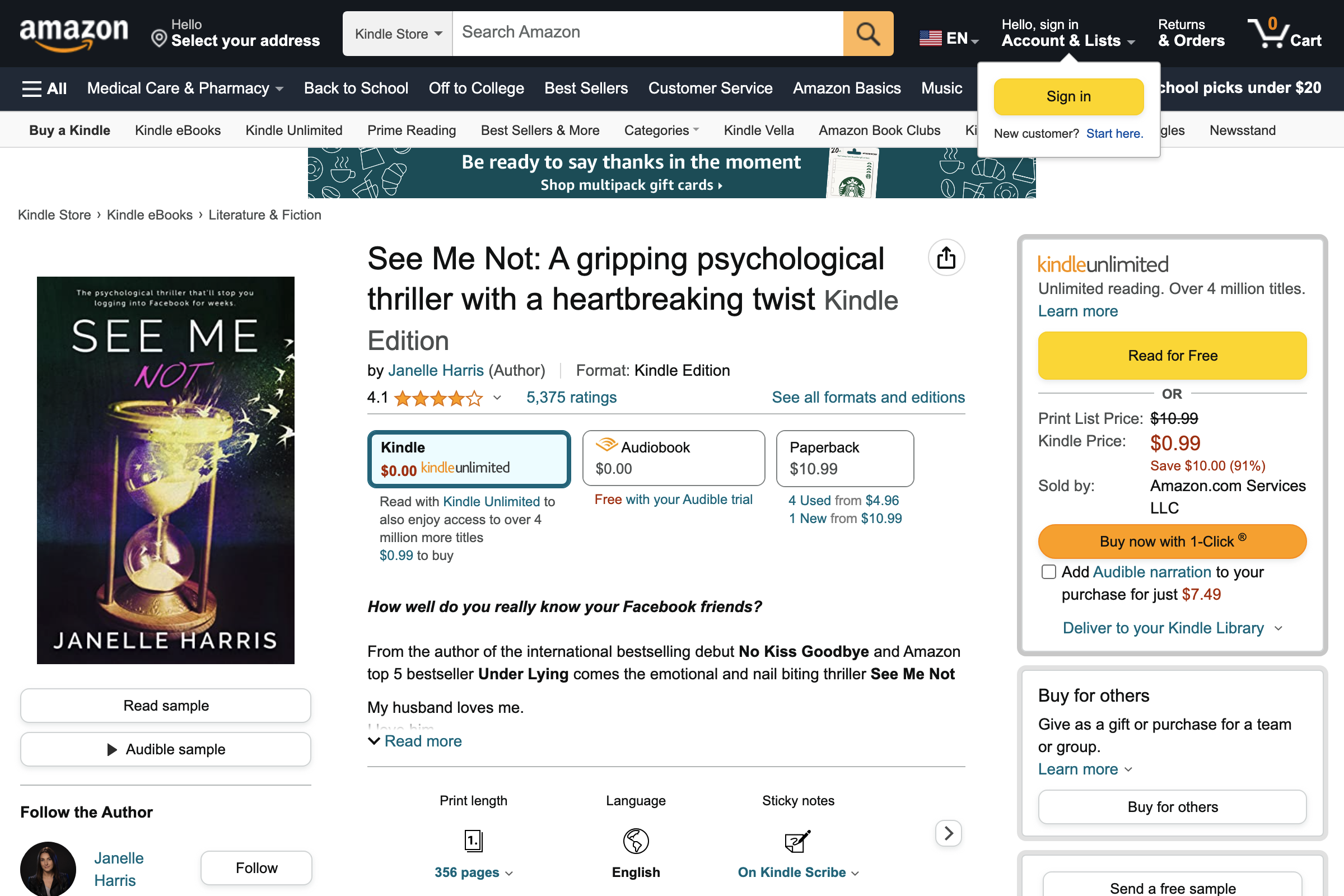}}}
    \hspace{8mm}
        \subfloat[\centering]{{\includegraphics[width=0.2\linewidth]{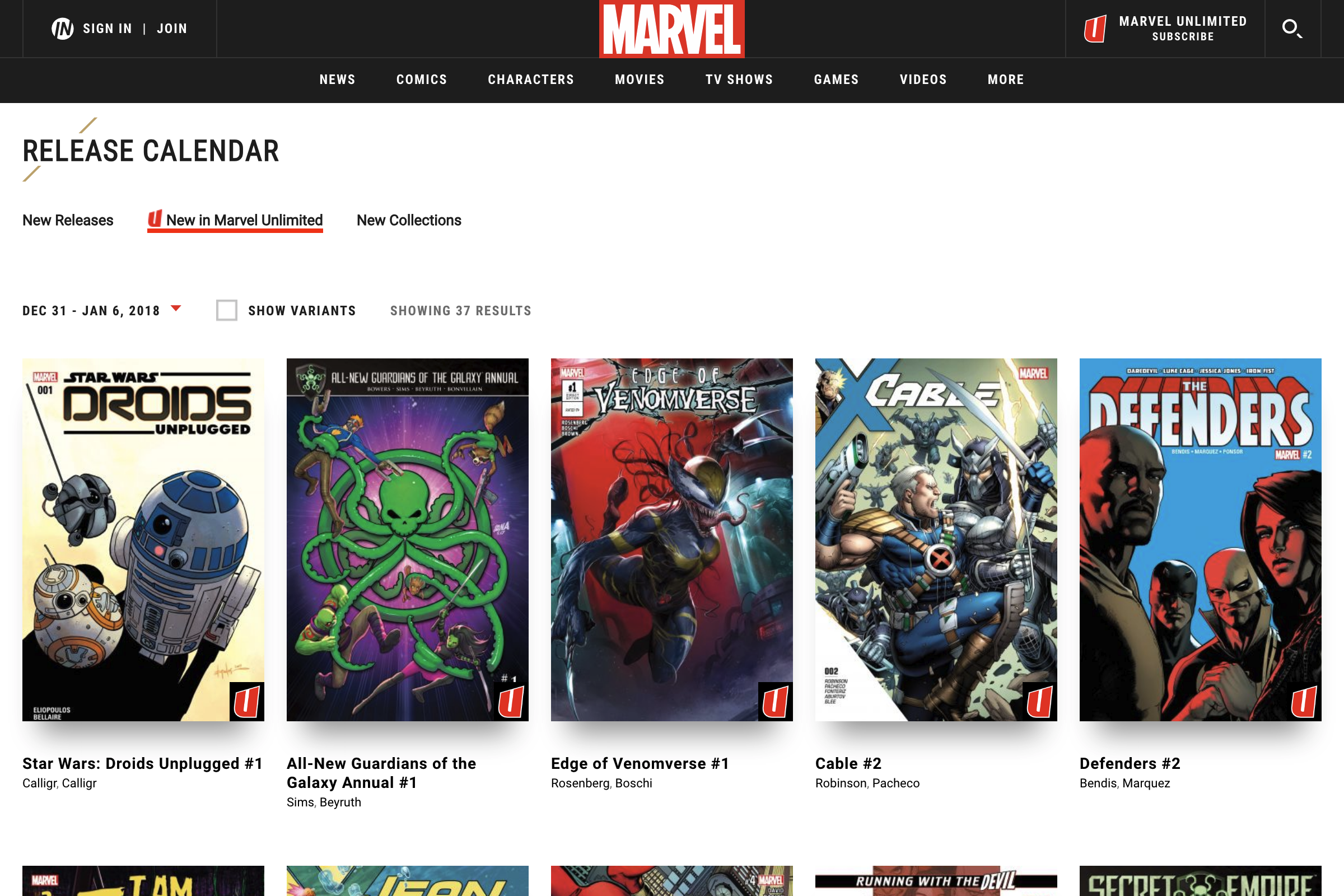}}}
        \hspace{8mm}
        \subfloat[\centering]{{\includegraphics[width=0.2\linewidth]{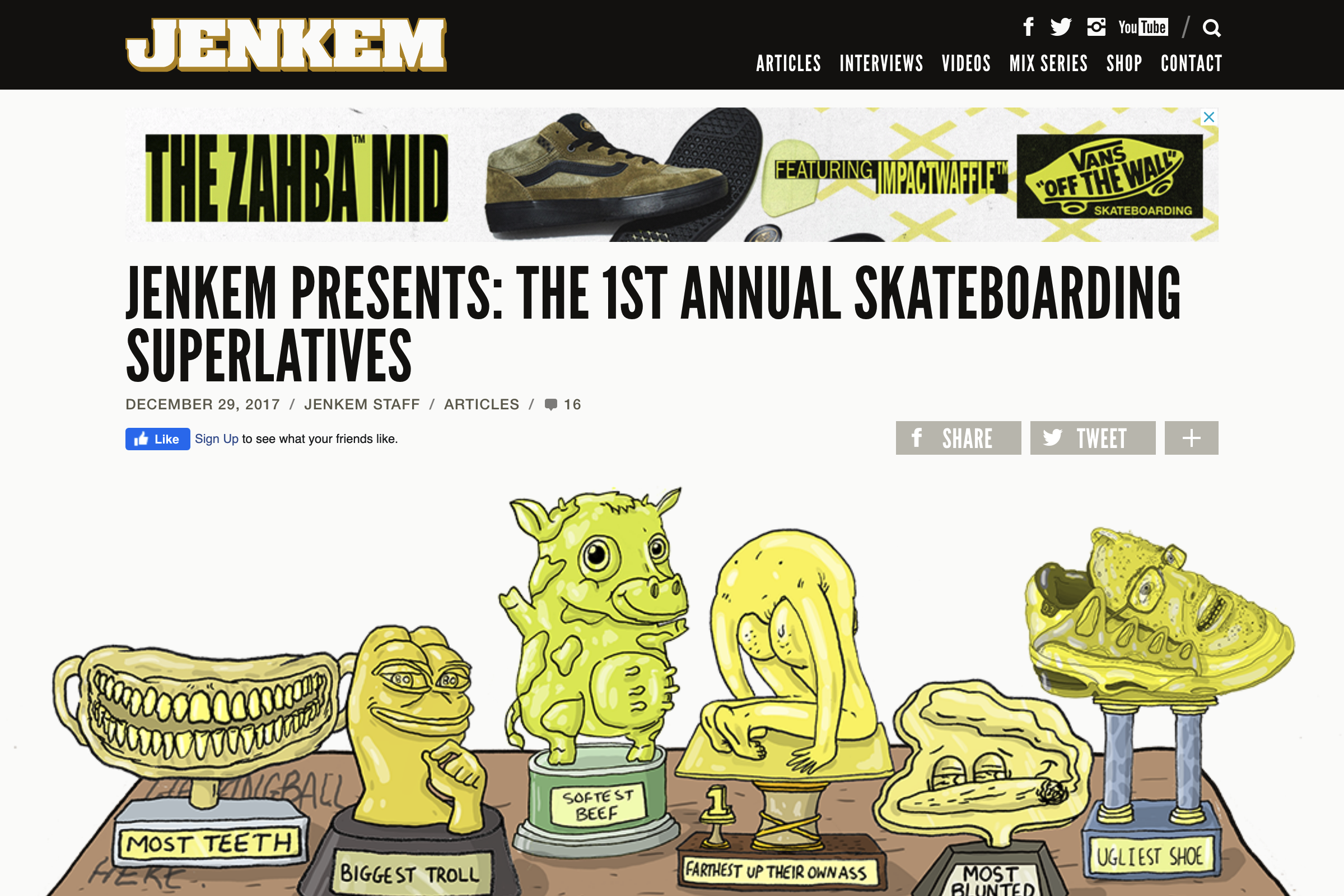}}}
    \caption{Examples of the \textit{Web Usage} visual scenario}
    \label{fig:web_usage_examples}
\end{figure*}


\begin{figure*}[h!]
    \centering
    \subfloat[\centering]{{\includegraphics[width=0.2\linewidth]{images/visual_examples/infographic/32692.jpeg}}}
    \hspace{8mm}
    \subfloat[\centering]{{\includegraphics[width=0.2\linewidth]{images/visual_examples/infographic/40436.jpeg}}}
    \hspace{8mm}
        \subfloat[\centering]{{\includegraphics[width=0.2\linewidth]{images/visual_examples/infographic/41181.jpeg}}}
        \hspace{8mm}
        \subfloat[\centering]{{\includegraphics[width=0.2\linewidth]{images/visual_examples/infographic/41656.jpeg}}}
    \caption{Examples of the \textit{Infographic} visual scenario}
    \label{fig:infographic_examples}
\end{figure*}


\begin{figure*}[h!]
    \centering
    \subfloat[\centering]{{\includegraphics[width=0.2\linewidth]{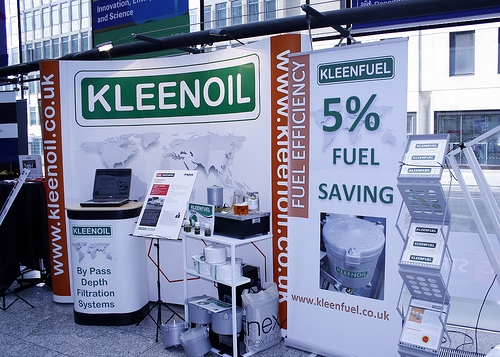}}}
    \hspace{8mm}
    \subfloat[\centering]{{\includegraphics[width=0.2\linewidth]{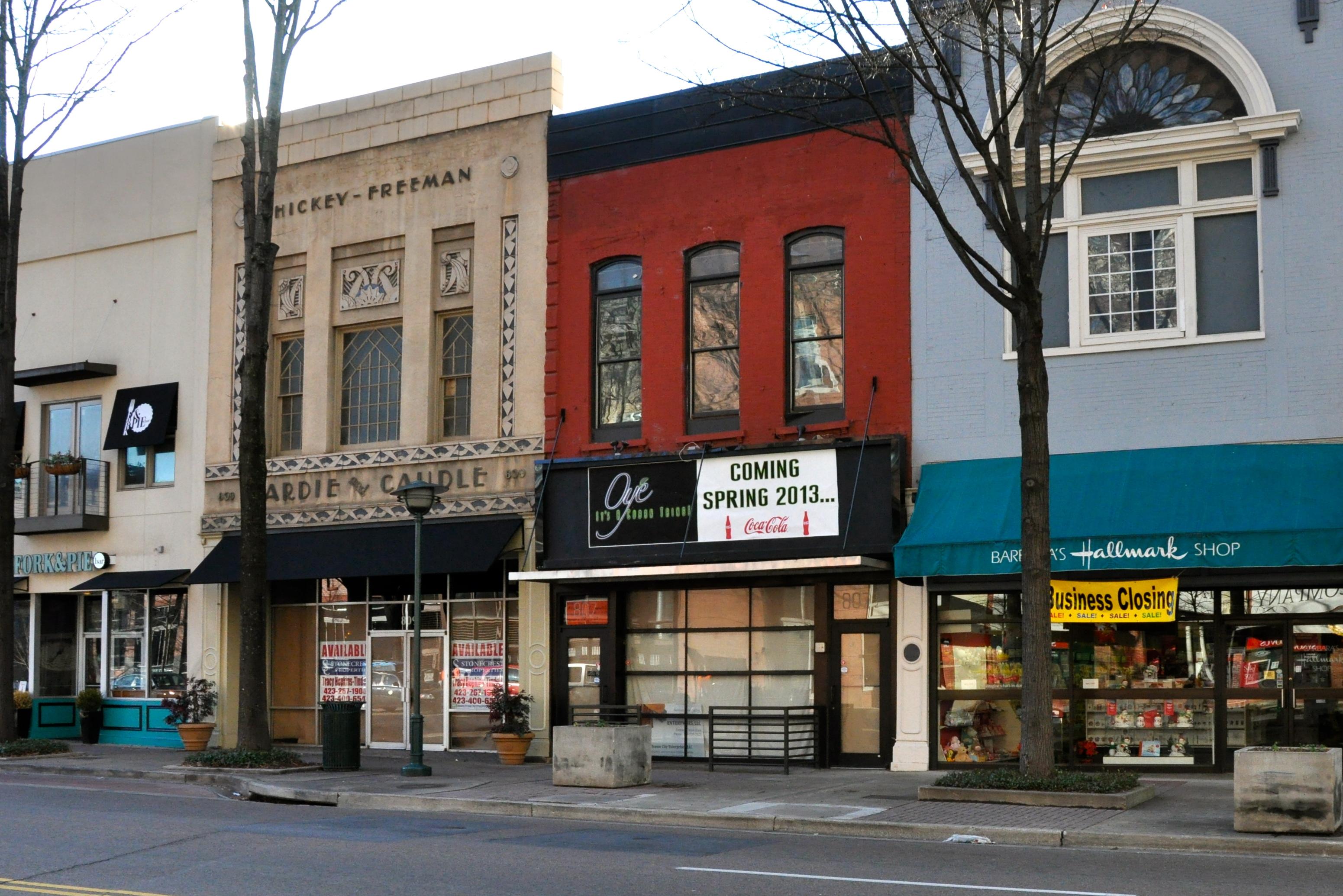}}}
    \hspace{8mm}
        \subfloat[\centering]{{\includegraphics[width=0.2\linewidth]{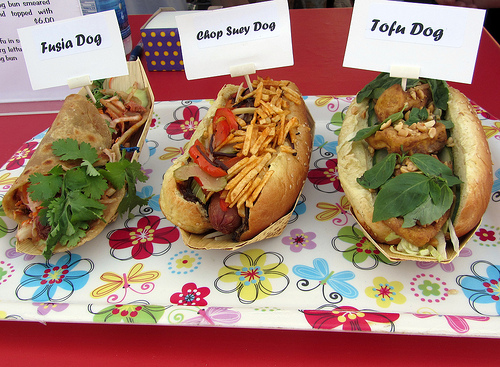}}}
        \hspace{8mm}
        \subfloat[\centering]{{\includegraphics[width=0.2\linewidth]{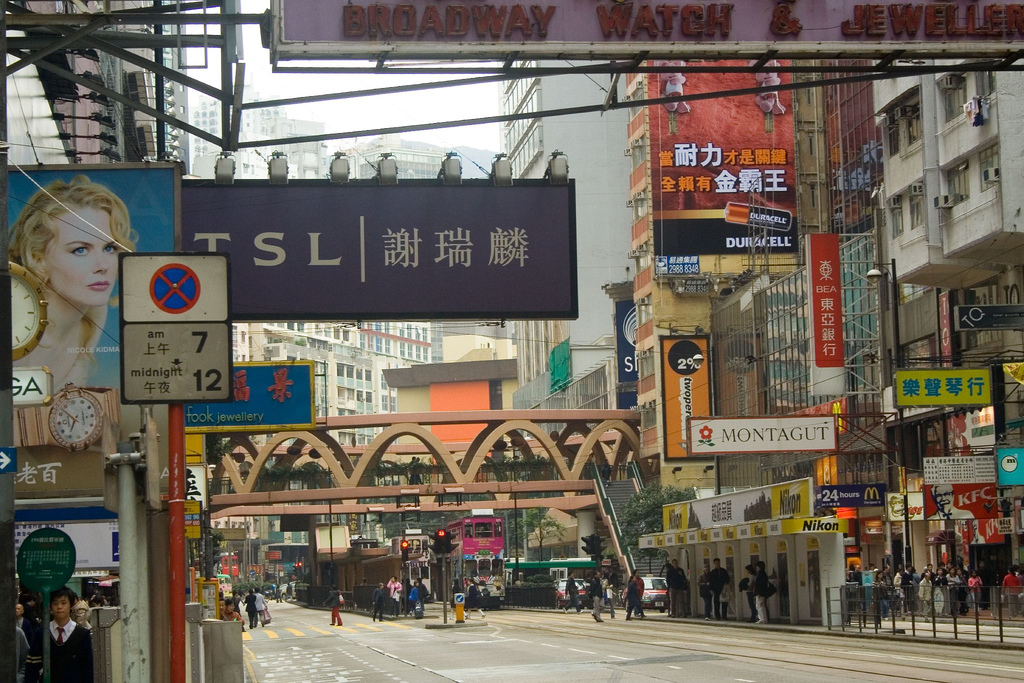}}}
    \caption{Examples of the \textit{Miscellaneous Natural Scenes} visual scenario}
    \label{fig:misc_examples}
\end{figure*}

\clearpage

\section{Human Annotation Screenshots}

\subsection{Human Performance Screenshot}
\label{app:human_responses_ss}

\begin{figure}[h!]
    \centering
    \includegraphics[scale=0.3]{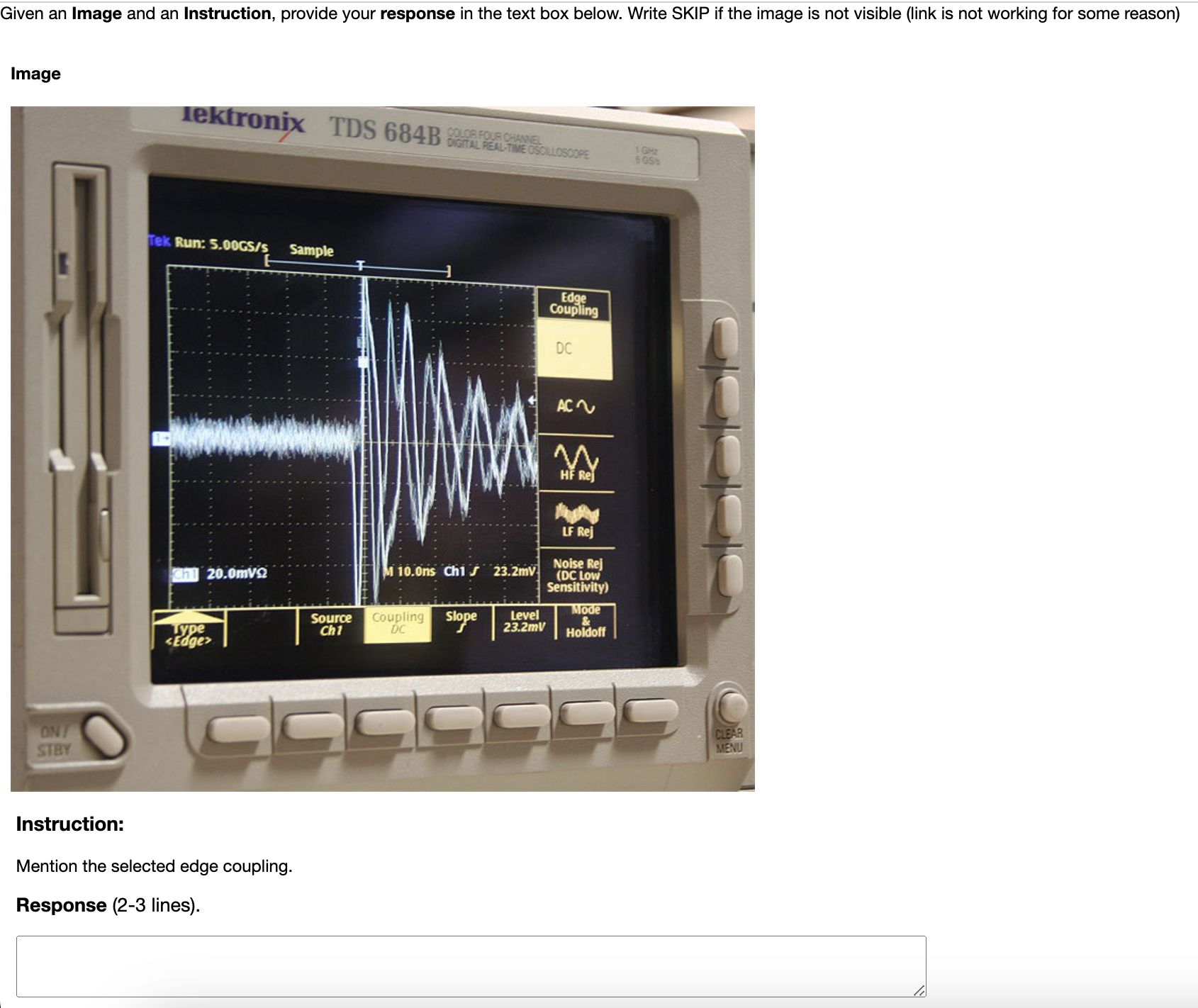}
    \caption{User interface of the human response collection.}
    \label{fig:mturk_human_response}
\end{figure}

We present the screenshot of the user interface used for acquiring human responses on the \name dataset in Figure \ref{fig:mturk_human_response}.

\subsection{Human Evaluation Screenshot}
\label{app:human_eval_ss}

\begin{figure}[h!]
    \centering
    \includegraphics[scale=0.35]{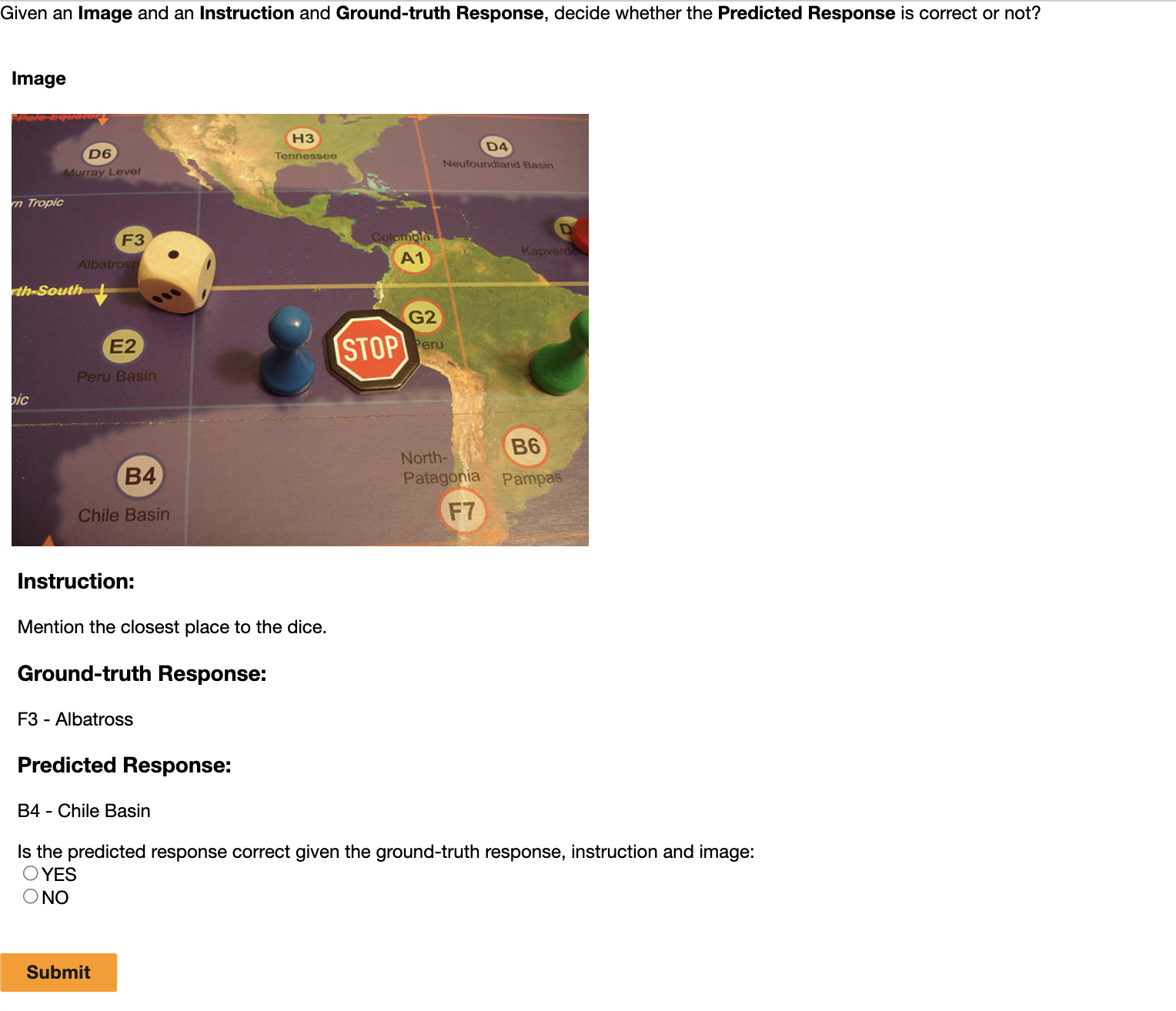}
    \caption{User interface of the human evaluation.}
    \label{fig:mturk_human_eval}
\end{figure}

We present the screenshot of the user interface used for human evaluation in Figure \ref{fig:mturk_human_eval}.

\clearpage

\section{Data Release}
\label{sec:data_release}
\name comprises 506 samples spanning eight visual categories (refer to Table \ref{tab:contextual_stats}). To facilitate model development, we will release a subset of 100 samples from the 506, as validation set, along with their reference responses, while keeping them hidden for the remaining 406 samples. We ensure that the distribution of validation samples closely mirrors the overall dataset distribution. To achieve this, we randomly select 30 samples from the `Miscellaneous Natural Scenes' category and 10 samples from the remaining categories, maintaining a proportional representation of each category in the validation samples, consistent with the overall benchmark. In this paper, all the results are reported on the entire dataset, unless stated otherwise.

\section{Few-Shot Setting}
\label{sec:few_shot}

\begin{figure}[h]
    \centering
    \includegraphics[width=0.6\linewidth]{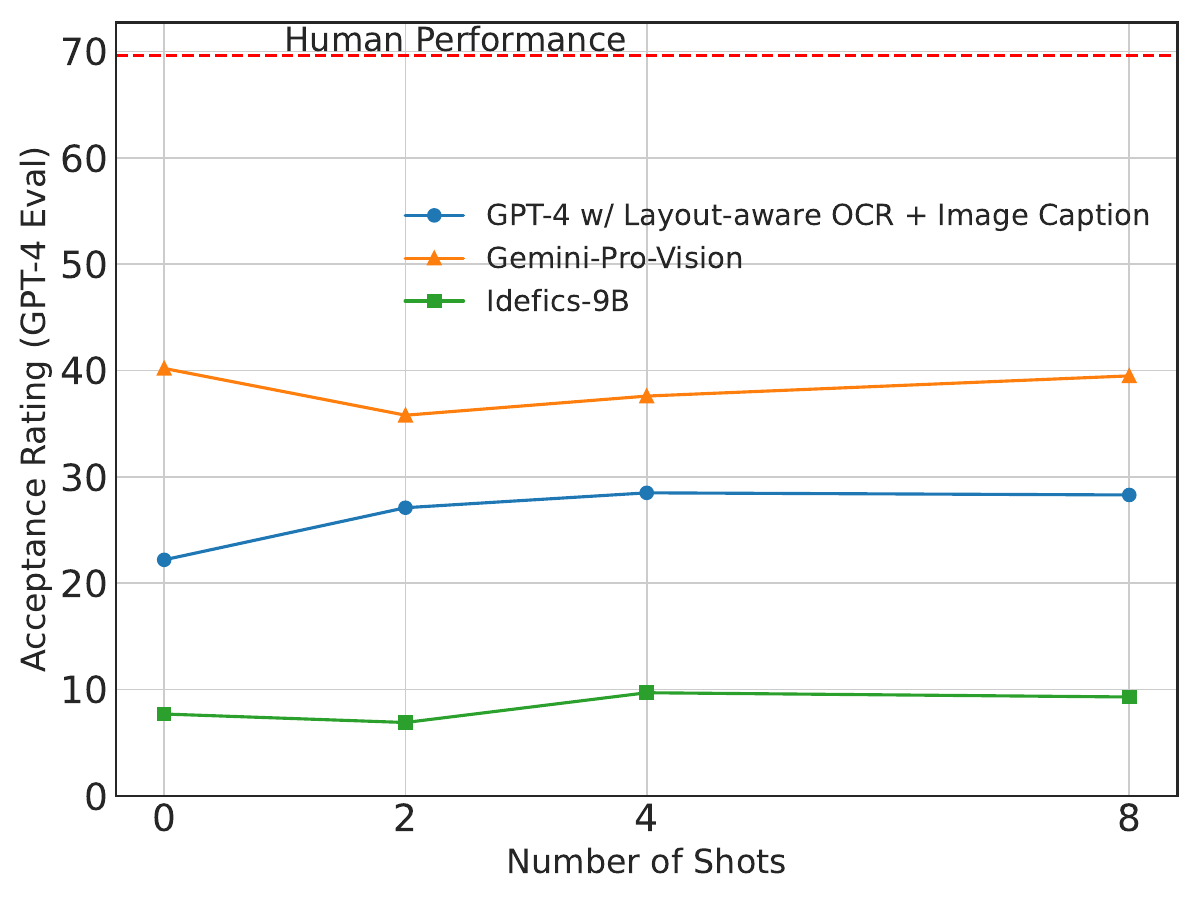}
    \caption{Few-shot performance on the \name dataset.}
    \label{fig:incontext}
\end{figure}

Here, we compare the performance of the foundation models on \name{} using GPT-4 evaluation with under the few-shot settings in Figure \ref{fig:incontext}. Specifically, we perform zero-shot, two-shot, four-shot, and eight-shot evaluation for augmented LLM (GPT-4 prompted w/ layout aware OCR and image caption), Gemini-Pro-Vision, and Idefics-9B. We select in-context examples at random from our dataset and evaluate the models on the remaining instances. 

In our experiments, we find that the performance of all the models does not change drastically with in-context examples. Specifically, we observe that Gemini-Pro-Vision response acceptance rating decreases by $5\%$ in the two-shot setting as compared to the zero-shot setting, and, increases monotonically from two-shot to eight-shots. In addition, we observe that the performance improvements stagnate for Idefics-9B after the four in-context examples. Recent studies highlight the instability and sensitivity of LMMs in few-shot settings \cite{li2023comprehensive}. For instance, a significant accuracy drop was observed in models like InstructBLIP in four-shot setting, especially in tasks requiring commonsense reasoning. Overall, we highlight that providing few-shot examples does not elicit context-sensitive text-rich visual reasoning in the foundation models.

\section{Augmented LLM Prompt}
\label{app:aug_llm_prompt}
In this section,  we discuss the design and elaborate on the prompts employed for the Augmented LLM approach (illustrated in Figure \ref{fig:ocr_prompt}, \ref{fig:la_ocr_prompt}, \ref{fig:la_ocr_ic_prompt}). We describe the three distinct prompt formats utilized, each differing in the extent of visual information presented. These formats encompass simple OCR of the image, OCR of the image arranged in the layout it appears in the image, and OCR presented in a layout format along with a comprehensive image caption. We prompt GPT4 with the above templates that does not take the image as input. However, the image is included in the illustration for reference purposes. 

\subsection{GPT-4 w/ OCR}
\begin{table*}[h!]
\fontsize{9.0pt}{\baselineskip}\selectfont
\linespread{0.9}\selectfont
\begin{mybody}
\includegraphics[scale=0.4]{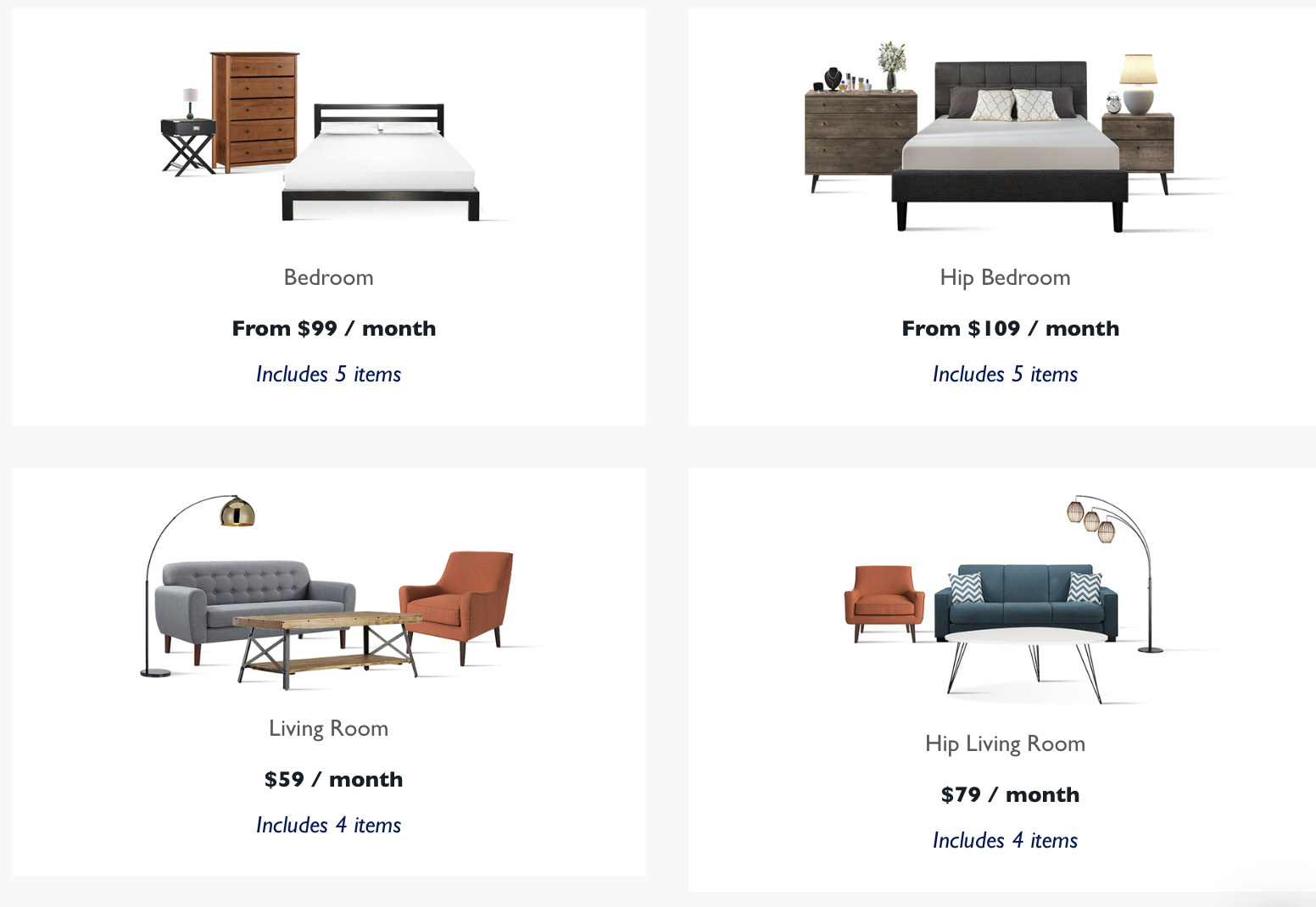}
\\
\textbf{Instruction:} Describe the most similar product between the living room and the hip living room.\\\\
\textbf{Reference Response:} 
The orangish-brown colored single-seating sofa is the most similar between the Living Room and the Hip Living Room.
\end{mybody}
\begin{mybody}
\textbf{Prompt:} \\ 
You are OCRGPT, an expert language model at responding to instructions posed for images. You have to respond to the instruction using the OCR Text of the image. More specifically, you will be given the following: \\

1. An instruction: This is a question, an imperative request, or something similar about the image that requires a response. \\

2. OCR Text: Text extracted from the image. \\

You have to respond with the Answer only. \\

NOW YOUR TURN:\\

Instruction : Provide the price of the upholstered dining set.\\

OCR Text: \\ Bedroom Hip Bedroom From \$99 / month From \$109 / month Includes 5 items Includes 5 items Living Room Hip Living Room. \$59 / month  \$79 / month Includes 4 items Includes 4 items \\\\
Answer:
\end{mybody}
\begin{mybody}
\textbf{GPT-4 w/ OCR Response:} \textbf{\red{Both the Living Room and the Hip Living Room include 4 items.}}
\end{mybody}
\captionof{figure}{Example prompt for Aug LLM with GPT4 w/ OCR provided without layout aware arrangement of it}
\label{fig:ocr_prompt}
\end{table*}

\clearpage

\subsection{GPT-4 w/ Layout-aware OCR}
\begin{table*}[th!]
\fontsize{9.0pt}{\baselineskip}\selectfont
\linespread{0.9}\selectfont
\begin{mybody}
\includegraphics[scale=0.4]{images/qualitative_examples/shopping/correct/screen-shot-2017-10-19-at-12-49-38-am.png}
\\
\textbf{Instruction:} Describe the most similar product between the living room and the hip living room.\\\\
\textbf{Reference Response:} 
The orangish-brown colored single-seating sofa is the most similar between the Living Room and the Hip Living Room.
\end{mybody}
\begin{mybody}
\textbf{Prompt:} \\ 
You are OCRGPT, an expert language model at responding to instructions posed for images. You have to respond to the instruction using the OCR Text of the image. More specifically, you will be given the following: \\

1. An instruction: This is a question, an imperative request, or something similar about the image that requires a response. \\

2. Layout Aware OCR Text: Text extracted from the image and arranged in the layout as it appears in the image, such that the relative position of among the pieces of text appearing in the image is maintained. \\

You have to respond with the Answer only. \\

NOW YOUR TURN: \\

Instruction : Provide the cost of hiring the machine type depicted in the image for nine months. \\

OCR TEXT:           \\
        Bedroom  \hspace{2cm}                 Hip Bedroom \\
        From \$99 / month \hspace{2cm}          From \$109 / month \\
         Includes 5 items  \hspace{2cm}         Includes 5 items \\
         Living Room    \hspace{2cm}            Hip Living Room. \\
         \$59 / month   \hspace{2cm}            \$79 / month \\
        Includes 4 items   \hspace{2cm}             Includes 4 items \\
Answer:
\end{mybody}
\begin{mybody}
\textbf{GPT-4 w/ Layout-aware OCR Response:} \textbf{\red{The most similar product between the living room and the hip living room is that both are available on a monthly basis, the living room costs \$59/month and the hip living room costs \$79/month, both including 4 items.}}
\end{mybody}
\captionof{figure}{Example prompt for Aug LLM with GPT-4 w/ Layout-aware OCR Response}
\label{fig:la_ocr_prompt}
\end{table*}

\clearpage
\subsection{GPT-4 w/ Layout-aware OCR + Caption Respons}
\begin{table*}[h!]
\fontsize{9.0pt}{\baselineskip}\selectfont
\linespread{0.9}\selectfont
\begin{mybody}
\includegraphics[scale=0.27]{images/qualitative_examples/shopping/correct/screen-shot-2017-10-19-at-12-49-38-am.png}
\\
\textbf{Instruction:} Describe the most similar product between the living room and the hip living room.\\\\
\textbf{Reference Response:} 
The orangish-brown colored single-seating sofa is the most similar between the Living Room and the Hip Living Room.
\end{mybody}
\begin{mybody}
\textbf{Prompt:} \\ 
You are OCRImageCaptionGPT, an expert language model at responding to instructions posed for images. You have to respond to the instruction using the OCR Text of the image and the Image Caption. More specifically, you will be given the following: \\

1. An instruction: This is a question, an imperative request, or something similar about the image that requires a response. \\

2. Layout Aware OCR Text: Text extracted from the image and arranged in the layout as it appears in the image, such that the relative position of among the pieces of text appearing in the image is maintained. \\

3. Image Caption: Natural language description of the contents of the image, like objects, text, etc, and their interaction with one another. \\

You have to respond with the Answer only. 
NOW YOUR TURN:\\

Instruction : Provide the cost of hiring the machine type depicted in the image for nine months. \\

OCR Text:           \\
        Bedroom  \hspace{2cm}                 Hip Bedroom \\
        From \$99 / month \hspace{2cm}          From \$109 / month \\
         Includes 5 items  \hspace{2cm}         Includes 5 items \\
         Living Room    \hspace{2cm}            Hip Living Room. \\
         \$59 / month   \hspace{2cm}            \$79 / month \\
        Includes 4 items   \hspace{2cm}             Includes 4 items \\

Image Caption: 
The image is an advertisement for a furniture store, showcasing a variety of furniture items. The items are arranged in a grid-like pattern, with each item occupying its own square. The furniture includes a couch, a dining table, a chair, a dresser, a bed, and a cabinet. The colors of the furniture range from brown to beige, with some items featuring black and white accents. The text in the image indicates that the furniture is available for purchase at prices ranging from 295 to 1945. The text also mentions that the furniture is of ""LUXURY FURNITURE DESIGN"". The overall layout and design of the advertisement suggest a focus on high-quality, stylish furniture. \\\\
Answer:
\end{mybody}
\begin{mybody}
\textbf{GPT-4 w/ Layout-aware OCR + Caption Response:} \textbf{\red{The most similar product between the Living Room and the Hip Living Room is in terms of the quantity of items. Both living room options include 4 items each - a couch, a coffee table, and a side table. The relative positioning and type of furniture also seem to be similar, although the styles and colors vary.}}
\end{mybody}
\captionof{figure}{Example prompt for Aug LLM with GPT-4 w/ Layout-aware OCR + Caption Response.}
\label{fig:la_ocr_ic_prompt}
\end{table*}
\clearpage

\section{GPT-4 Evaluation Prompt}
\label{app:gpt_4_eval_prompt}
In this section, we provide an illustration of a GPT4 prompt used to assess both model-generated and human responses. Figure \ref{fig:gpt4_prompt} displays an instance within the Shopping category, featuring a reference response generated by the Gemini-Pro-Vision model. It's important to observe that the prompt does not include any information about the predicting model.

\begin{table*}[h!]
\fontsize{9.0pt}{\baselineskip}\selectfont
\linespread{0.9}\selectfont
\begin{mybody}
\includegraphics[scale=0.25]{images/qualitative_examples/shopping/correct/screen-shot-2017-10-19-at-12-49-38-am.png}
\\
\textbf{Instruction:} Describe the most similar product between the living room and the hip living room.\\\\
\textbf{Reference Response:} 
The orangish-brown colored single-seating sofa is the most similar between the Living Room and the Hip Living Room.
\end{mybody}
\begin{mybody}
\textbf{Prompt:} \\ 
You are ImageTaskEvaluatorGPT, an expert language model at judging whether or not a response adequately addresses an instruction in the context of an image. More specifically, you will be given the following:\\

1. An instruction: This is a question, an imperative request, or something similar about the image which requires a response.\\
2. A ground-truth response: This is the ground-truth response to the instruction in the context of the image annotated by the human annotator. \\
3. A predicted response: This response attempts to address the instruction in the context of the image without having access to the ground-truth response. \\

Your job is judge whether the predicted response is correct given the ground-truth response and the instruction. \\
 
Some things to remember: \\

- Even though you are just a language model, the instructions mostly require an objective answer i.e., the ground-truth response and instruction should be sufficient for you to judge the correctness of the predicted response. You do not need to have access to the complete image description. \\
- You are capable of judging response quality, accounting for important factors like correctness, relevance, fluency, specificity, etc.  \\
- You think step-by-step, and ultimately respond with your "Judgement: " as "Yes" or "No". Here, "Yes" implies that the predicted response is correct according to you, and "No" implies that the predicted response is not correct. \\
- Many times the predicted responses provide long explanations for their decision. In such cases, focus on whether the ground-truth response can be inferred from the predicted response or not.  \\

Instruction: Describe the most similar product between the living room and the hip living room. \\
Ground-truth Response: The orangish-brown colored single-seating sofa is the most similar between the Living Room and the Hip Living Room.  \\
Predicted Response: The most similar product between the living room and the hip living room is the sofa. Both sofas are blue and have a similar shape. \\
\end{mybody}
\begin{mybody}
\textbf{Response:} No
\end{mybody}
\captionof{figure}{Example prompt for GPT4 evaluation. Here, the predicted response is taken from Gemini Pro-Vision}
\label{fig:gpt4_prompt}
\end{table*}
\clearpage

\section{Additional Fine-grained Evaluation}
\label{sec:additional_fine_grained}

\begin{figure*}[h!]
    \centering
    \subfloat[\centering \label{fig:ex_math}  Performance on different types of tasks.]{{\includegraphics[width=0.45\linewidth]{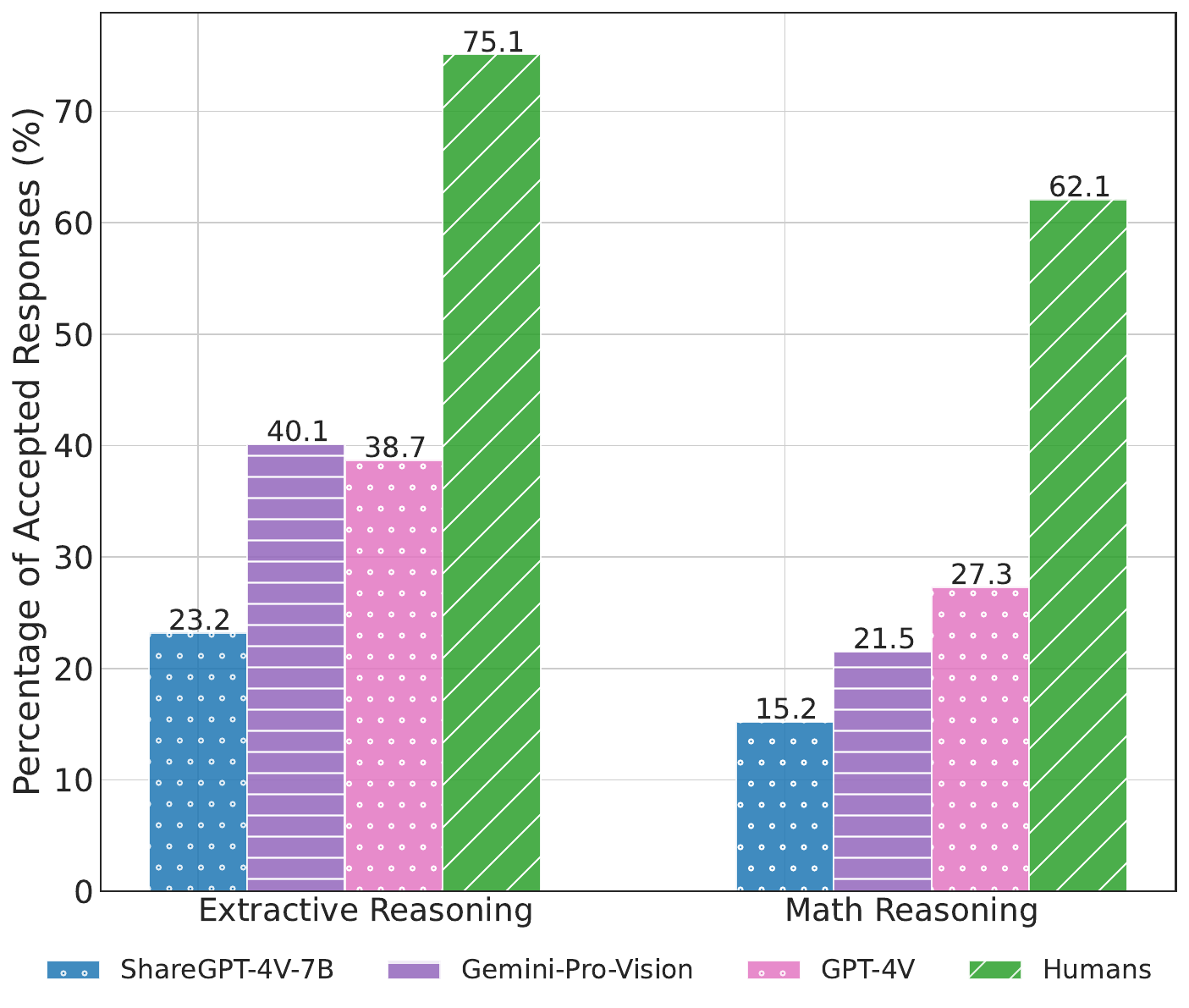}}}
    \hspace{8mm}
    \subfloat[\centering \label{fig:scenes} Performance on natural and digital scenes.]{{\includegraphics[width=0.45\linewidth]{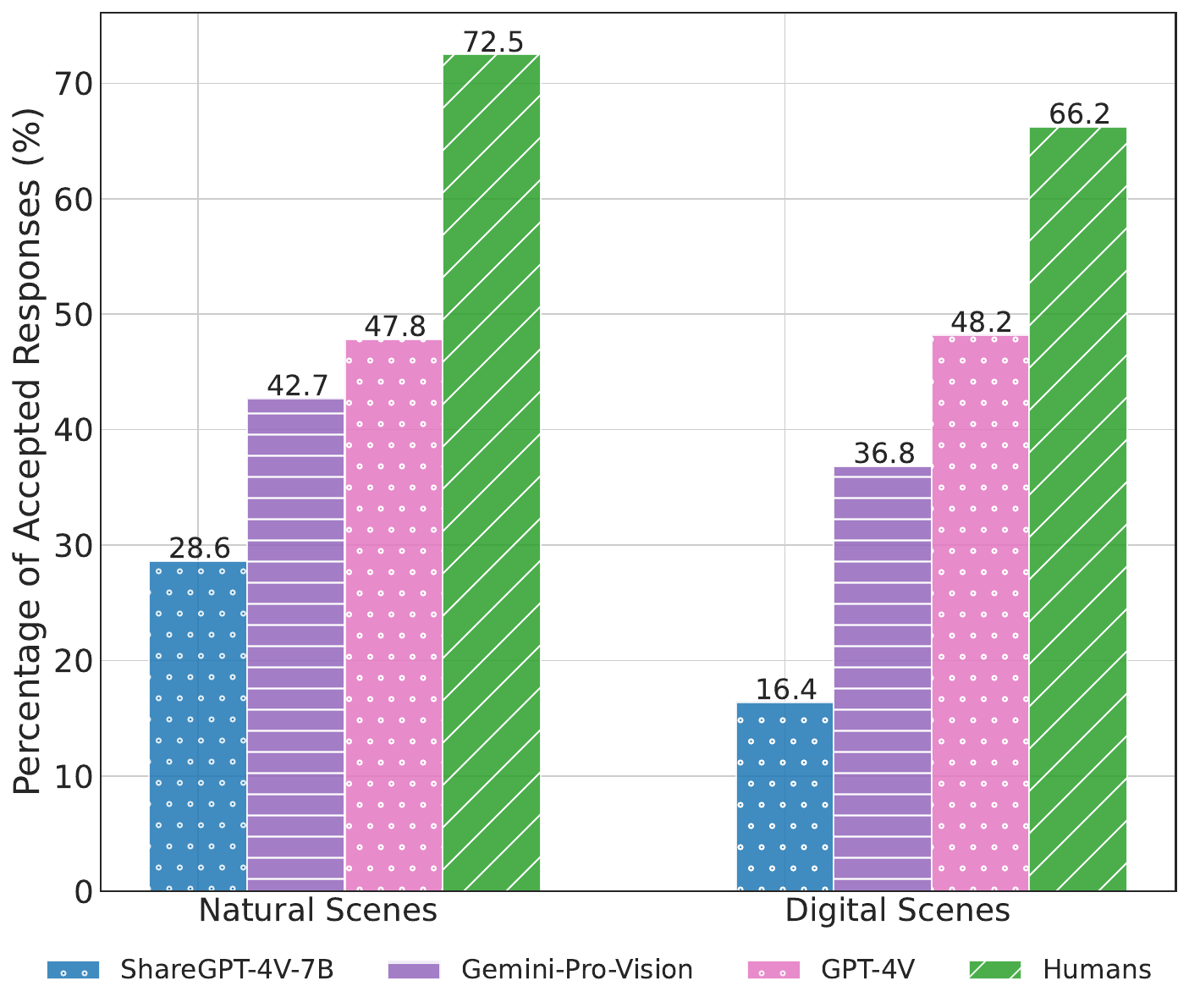}}}
    \caption{Additional fine-grained evaluation results.}
    \label{fig:visual_analysis}
\end{figure*}

\subsection{Types of Tasks}
\label{app:task}
We compare the performance of the foundation models with varying types of tasks in Figure \ref{fig:ex_math} using GPT-4 evaluation. Specifically, we assess the quality of the responses when the instructions require the models to extract text or visual elements in the image (e.g., \textit{List the exercises where the corresponding illustration showcases a single movement.}). There are $285$ such instances in the \name dataset. While these tasks require complex perception and reasoning abilities, they do not require additional operations on top of the information already presented in the image explicitly. We observe that the humans achieve $75.1\%$ on such instructions while the proprietary models GPT-4V and Gemini-Pro-Vision achieve $38.7\%$ and $40.1\%$, respectively. This indicates that humans are very good at identify the key information that needs to be extracted to respond to the instructions. 

In addition, we assess the responses when the instructions require the models to go beyond information extraction, and perform math reasoning for the instruction (e.g., \textit{What is the total price for the two cars listed here?}). There are $66$ instances in the \name dataset. We find that humans achieve $62.1\%$ on such tasks while the proprietary models GPT-4V achieve $27.3\%$, again highlighting at the large gap in their math reasoning. 

\subsection{Visual Scenes}

We compare the performance of the foundation models with varying visual scenes (e.g., natural scenes and digital scenes) in Figure \ref{fig:scenes}. Majorly, shopping, navigation, and misc. natural scenes constitute natural scenes, and web usage, mobile usage, abstract, infographics and time reading constitute digital scenes. We find that humans achieve the highest performance in both the visual scenes i.e., $72.5\%$ and $66.2\%$ on natural scenes and digital scenes, respectively. In addition, we observe that GPT-4V achieve $47.8\%$ and $48.2\%$ on natural and digital scenes, respectively. Interestingly, we find that Gemini-Pro-Vision and ShareGPT-4V-7B achieve higher performance on the natural scenes than the digital scenes. It indicates these models may not have seen many examples with digital scenes during their pretraining. Thus, our \name dataset highlights the gaps in the training data of the modern LMMs.

\clearpage

\section{Synthetically Scaling Data}
\label{app:data_scaling}
We develop an synthetic data generation pipeline that could be useful to create samples that more likely require context-sensitive text-rich visual reasoning. We use the existing OCR filtering strategy to obtain 200 candidate images belonging to Misc. Natural scenes category. Then, we use in-context learning capabilities of GPT-4V and prompt it to generate instruction response pairs (as shown in Fig \ref{fig:data_scaling}. We observed that use of negative demonstration within in-context examples, as shown in previous studies \cite{wang2022super}, along with reasons was critical for GPT4V to understand the nuanced difference between text only, visual element only and context-sensitive text-rich visual reasoning instructions. After generating the examples, we evaluated representative models from the different model categories: open-source LMMs (LLaVA, ShareGPT), closed-source LMMs (GPT-4V), and Augmented LLM (GPT-4 w/ Layout-aware OCR + Caption), as shown in Table \ref{tab:data_scaling}.

\begin{table*}[h]
\centering
\small
\setlength{\tabcolsep}{1mm}
\resizebox{0.7\linewidth}{!}{
\begin{tabular}{ccc}
\hline
 \textbf{Model} & \textbf{Synthetic examples} & \textbf{\name examples} \\\hline
LLaVa-v1.5-13B & 39.8 & 29.7 \\
ShareGPT4v-7B & 44.9 & 37.7 \\
GPT-4 w/ Layout-aware OCR & 51.1 & 27.3  \\
GPT4V  & 68.6 & 48.0\\
\hline
\end{tabular}
}
\caption{\small{Comparison of model performance (accuracy in \%) using GPT4 evaluation of synthetically generated samples belonging to Misc. Natural Scenes category (200 samples) and human annotated samples (\name) belonging to Misc. Natural Scences Category (156 samples).}}
\label{tab:data_scaling}
\end{table*}

We observe that the accuracy of models on synthetic samples is greater than the accuracy of human-made ConTextual examples. This is expected because it is difficult for the model to understand context-sensitive in the first place, and asking it to make such tasks would be difficult. Despite this, it can create good instructions because the performance gap of open-source LMMs is relatively small. The only exceptions are Aug LLM and GPT4V, which both use GPT models. Due to this, the GPT4 evaluator may favor their responses to the instruction generated by a GPT model. This demonstrates that we can scale these instructions by filtering good candidates for data generation and nuanced prompt engineering.

\begin{table*}[th!]
\fontsize{9.0pt}{\baselineskip}\selectfont
\linespread{0.9}\selectfont
\begin{mybody}
\centering
\includegraphics[scale=0.18]{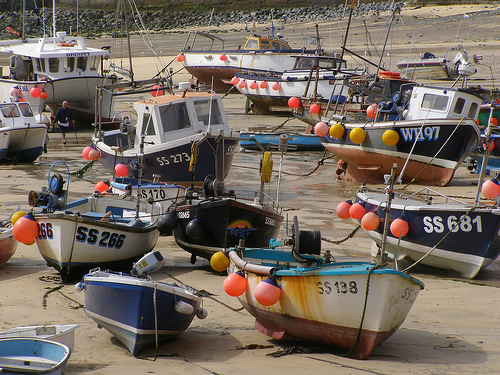}
\hspace{3em}
\includegraphics[scale=0.18]{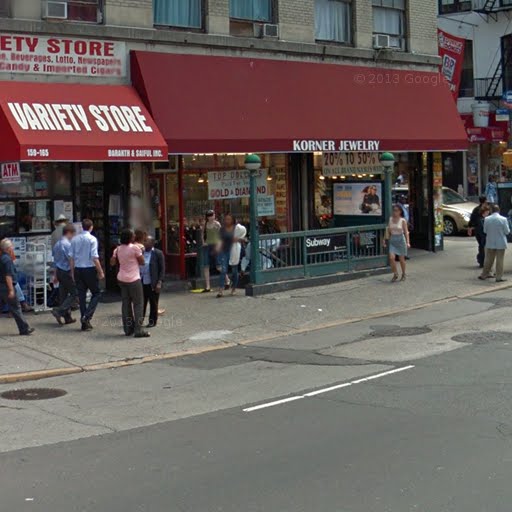}
\hspace{3em}
\includegraphics[scale=0.18]{images/uploaded_image_3.jpg}
\end{mybody}
\begin{mybody}
\textbf{Prompt:} \\ 
You are InstructionResponseGPT4Vision, an expert in understanding images and generating appropriate instruction-response pairs based on exact point Text-Vision Context. The model has access to two images and their corresponding Bad Instruction 1, Bad Instruction 1 reason, Bad Instruction 2, Bad Instruction 2 reason, Good Instruction, Good Instruction response and Good Instruction reason. You need to generate a Good Instruction-Response pair for the third image, that follows the constraints of the Good Instruction. You must always follow the next five guidelines strictly.\\
1. Instruction must be specific about the details in the image and not be vague or open ended\\
2. Instructions must always be complex, creative and use different imperative and task oriented main verbs.\\
3. Note text is not a visual element, a visual element is an object and its attributes / actions/ relationships.\\
4. An instruction cannot be based on just text as it can be answered using OCR. An instruction cannot be based on just visual elements as it can be answered by vision models. It must always require joint text and visual element reasoning.\\
5. The response should contain information present in the image, no external or general knowledge can be used or assumptions made.\\\\
Input Examples:\\
Uploaded Image 1:\\
Bad Instruction 1: Identify the color of the round items on the boats. \\
Reason: Bad instruction, only requires to color attributes of visual element, here round items, from the image. Does not use joint-text-vision-context to frame the instruction. \\
Bad Instruction 2: Count the number of red round items if the serial number is SS681. \\
Reason: Bad instruction, has a conditional  “if” in the instruction, conditionals cannot appear in the instruction. \\
Good Instruction: Get the number of the boat with three yellow and one red round items hanging from it. \\
Response: WH97 \\
Reason: Good instruction, requires first identifying which boat has three yellow and one round red item and then requires to extract its serial number WH97. \\\\
Uploaded Image 2: \\
Bad Instruction 1: Which stores are visible in this image?\\
Reason: Bad instruction, as one can obtain the answers using OCR and we do not want that. Does not use joint-text-vision-context to frame the instruction. \\
Bad Instruction 2: Is the signage of the stores shown in the image of the same color? \\
Reason: Bad instruction, we only need visual element extraction to respond to the question. Does not use joint-text-vision-context to frame the instruction. \\
Good Instruction:  Name the shop whose main door directly leads to the subway's entrance. \\
Response: Variety Store \\
Reason: Good Instruction, it requires to first localize physical elements of the stores, like the main door and see which one is before the subway because that will lead to the subways entrace. After identfying the correct store, it needs to extract the name of that store. This uses joint-text-vision context to frame the instruction.\\\\
Now your turn: \\
Uploaded Image 3: \\
Good Instruction: \\
Response: 
\end{mybody}
\begin{mybody}
 \textbf{Good Instruction}: Determine the message conveyed by the combination of the STOP sign and the movement implied by the blurred background.\\
\textbf{Response}: The image conveys the message of a railroad crossing where vehicles are required to stop for passing trains, indicated by the combination of a STOP sign and the motion blur of a train in the background.
\end{mybody}
\captionof{figure}{Example prompt for synthetically generating context-sensitive instructions using GPT4V, with images from left-to-right: Uploaded Image 1, Uploaded Image 2, Uploaded Image 3.}
\label{fig:data_scaling}
\end{table*}

\clearpage

\section{Detailed Qualitative Analysis}
\label{app:qualitative_examples}
In this section, we qualitatively analyze the responses of the different visual categories. In the caption annotations, we follow the following color notations. \green{green}: correct response, \red{red}: incorrect response, and \purple{purple}: intuition behind incorrect response.

\subsection{Time}
This section provides qualitative analysis of the \textit{Time} visual scenario and our findings across the models, as shown in Figures \ref{fig:time_correct_1}, \ref{fig:time_correct_2}, \ref{fig:time_wrong_1}, \ref{fig:time_wrong_2}.

\begin{table*}[th!]
\fontsize{9.0pt}{\baselineskip}\selectfont
\linespread{0.9}\selectfont
\begin{mybody}
\includegraphics[scale=0.4]{images/qualitative_examples/time/correct/original-2563140-4.png}
\\
\textbf{Instruction:} Identify the object on the dial that shows the time 10:10. \\\\
\textbf{Reference Response:} The clock at 10:10 features an octopus on its dial. 
\end{mybody}
\begin{mybody}
\textbf{Human Response:} The only clock with an object on the dial at 10:10 is the white one with a \textbf{\green{tall purple octopus}} on it. 
\end{mybody}
\begin{mybody}
\textbf{GPT-4V Response:} The clock with the image of \textbf{\green{an octopus}} on the face is showing the time 10:10.
\end{mybody}
\begin{mybody}
\textbf{Gemini Pro-Vision Response:} \textbf{\green{An Octopus}} 
\end{mybody}
\begin{mybody}
\textbf{LLaVA-Next-34B:} The object on the dial that shows the time 10:10 is the \textbf{\red{hour hand and the minute hand}}.\\
\textbf{\purple{Lack of context-sensitive text-rich visual reasoning}}
\end{mybody}
\begin{mybody}
\textbf{ShareGPT-4V-7B Response:} The object on the dial that shows the time 10:10 is an \textbf{\red{apple}}. \\
\textbf{\purple{Lack of context-sensitive text-rich visual reasoning}}
\end{mybody}
\begin{mybody}
\textbf{LLaVA-1.5-13B Response:} The object on the dial that shows the time 10:10 is a \textbf{\red{clock}}. \\
\textbf{\purple{Lack of context-sensitive text-rich visual reasoning}}
\end{mybody}
\begin{mybody}
\textbf{GPT-4 w/ Layout-aware OCR + Caption Response:} \textbf{\red{The instruction does not provide enough specific information to identify the object on the dial that shows the time 10:10.}} \\
\textbf{\purple{Lack of context-sensitive text-rich visual reasoning}}
\end{mybody}
\captionof{figure}{In this task, \textit{GPT-4V} and \textit{Gemini Pro-Vision} correctly respond to the instruction. This can be attributed to their ability for visual perception, text recognition, and context-sensitive text-rich visual reasoning. Open source LMMs like \textit{LLaVA-Next-34B}, \textit{ShareGPT4V} and \textit{LLaVA-1.5-13B} and \textit{GPT-4 w/ Layout-aware OCR + Caption Response} (Augmented LLM approach) predict the wrong response.}
\label{fig:time_correct_1}
\end{table*}


\begin{table*}[th!]
\fontsize{9.0pt}{\baselineskip}\selectfont
\linespread{0.9}\selectfont
\begin{mybody}
\includegraphics[scale=0.4]{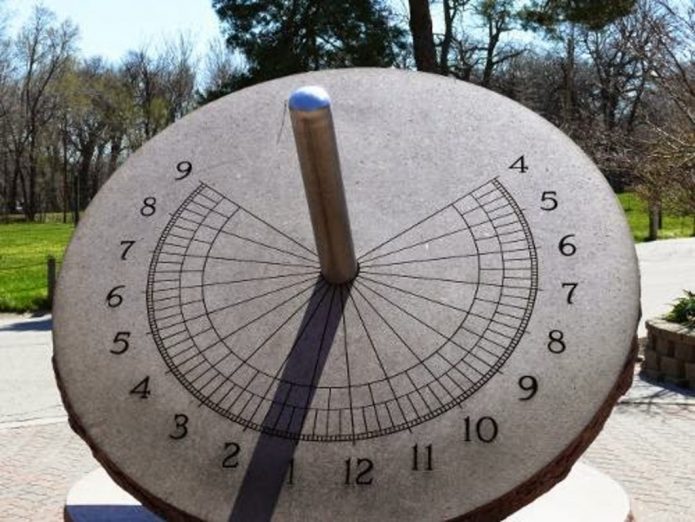}
\\
\textbf{Instruction:} What time is the sun dial indicating? \\\\
\textbf{Reference Response:} The time shown sun dial is between 1pm and 2pm.
\end{mybody}
\begin{mybody}
\textbf{Human Response:} The sun dial Depicts \textbf{\green{approximately 1:30}}.
\end{mybody}
\begin{mybody}
\textbf{GPT-4V Response:} The sundial in the image is designed to tell time by casting a shadow on its face with the gnomon, which is the part sticking out that catches the sunlight. The shadow appears to be falling between the 1 and 2 on the right side of the dial. \textbf{\green{Given the position of the shadow, it suggests that the time is approximately 1:30}}.
\end{mybody}
\begin{mybody}
\textbf{Gemini Pro-Vision Response:} The sundial is indicating that it is \textbf{\green{between 1 and 2 o' clock.}} 
\end{mybody}
\begin{mybody}
\textbf{LLaVA-Next-34B:} The sundial in the image indicates that it is approximately \textbf{\red{12:30}}. The shadow of the gnomon (the central rod) is pointing towards the number 12, which represents noon, and the shadow is about halfway between the numbers 12 and 1, suggesting that it is about 30 minutes past noon. Sundials are used to tell time by the position of the shadow cast by the gnomon, which is aligned with the Earth's axis and the position of the sun. \\
\textbf{\purple{Reasoning correct but needs accurate fine-grained perception}}
\end{mybody}
\begin{mybody}
\textbf{ShareGPT-4V-7B Response:} The sun dial is indicating \textbf{\red{12:20}}. \\
\textbf{\purple{Wrong response}}
\end{mybody}
\begin{mybody}
\textbf{LLaVA-1.5-13B Response:} The sun dial is indicating that it is \textbf{\green{1:20 pm}}.
\end{mybody}
\begin{mybody}
\textbf{GPT-4 w/ Layout-aware OCR + Caption Response:} \textbf{\red{The image and the OCR text do not provide enough information to determine the exact time the sundial is indicating.}}\\
\textbf{\purple{Lack of context-sensitive text-rich visual reasoning}}
\end{mybody}
\captionof{figure}{In this task, closed-source LMMs \textit{GPT-4V} and \textit{Gemini Pro-Vision} correctly respond to the instruction. This can be attributed to their ability for visual perception, text recognition, and context-sensitive text-rich visual reasoning. \textit{GPT-4V} is also able to explain the reasoning behind its response correctly. Amongst the open source LMMs, \textit{LLaVA-1.5} correctly responds to the instruction, suggesting its ability to perform context-sensitive text-rich visual reasoning in this case. Notably, \textit{LLaVA-Next-34B} gets the answer wrong but the reasoning correctly justifies the incorrect answer, thereby highlighting the need for more accurate fine-grained perception.. On the other, \textit{ShareGPT4V} and \textit{GPT-4 w/ Layout-aware OCR + Caption Response} (Augmented LLM approach) predict the wrong response.}
\label{fig:time_correct_2}
\end{table*}


\begin{table*}[th!]
\fontsize{9.0pt}{\baselineskip}\selectfont
\linespread{0.9}\selectfont
\begin{mybody}
\includegraphics[scale=0.6]{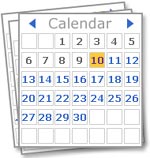}
\\
\textbf{Instruction:} Explain the idea behind using a color scheme for the different numbers in the calendar. \\\\
\textbf{Reference Response:} The calendar employs three distinct colors: grey for days prior to the current date, yellow highlights for the current day, and blue for the remaining days in the month.
\end{mybody}
\begin{mybody}
\textbf{Human Response:} \textbf{\green{The grey numbers are the dates that already passed. The yellow color marked today's date. The blue color marked tomorrow and future dates.}}.
\end{mybody}
\begin{mybody}
\textbf{GPT-4V Response:} 
\textbf{\green{The use of a color scheme in a calendar can help users quickly distinguish between different types of days or events.}} For example: \\
Weekdays vs. Weekend Days: Standard calendars often utilize color to differentiate between weekdays and weekends, with Saturdays and Sundays sometimes colored to stand out from Monday through Friday. \\
\textbf{\green{Current Day: The current date might be highlighted in a unique color, as shown in the image where the number 10 is in orange. This helps users quickly identify the present day without searching. \\
Past, Present, and Future Distinguishing: Some calendars use color gradients or different tones to differentiate between past, present, and future dates.}}\\
\textbf{\purple{Partly correct response but not grounded to this image.}}
\end{mybody}
\begin{mybody}
\textbf{Gemini Pro-Vision Response:} \textbf{\green{The color scheme used in the calendar is to highlight the current day. The current day is highlighted in yellow, making it easy to identify.}} \\
\textbf{\purple{Parially correct response}}
\end{mybody}
\begin{mybody}
\textbf{LLaVA-Next-34B:} The color scheme used for the numbers in the calendar likely serves to highlight specific dates that are important or relevant to the user. For example, \textbf{\green{the yellow number \"10\" is highlighted, which could indicate that it is a date of particular significance, such as a birthday, anniversary, or an event that the user wants to easily identify. The use of color in this way helps to quickly draw attention to the important dates, making it easier for the user to remember or plan around them.}} \\
\textbf{\purple{Partially correct response}}
\end{mybody}
\begin{mybody}
\textbf{ShareGPT-4V-7B Response:} \textbf{\red{The use of a color scheme for the different numbers in the calendar serves a few purposes. First, it adds visual interest and aesthetic appeal to the calendar, making it more visually engaging and easier to read. The color-coded numbers can also help users quickly identify specific dates or events, as the distinct colors can be more noticeable and memorable than the traditional black or gray numbers.}} \\
\textbf{\purple{Response not grounded to this image}}
\end{mybody}
\begin{mybody}
\textbf{LLaVA-1.5-13B Response:} The idea behind using a color scheme for the different numbers in the calendar is to make it easier for users to identify and distinguish between specific dates or events. \textbf{\green{In the image, the number 10 is highlighted in yellow, which draws attention to that particular day.}}\\
\textbf{\purple{Partially correct response}}

\end{mybody}
\begin{mybody}
\textbf{GPT-4 w/ Layout-aware OCR + Caption Response:} \textbf{\red{The idea behind using a color scheme for the different numbers in the calendar is to create a clear distinction between the types of information presented. In this case, the blue color is used for the days of the week and black for the dates. This way, users can easily differentiate between these two types of information, enhancing readability and understanding.}} \\
\textbf{\purple{Response not grounded to this image}}
\end{mybody}
\captionof{figure}{In this task, all the models fail to ground their response to the image, with the best performing ones being \textit{GPT4V, Gemini-Vision-Pro, LLaVA-Next-34B \& LLaVA-1.5-13B} and \textit{ShareGPT4V-7B \& GPT-4 w/ Layout-aware OCR + Caption} producing a generic response}
\label{fig:time_wrong_1}
\end{table*}


\begin{table*}[th!]
\fontsize{9.0pt}{\baselineskip}\selectfont
\linespread{0.9}\selectfont
\begin{mybody}
\includegraphics[scale=0.4]{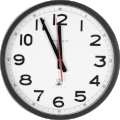}
\\
\textbf{Instruction:} How much time remains until the hour and minute hands align? \\\\
\textbf{Reference Response:} The hour and the minute hand will align at 12 o'clock, 5 minutes from the depicted time.
\end{mybody}
\begin{mybody}
\textbf{Human Response:} \textbf{\green{5 minutes}}.
\end{mybody}
\begin{mybody}
\textbf{GPT-4V Response:}  
To answer this, we'll have to calculate the relative speeds of the hour and minute hands and determine when they will next align.\\
The hour hand of a clock completes a full cycle every 12 hours, so it moves at a speed of 360 degrees per 12 hours, which is 30 degrees per hour. The minute hand completes a full cycle every 60 minutes, so it moves at a speed of 360 degrees per 60 minutes, which is 6 degrees per minute.\\
At 10 o'clock, the hour hand is at the 300-degree mark, and the minute hand is at the 0-degree mark. The minute hand will need to "catch up" to the hour hand. \\
Every hour, the hour hand moves 30 degrees, while the minute hand moves 360 degrees. The relative speed is the difference in their speeds, so the minute hand catches up to the hour hand at a rate of 360 - 30 = 330 degrees per hour, or 5.5 degrees per minute (since there are 60 minutes in an hour).
We need to find out how many degrees the minute hand needs to catch up from the 10 o'clock position. At 10 o'clock, there is a 300 - 0 = 300-degree difference between the two hands. \\
\textbf{\purple{Hallucination and Incorrect reasoning}}
\end{mybody}
\begin{mybody}
\textbf{Gemini Pro-Vision Response:} 
The minute hand moves 6 degrees per minute, and the hour hand moves 0.5 degrees per minute. So, the minute hand moves 1. \\
\textbf{\purple{Does not ground the task to the image}}
\end{mybody}
\begin{mybody}
\textbf{LLaVA-Next-34B:} The hour and minute hands on the clock are currently aligned at approximately \textbf{\red{1:15}}. To determine how much time remains until the hour and minute hands align again, we need to find the next time when the hour hand will be at the same position as it is now, which is at the number 1. Since the hour hand moves one number every hour, we can calculate the time remaining by subtracting the current time from the next time the hour hand will be at 1. The next time the hour hand will be at 1 is 2 hours from now, which is 3:15. Now, we subtract the current time from 3:15 to find the time remaining: 3:15 - 1:15 = 2 hours. So, there are 2 hours remaining until the hour and minute hands align again. \\
\textbf{\purple{Hallucination and Incorrect reasoning}}
\end{mybody}
\begin{mybody}
\textbf{ShareGPT-4V-7B Response:} 
The hour hand of the clock is pointing to 12, and the minute hand is pointing to 3. To determine how much time remains until the hour and minute hands align, we need to calculate the difference between the current time and the time when the hands align. Current time: 12:03, Hour hand position: 12, \textbf{\red{Minute hand position: 3}}. To find the time when the hands align, we need to add the number of minutes to the current hour. In this case, the time when the hands align will be 12:03 + 3 = 12:06. So, there are 3 minutes remaining until the hour and minute hands align.\\
\textbf{\purple{Hallucination and Incorrect reasoning}}
\end{mybody}
\begin{mybody}
\textbf{LLaVA-1.5-13B Response:} 
The hour hand on the clock is at 12, and \textbf{\red{the minute hand is at 10}}. To determine the time remaining until the hour and minute hands align, we need to find the difference between the current time and the next multiple of 12. Since the hour hand moves in increments of 12, the next multiple of 12 will be at 1 o'clock. The time difference between the current time (12:10) and the next multiple of 12 (1:00) is 1 hour and 50 minutes. This means that there are 1 hour and 50 minutes left until the hour and minute hands align. \\
\textbf{\purple{Hallucination and Incorrect reasoning}}
\end{mybody}
\begin{mybody}
\textbf{GPT-4 w/ Layout-aware OCR + Caption Response:} 
\textbf{\red{30 minutes}} \\
\textbf{\purple{Wrong response}}
\end{mybody}
\captionof{figure}{In this task, \textit{Gemini-Vision-Pro} fails to ground the task to this image. Whereas, \textit{GPT4V, LLaVA-Next-34B, LLaVA-1.5-13B \& ShareGPT4V-7B} suffer from hallucination and incorrect reasoning. Lastly, the \textit{GPT-4 w/ Layout-aware OCR + Caption Response} (Augmented LLM approach) produces the wrong response.}
\label{fig:time_wrong_2}
\end{table*}

\clearpage

\subsection{Shopping}
This section provides qualitative analysis of the \textit{Shopping} visual scenario and our findings across the models, as shown in Figures \ref{fig:shopping_correct_1}, \ref{fig:shopping_correct_2}, \ref{fig:shopping_wrong_1}, \ref{fig:shopping_wrong_2}..

\begin{table*}[th!]
\fontsize{9.0pt}{\baselineskip}\selectfont
\linespread{0.9}\selectfont
\begin{mybody}
\includegraphics[scale=0.4]{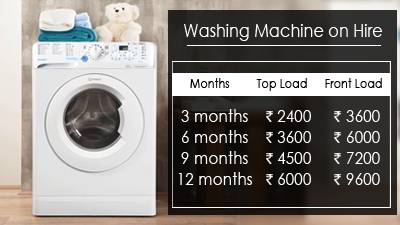}
\\
\textbf{Instruction:} Provide the cost of hiring the machine type depicted in the image for nine months.\\\\
\textbf{Reference Response:} 
The machine shown in the image is a front-loading machine, and the rental fee for a period of 9 months amounts to INR 7200. 
\end{mybody}
\begin{mybody}
\textbf{Human Response:} \textbf{\green{Front load machines cost 7200 for 9 months.}}
\end{mybody}
\begin{mybody}
\textbf{GPT-4V Response:} The image displays a front load washing machine along with a price list for hiring different types of washing machines. For hiring a front load washing machine for a period of 9 months, the cost would be \textbf{\green{INR 7,200}} according to the given price list.
\end{mybody}
\begin{mybody}
\textbf{Gemini Pro-Vision Response:} The cost of hiring the machine depicted in the image for nine months is \textbf{\red{INR 4500.}}\\
\textbf{\purple{Lack of context-sensitive text-rich visual reasoning}}
\end{mybody}
\begin{mybody}
\textbf{LLaVA-Next-34B:} The cost of hiring the machine type depicted in the image for nine months is 4500 rupees for a top load machine and \textbf{\green{7200 rupees for a front load machine}}.
\end{mybody}
\begin{mybody}
\textbf{ShareGPT-4V-7B Response:} The cost of hiring the washing machine for nine months is \textbf{\red{4500 rupees}}. \\
\textbf{\purple{Lack of context-sensitive text-rich visual reasoning}}
\end{mybody}
\begin{mybody}
\textbf{LLaVA-1.5-13B Response:} The cost of hiring the machine type depicted in the image for nine months is \textbf{\red{60,000 rupees}}. \\
\textbf{\purple{Lack of context-sensitive text-rich visual reasoning \& Hallucination}}
\end{mybody}
\begin{mybody}
\textbf{GPT-4 w/ Layout-aware OCR + Caption Response:} The cost of hiring the depicted machine type (Top Load) for nine months is \textbf{\red{4500 units (currency not specified in the provided text)}}\\
\textbf{\purple{Lack of context-sensitive text-rich visual reasoning}}
\end{mybody}
\captionof{figure}{In this task, apart from \textit{GPT4V} and \textit{LLaVA-Next-34B}, all other models produce the wrong response. This can be attributed to the strong fine-grained visual perception and context-sensitive text-rich visual reasoning abilities} of GPT4V. This also shows that improvement in visual encoding capabilities of open-source LMMs, like \textit{LLaVA-Next-34B} over \textit{LLaVA-1.5-13B} improves performance on context-sensitive text-rich visual reasoning tasks. 
\label{fig:shopping_correct_1}
\end{table*}


\begin{table*}[th!]
\fontsize{9.0pt}{\baselineskip}\selectfont
\linespread{0.9}\selectfont
\begin{mybody}
\includegraphics[scale=0.4]{images/qualitative_examples/shopping/correct/screen-shot-2017-10-19-at-12-49-38-am.png}
\\
\textbf{Instruction:} Describe the most similar product between the living room and the hip living room.\\\\
\textbf{Reference Response:} 
The orangish-brown colored single-seating sofa is the most similar between the Living Room and the Hip Living Room. 
\end{mybody}
\begin{mybody}
\textbf{Human Response:} \textbf{\green{Both living rooms have similar orange chairs.}}
\end{mybody}
\begin{mybody}
\textbf{GPT-4V Response:} The most similar products between the "Living Room" and "Hip Living Room" sets appear to be \textbf{\green{the armchairs. Both sets feature a single armchair with a similar style and design, characterized by a mid-century modern aesthetic, with splayed legs and an angled backrest.}}
\end{mybody}
\begin{mybody}
\textbf{Gemini Pro-Vision Response:} The most similar product between the living room and the hip living room \textbf{\red{is the sofa. Both sofas are blue and have a similar shape.}}\\
\textbf{\purple{Lack of fine-grained visual perception.}}
\end{mybody}
\begin{mybody}
\textbf{LLaVA-Next-34B:} The most similar product between the living room and the hip living room is \textbf{\red{the sofa. Both images feature a sofa as a central piece of furniture in the living room setup. The color of the sofa in the living room is a neutral gray, while the hip living room has a blue sofa.}} The other items in the living room and hip living room are different, with the living room having a wooden coffee table and a floor lamp. The hip living room has a more contemporary coffee table and a floor lamp with a unique design. \\
\textbf{\purple{Lack of fine-grained visual perception.}}
\end{mybody}
\begin{mybody}
\textbf{ShareGPT-4V-7B Response:} The most similar product between the living room and the hip living room is the \textbf{\red{couch. Both living room and hip living room couches are designed to provide comfort and serve as a focal point in the room.}}\\
\textbf{\purple{Lack of fine-grained visual perception.}}
\end{mybody}
\begin{mybody}
\textbf{LLaVA-1.5-13B Response:} The most similar product between the living room and the hip living room is \textbf{\red{the couch. Both living rooms feature a couch as a central piece of furniture, providing a comfortable seating area for relaxation and socializing. The couches in both rooms have a similar design.}}\\
\textbf{\purple{Lack of fine-grained visual perception.}}
\end{mybody}
\begin{mybody}
\textbf{GPT-4 w/ Layout-aware OCR + Caption Response:} The most similar product between the Living Room and the Hip Living Room is in terms of the quantity of items. \textbf{\red{Both living room options include 4 items each - a couch, a coffee table, and a side table. The relative positioning and type of furniture also seem to be similar, although the styles and colors vary.}}\\
\textbf{\purple{Lack of context-sensitive text-rich visual reasoning}}

\end{mybody}
\captionof{figure}{In this task, apart from \textit{GPT4V}, all other models produce the wrong response. This can be attributed to the strong fine-grained visual perception and context-sensitive text-rich visual reasoning of GPT4V. Further, \textit{GPT-4 w/ Layout-aware OCR + Caption Response} (Augmented LLM approach) shows inability to correctly reason with the provided dense image information as text, thereby, highlighting the need for context-sensitive text-rich visual reasoning.}
\label{fig:shopping_correct_2}
\end{table*}



\begin{table*}[th!]
\fontsize{9.0pt}{\baselineskip}\selectfont
\linespread{0.9}\selectfont
\begin{mybody}
\includegraphics[scale=0.3]{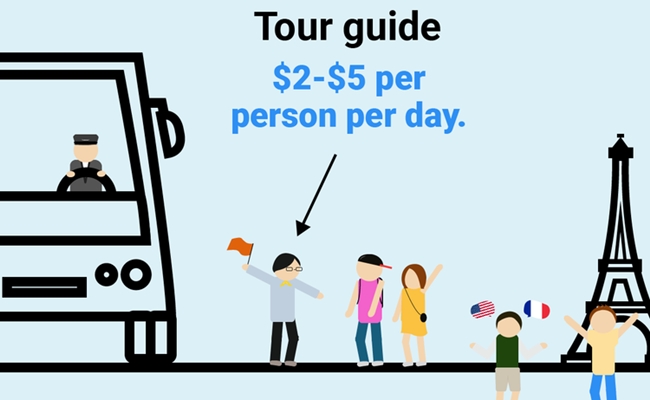}
\\
\textbf{Instruction:} Calculate the maximum money the tour guide can make from the tourists shown in this image.\\\\
\textbf{Reference Response:} 
The picture depicts four potential tourists, one tour guide, and one bus driver. Considering that the tour guide earns between \$2 and \$5 per person, the highest amount they can earn from four tourists is \$20.
\end{mybody}
\begin{mybody}
\textbf{Human Response:} \textbf{\green{There are four tourists shown in the image. 5 x 4 = \$20.}}
\end{mybody}
\begin{mybody}
\textbf{GPT-4V Response:} To calculate the maximum amount of money the tour guide can make from the tourists shown in this image in a day, we need to count the number of tourists and multiply that by the maximum rate of \$5 per person per day. \textbf{\red{In the image, there are five tourists visible. So, the calculation is as follows: 5 tourists x \$5/tourist/day = \$25/day. Therefore, the maximum amount of money the tour guide can make from the tourists shown in this image per day is \$25.}}\\
\textbf{\purple{Lack of fine-grained visual perception.}}
\end{mybody}
\begin{mybody}
\textbf{Gemini Pro-Vision Response:} \textbf{\red{The image shows 5 tourists and the tour guide charges \$2-5 per person per day. So the maximum money the tour guide can make is \$25 per day}}\\
\textbf{\purple{Lack of fine-grained visual perception.}}
\end{mybody}
\begin{mybody}
\textbf{LLaVA-Next-34B:} \textbf{\red{The image shows a group of tourists, but it does not provide an exact number of people. To calculate the maximum money the tour guide can make from the tourists shown, I would need to know the total number of people in the group. If you can provide that information, I can help you with the calculation.}} \\
\textbf{\purple{Lack of fine-grained visual perception}}
\end{mybody}
\begin{mybody}
\textbf{ShareGPT-4V-7B Response:} 
The tour guide can make a maximum of \textbf{\red{\$25}} per day from the tourists shown in the image. \textbf{\red{This is because the image shows five tourists standing next to each other, and the tour guide is holding an orange flag. The tour guide's maximum earnings would be \$25 per day, per person, considering there are five tourists in the group.}}\\
\textbf{\purple{Lack of fine-grained visual perception.}}

\end{mybody}
\begin{mybody}
\textbf{LLaVA-1.5-13B Response:} To calculate the maximum money the tour guide can make from the tourists, we need to determine the number of tourists and the cost per person. \textbf{\red{In the image, there are five people standing in front of the bus, and the tour guide is offering a \$25-\$5 per person per day deal. Since there are five people, the maximum amount of money the tour guide can make is \$25 x 5 = \$125 per day.}}\\
\textbf{\purple{Lack of fine-grained visual perception.}}
\end{mybody}
\begin{mybody}
\textbf{GPT-4 w/ Layout-aware OCR + Caption Response:} \textbf{\red{The maximum money the tour guide can make from the five tourists shown in this image is \$25 per day (\$5 per person per day).}}\\
\textbf{\purple{Lack of fine-grained visual perception.}}

\end{mybody}
\captionof{figure}{In this task, all  models produce the wrong response. This can be attributed to the lack of fine-grained visual perception} for this task.
\label{fig:shopping_wrong_1}
\end{table*}



\begin{table*}[th!]
\fontsize{9.0pt}{\baselineskip}\selectfont
\linespread{0.9}\selectfont
\begin{mybody}
\includegraphics[scale=0.2]{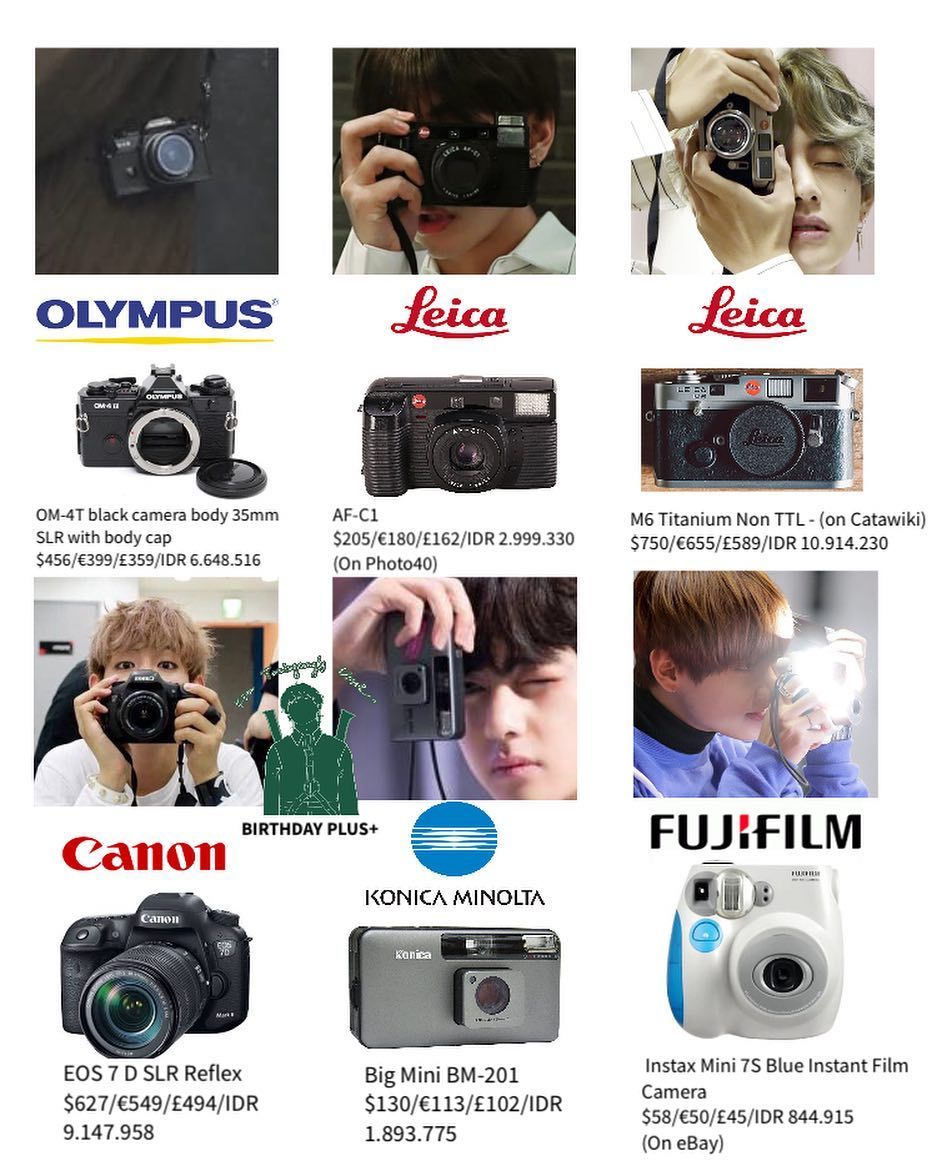}
\\
\textbf{Instruction:} Discuss the features of a camera being held vertically with a single hand.\\\\
\textbf{Reference Response:} 
Konica Minolta is the camera held vertically in one hand, specifically the Big Mini BM-201 model, priced at \$130, €113, £102, or IDR 1,893,775.
\end{mybody}
\begin{mybody}
\textbf{Human Response:} \textbf{\green{The camera held by a single hand vertically is a Big Mini and it features a price tag of \$130.}}
\end{mybody}
\begin{mybody}
\textbf{GPT-4V Response:} \textbf{\red{The image appears to be a collage of various cameras with a person holding one camera vertically in the top right corner. Since we're focusing on the vertically held camera, here are some general features and considerations for using a camera in a vertical orientation with a single hand.}}  \\\textbf{\purple{Lack of context-sensitive text-rich visual reasoning}}
\end{mybody}
\begin{mybody}
\textbf{Gemini Pro-Vision Response:} \textbf{\red{The camera being held vertically with a single hand is a Fujifilm Instax Mini 7s. It is a compact and lightweight instant camera that is perfect.}} \\
\textbf{\purple{Wrong response}}
\end{mybody}
\begin{mybody}
\textbf{LLaVA-Next-34B:} \textbf{\red{The image shows a person holding a camera vertically with a single hand. This type of camera is likely a mirrorless or compact camera, as it appears to be relatively small and lightweight, allowing for such a grip. The camera is equipped with a lens, which is typical for cameras that capture images with a higher level of detail and quality compared to smartphones...}} \\
\textbf{\purple{Lack of context-sensitive text-rich visual reasoning}}
\end{mybody}
\begin{mybody}
\textbf{ShareGPT-4V-7B Response:} \textbf{\red{ The camera being held vertically with a single hand is a Canon EOS 5D Mark IV. This camera is known for ...}} \\
\textbf{\purple{Lack of context-sensitive text-rich visual reasoning \& Hallucination}}
\end{mybody}
\begin{mybody}
\textbf{LLaVA-1.5-13B Response:}
\textbf{\red{A camera being held vertically with a single hand is typically a compact or point-and-shoot camera ...}} \\
\textbf{\purple{Lack of context-sensitive text-rich visual reasoning}}
\end{mybody}
\begin{mybody}
\textbf{GPT-4 w/ Layout-aware OCR + Caption Response:} \textbf{\red{The image does not specify the features of a camera being held vertically with a single hand. However, a camera designed for this use would typically be lightweight and compact...}} \\
\textbf{\purple{Lack of context-sensitive text-rich visual reasoning}}

\end{mybody}
\captionof{figure}{In this task, all  models except \textit{Gemini-Vision-Pro}, show lack of context-sensitive text-rich visual reasoning.}
\label{fig:shopping_wrong_2}
\end{table*}

\clearpage

\subsection{Navigation}
This section provides qualitative analysis of the \textit{Navigation} visual scenario and our findings across the models, as shown in Figures \ref{fig:navigation_correct_1}, \ref{fig:navigation_correct_2}, \ref{fig:navigation_wrong_1}, \ref{fig:navigation_wrong_2}.

\begin{table*}[th!]
\fontsize{9.0pt}{\baselineskip}\selectfont
\linespread{0.9}\selectfont
\begin{mybody}
\includegraphics[scale=0.4]{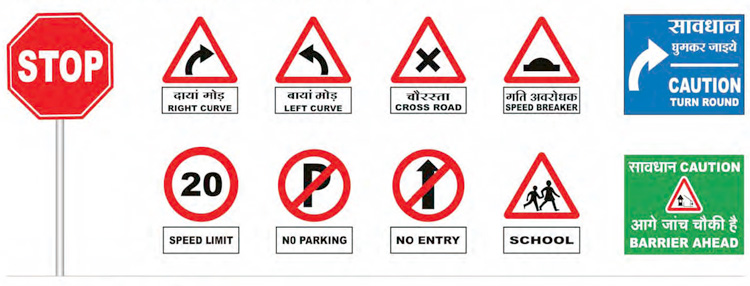}
\\
\textbf{Instruction:} Determine the triangular-shaped signs.\\\\
\textbf{Reference Response:} 
Triangular-shaped signs are Right Curve, Left Curve, Cross Road, Speed Breaker, and School.
\end{mybody}
\begin{mybody}
\textbf{Human Response:} \textbf{\green{The triangular shaped signs are Right Curve, Left Curve, Cross Road, and Speed Breaker, in order from left to right, in the top row. The the second row, the last image is triangular and is for School.
}}
\end{mybody}
\begin{mybody}
\textbf{GPT-4V Response:} The triangular shaped signs in the image are as follows:
1. The second sign on the top row, indicating a \textbf{\green{"RIGHT CURVE."}}
2. The third sign on the top row, indicating a \textbf{\green{"LEFT CURVE."}}
3. The fourth sign on the top row, indicating a \textbf{\green{"CROSS ROAD."}}
4. The fifth sign on the top row, with an icon of a bump, indicating a \textbf{\green{"SPEED BREAKER."}}
5. The third sign on the bottom row, with an icon of two people, indicating a \textbf{\green{"SCHOOL"}} zone.
\end{mybody}
\begin{mybody}
\textbf{Gemini Pro-Vision Response:} There are \textbf{\red{6}} triangular-shaped signs:
\textbf{\green{1. Right curve}}
\textbf{\green{2. Left curve}}
\textbf{\green{3. Cross road}}
\textbf{\green{4. Speed breaker}} \\
\textbf{\purple{Partially correct response and hallucination.}}
\end{mybody}
\begin{mybody}
\textbf{LLaVA-Next-34B:} The triangular-shaped signs in the image are: \textbf{\green{"LEFT CURVE"}} in white, \textbf{\green{"RIGHT CURVE"}} in white, \textbf{\red{"LEFT TURN"}} in white, \textbf{\red{"RIGHT TURN"}} in white,\textbf{\red{"NO ENTRY"}} in white. \\
\textbf{\purple{Partially correct response and hallucination.}}
\end{mybody}
\begin{mybody}
\textbf{ShareGPT-4V-7B Response:} \textbf{\red{The triangular-shaped signs in the image are the "No Parking" sign and the "No Entry" sign.}}. \\
\textbf{\purple{Lack of fine-grained visual perception.}}
\end{mybody}
\begin{mybody}
\textbf{LLaVA-1.5-13B Response:} \textbf{\red{The triangular-shaped signs in the image are warning signs, which are typically used to convey important information or instructions to drivers and pedestrians ...}}. \\
\textbf{\purple{Lack of context-sensitive text-rich visual reasoning}}
\end{mybody}
\begin{mybody}
\textbf{GPT-4 w/ Layout-aware OCR + Caption Response:} The triangular-shaped signs in the image are:
\textbf{\red{1. STOP}}
\textbf{\red{2. CAUTION}}
\textbf{\red{3. RIGHT CURVE}}
\textbf{\red{4. LEFT CURVE}}
\textbf{\red{5. CROSS ROAD}}
\textbf{\red{6. SPEED BREAKER}}
\textbf{\red{7. TURN ROUND}}
\textbf{\red{8. RIGHT CAUTION}}
\textbf{\red{9. SPEED LIMIT}}
\textbf{\red{10. NO PARKING}}
\textbf{\red{11. NO ENTRY}}
\textbf{\red{12. SCHOOL}}
\textbf{\red{13. BARRIER AHEAD}} \\
\textbf{\purple{Lack of context-sensitive text-rich visual reasoning}}
\end{mybody}
\captionof{figure}{In this task \textit{GPT4V} produces the correct response, \textit{Gemini-Vision-Pro} \& \textit{LLaVA-Next-34B} produce a partially correct answer with minor hallucinations. All other models hallucinate, and produce the wrong response or both.}
\label{fig:navigation_correct_1}
\end{table*}


\begin{table*}[th!]
\fontsize{9.0pt}{\baselineskip}\selectfont
\linespread{0.9}\selectfont
\begin{mybody}
\includegraphics[scale=0.4]{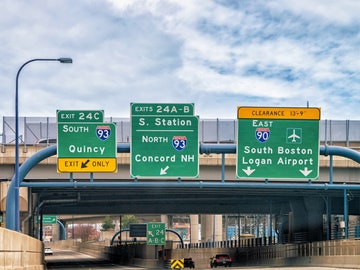}
\\
\textbf{Instruction:} Itemize the destinations and exits one can take while driving on the left. \\\\
\textbf{Reference Response:} When driving in the left lane, the options for directions and exits include taking Exit 24C for South Quincy on Route 93 or Exits 24 A-B, which lead to South Station or North Concord (NH on Route 93), respectively.
\end{mybody}
\begin{mybody}
\textbf{Human Response:} \textbf{\green{24A-B S. Station, North 93, Concord NH
}}
\end{mybody}
\begin{mybody}
\textbf{GPT-4V Response:} \textbf{\green{ If you're driving on the left and considering your potential destinations and exits based on the signage in the image, here they are:
\\Exit 24C: Take this exit if your destination is Quincy; it's an "EXIT ONLY" lane.
\\Exits 24A-B: Options available from these exits: 24A South Station, 24B North 93, Concord NH
}}
\end{mybody}
\begin{mybody}
\textbf{Gemini Pro-Vision Response:} \textbf{\green{Take Exit 24C for Quincy.
Take Exit 24A-B for South Station and North for Concord, NH.
}}
\end{mybody}
\begin{mybody}
\textbf{LLaVA-Next-34B:} \textbf{\red{While driving on the left side of the highway, as indicated by the signage in the image, you can take the following destinations and exits:1. Exit 24C: South 93 to Quincy 2. Exit 24A: North 93 to Concord, NH 3. Exit 24B: South Station 4. Exit 24A-B: S. Station 5. Clearance 13'-9'6. East 90 7. South Boston 8. Logan Airport.Please note that the specific lanes for each exit may vary, and it's always best to check the road signs and markings for the most accurate information while driving.}} \\
\textbf{\purple{Lack of context-sensitive text-rich visual reasoning}}
\end{mybody}
\begin{mybody}
\textbf{ShareGPT-4V-7B Response:} \textbf{\red{While driving on the left, there are several destinations and exits to consider. These include Quincy, South Boston Logan Airport, Concord NH, and S Station. Additionally, there are exits for 24A, 24B, 93, and 90. These exits provide drivers with the opportunity to reach various locations and navigate through the city effectively.
}} \\
\textbf{\purple{Lack of context-sensitive text-rich visual reasoning \& Hallucination}}
\end{mybody}
\begin{mybody}
\textbf{LLaVA-1.5-13B Response:} \textbf{\red{While driving on the left, there are several destinations and exits available for drivers to take. These include: Exit 24A-B: This exit is for drivers heading towards the south station, Exit 24C: This exit is for drivers heading towards Quincy, Exit 24D: This exit is for drivers heading towards Concord, NH, Exit 24E: This exit is for drivers heading towards South Boston, Exit 24F: This exit is for drivers heading towards Logan Airport. These exits are located under a highway overpass, and drivers can choose the appropriate exit based on their desired destination.}} \\
\textbf{\purple{Lack of context-sensitive text-rich visual reasoning \& Hallucination}}
\end{mybody}
\begin{mybody}
\textbf{GPT-4 w/ Layout-aware OCR + Caption Response:} \textbf{\red{The destinations and exits one can take while driving on the left are: South 93, Concord NH, and South Boston Logan Airport.}} \\
\textbf{\purple{Lack of context-sensitive text-rich visual reasoning}}
\end{mybody}
\captionof{figure}{In this task, closed-source LMMs \textit{GPT-4V} and \textit{Gemini Pro-Vision} correctly respond to the instruction. This can be attributed to their ability for visual perception, text recognition, and context-sensitive text-rich visual reasoning. All other models lack context-sensitive text-rich visual reasoning and some even contain hallucinations in their responses.}
\label{fig:navigation_correct_2}
\end{table*}



\begin{table*}[th!]
\fontsize{9.0pt}{\baselineskip}\selectfont
\linespread{0.9}\selectfont
\begin{mybody}
\includegraphics[scale=1.5]{images/qualitative_examples/navigation/incorrect/102_space.jpeg}
\\
\textbf{Instruction:} Specify the colors of the rocket, space shuttle, and external tank.\\\\
\textbf{Reference Response:} 
The space shuttle is depicted in yellow, the rockets are depicted in grey, and the external tank is depicted in red.
\end{mybody}
\begin{mybody}
\textbf{Human Response:} \textbf{\green{The rocket is gray. The space shuttle is yellow. The external tank is red.}}
\end{mybody}
\begin{mybody}
\textbf{GPT-4V Response:} 
In the image, the solid rocket boosters of the \textbf{\red{space shuttle are colored in red and light gray, with the red at the top and the gray on the bottom. The external tank is colored orange. The main body of the space shuttle, also known as the orbiter, is primarily white with black on its nose and wing edges. There is also a bit of red on the tail section.}} The image depicts various stages of a space shuttle flight sequence from pre-launch to landing. \\
\textbf{\purple{Hallucination}}

\end{mybody}
\begin{mybody}
\textbf{Gemini Pro-Vision Response:} \textbf{\red{The rocket is red and white, the space shuttle is white, and the external tank is orange.}} \\
\textbf{\purple{Hallucination}}
\end{mybody}
\begin{mybody}
\textbf{LLaVA-Next-34B:} The \textbf{\red{rocket is red and white}}, \textbf{\green{the space shuttle is yellow}}, and \textbf{\red{the external tank is white.}} \\
\textbf{\purple{Partially correct response with Hallucination}}
\end{mybody}
\begin{mybody}
\textbf{ShareGPT-4V-7B Response:} The \textbf{\red{rocket is red}}, \textbf{\green{the space shuttle is yellow}}, and \textbf{\red{the external tank is orange.}} \\
\textbf{\purple{Partially correct response with Hallucination}}
\end{mybody}
\begin{mybody}
\textbf{LLaVA-1.5-13B Response:} \textbf{\red{The rocket is red}}, \textbf{\green{the space shuttle is yellow}}, and \textbf{\red{the external tank is white.}}\\
\textbf{\purple{Partially correct response with Hallucination}}
\end{mybody}
\begin{mybody}
\textbf{GPT-4 w/ Layout-aware OCR + Caption Response:} \textbf{\red{The rocket is red and yellow. The space shuttle is yellow and red. The image did not specify the color of the external tank.}}
\\
\textbf{\purple{Hallucination}}
\end{mybody}
\captionof{figure}{In this task, all models hallucinate, due to lack of grounding the response to the given image \& fine-grained visual perception and using prior visual knowledge. Notably, the open-source LMMs, \textit{LLaVA-Next-34B}. \textit{LLaVA-1.5-13B} \& \textit{ShareGPT-4V-7B} provide partially accurate responses, suggesting a reduced reliance on prior visual knowledge, possibly due to not having seen it.}
\label{fig:navigation_wrong_1}
\end{table*}

\begin{table*}[th!]
\fontsize{9.0pt}{\baselineskip}\selectfont
\linespread{0.9}\selectfont
\begin{mybody}
\includegraphics[scale=0.4]{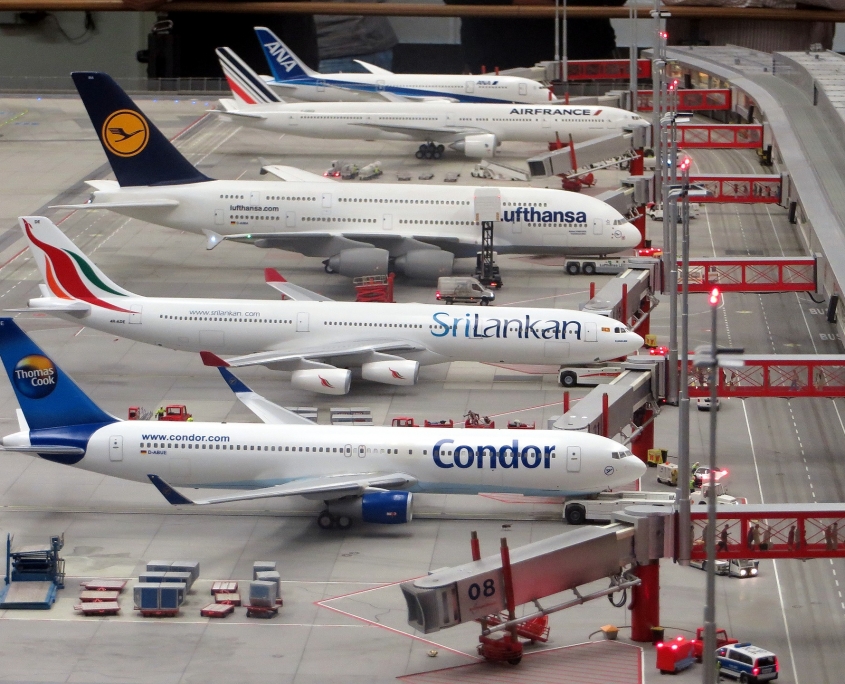}
\\
\textbf{Instruction:} Pick out the airlines immediately in front and behind the one featuring a flying bird in yellow and blue.\\\\
\textbf{Reference Response:} 
The airline right in front of the airline with a bird logo is SriLankan Airlines, and the one situated behind it is Air France.
\end{mybody}
\begin{mybody}
\textbf{Human Response:} \textbf{\green{Sri Lankan and Air France.
}}
\end{mybody}
\begin{mybody}
\textbf{GPT-4V Response:} 
\textbf{\red{The airline with a flying bird on its fuselage is "SriLankan Airlines." Directly in front of it, the airplane features the livery of "Lufthansa," and right behind the SriLankan Airlines aircraft, the airplane with the dark blue tail and the text "Condor" represents Condor Airlines.}}\\
\textbf{\purple{Lack of fine-grained visual perception}}
\end{mybody}
\begin{mybody}
\textbf{Gemini Pro-Vision Response:} \textbf{\red{ In front: Condor, Behind: Sri Lankan}}\\
\textbf{\purple{Lack of fine-grained visual perception \& Hallucination}}
\end{mybody}
\begin{mybody}
\textbf{LLaVA-Next-34B:} \textbf{\red{The airline immediately in front of the one featuring a flying bird in yellow and blue is "SriLankan". The airline immediately behind it is "Condor"}} \\
\textbf{\purple{Lack of fine-grained visual perception \& Hallucination}}
\end{mybody}
\begin{mybody}
\textbf{ShareGPT-4V-7B Response:} \textbf{\red{The airlines in front of the one with the flying bird are Condor and SriLankan. The airline behind the one with the flying bird is Lufthansa.}}\\
\textbf{\purple{Lack of fine-grained visual perception \& Hallucination}}
\end{mybody}
\begin{mybody}
\textbf{LLaVA-1.5-13B Response:}
\textbf{\red{In the image, the airline with a flying bird logo is Sri Lankan Airlines. The airline in front of it is Condor, and the airline behind it is Air France.
}}\\
\textbf{\purple{Lack of fine-grained visual perception}}
\end{mybody}
\begin{mybody}
\textbf{GPT-4 w/ Layout-aware OCR + Caption Response:} 
\textbf{\red{The airlines immediately in front and behind the one featuring a flying bird, Condor, are Srilankan and Lufthansa respectively.}}\\
\textbf{\purple{Lack of context-sensitive text-rich visual reasoning}}

\end{mybody}
\captionof{figure}{In this task, \textit{all models} produce the wrong response, due to lack of fine-grained visual perception.}
\label{fig:navigation_wrong_2}
\end{table*}

\clearpage

\subsection{Abstract}
This section provides qualitative analysis of the \textit{Abstract} visual scenario and our findings across the models, as shown in Figures \ref{fig:abstract_correct_1}, \ref{fig:abstract_wrong_1}.

\begin{table*}[th!]
\fontsize{9.0pt}{\baselineskip}\selectfont
\linespread{0.9}\selectfont
\begin{mybody}
\includegraphics[scale=0.2]{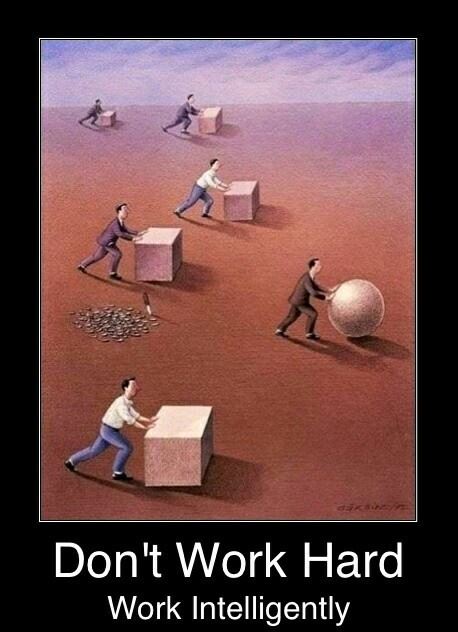}
\\
\textbf{Instruction:} Contextualize the saying with the visual contents of the image. \\\\
\textbf{Reference Response:} 
The saying "Work smart, not hard" is illustrated in the accompanying image. It depicts multiple people pushing square wooden blocks across a flat surface, while one person carves a wooden block into a sphere and easily transports it across the same surface. 
\end{mybody}
\begin{mybody}
\textbf{Human Response:} \textbf{\green{The person working intelligently turned their square into a circle so they could roll it. They aren't working hard, but intelligently.
}}
\end{mybody}
\begin{mybody}
\textbf{GPT-4V Response:} \textbf{\green{The image depicts a series of individuals pushing cubes across a flat surface, which appears to be a difficult and labor-intensive task. In contrast, one individual is shown pushing a spherical object—a ball—which naturally rolls and thereby requires much less effort to move. The saying, "Don't Work Hard, Work Intelligently," encapsulates the idea that the approach to a task can greatly affect the amount of effort required to complete it. The person with the ball has found a more efficient method, using a shape that is easier to move, demonstrating intelligence in their approach to the work, as opposed to the others who are working hard but not necessarily in the most effective way with the cube shapes.}}

\end{mybody}
\begin{mybody}
\textbf{Gemini Pro-Vision Response:} \textbf{\green{The image shows a group of people pushing square blocks. Only one person is pushing a round object, which is much easier to push than a square block.}} 
\end{mybody}
\begin{mybody}
\textbf{LLaVA-Next-34B:} \textbf{\green{The image shows a sequence of figures, each depicted in a different pose, with the last figure holding a large ball. The sequence appears to represent a progression from a more challenging task to a more efficient or intelligent approach. The saying "Don't Work Hard, Work Intelligently" is superimposed on the image, which suggests that the figures are illustrating the concept of finding smarter ways to accomplish tasks rather than merely exerting effort. The progression from the first figure, who is struggling with a small box, to the last figure, who is effortlessly holding a large ball, implies that working intelligently can lead to greater success with less effort.}}
\end{mybody}
\begin{mybody}
\textbf{ShareGPT-4V-7B Response:} \textbf{\red{The image portrays a scene of six men, each engaged in a different activity, all working together to move a large white ball. The men are dressed in suits and are using various tools, such as shovels and a hammer, to maneuver the ball. The setting appears to be a desert-like environment, with the men and the ball being the main focus of the image...}} \\
\textbf{\purple{Hallucination}}
\end{mybody}
\begin{mybody}
\textbf{LLaVA-1.5-13B Response:} \textbf{\red{The image features a painting of a man pushing a large cube on wheels, which is a representation of the saying "Don't waste your time." The man is depicted in various positions, including pushing the cube, pulling it, and even running away from it ...}}\\
\textbf{\purple{Hallucination}}
\end{mybody}
\begin{mybody}
\textbf{GPT-4 w/ Layout-aware OCR + Caption Response:} \textbf{\red{The saying "Don't Work Hard, Work Intelligently" is contextualized by the visual representation of people engaged in a seemingly endless task of moving large cubes. This serves as a metaphor to highlight the concept behind the saying. The act of pushing, pulling, and lifting these blocks signifies the traditional 'hard work.' ... }}\\
\textbf{\purple{Hallucination}}
\end{mybody}
\vspace{-1em}
\captionof{figure}{In this task, closed-source LMMs \textit{GPT-4V}, \textit{Gemini Pro-Vision} \& \textit{LLaVA-Next-34B} correctly respond to the instruction. This can be attributed to their ability for visual perception, text recognition, and context-sensitive text-rich visual reasoning. All other models hallucinate their responses.}
\label{fig:abstract_correct_1}
\end{table*}



\begin{table*}[th!]
\fontsize{9.0pt}{\baselineskip}\selectfont
\linespread{0.9}\selectfont
\begin{mybody}
\includegraphics[scale=0.22]{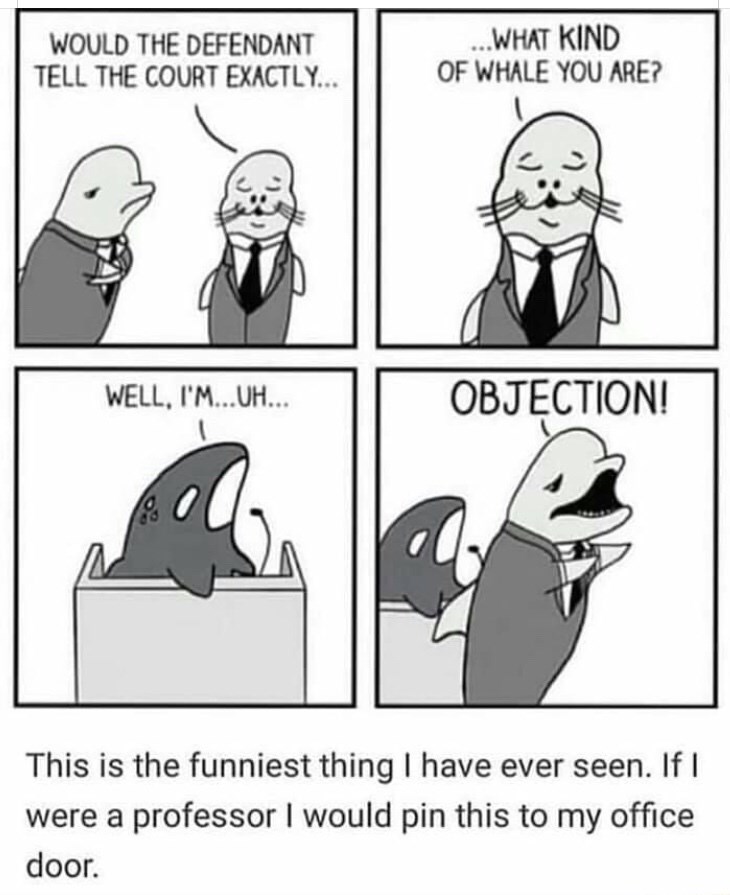}
\\
\textbf{Instruction:} Comment on the funny aspect of the comic strip.\\\\
\textbf{Reference Response:} 
In this four-panel comic, the lawyer (seal) asks the defendant (killer whale) to identify itself, but the whale hesitates because admitting to being a "killer" would mean confessing to the crime. In the final panel, the defense lawyer (dolphin) exclaims "objection, " evoking humor through this conundrum.
\end{mybody}
\begin{mybody}
\textbf{Human Response:} \textbf{\green{The whale depicted in the strip is a "Killer" Whale.}}
\end{mybody}
\begin{mybody}
\textbf{GPT-4V Response:} 
\textbf{\red{This comic strip might be considered particularly funny for a few reasons: Anthropomorphism: The characters are animals dressed in human clothes and partaking in a court trial, which is inherently comical. Visual humor: The expressions and body language of the animals add to the comedic effect, as they mimic human courtroom drama. Irony: It's humorous and ironic because whales obviously cannot speak, let alone participate in a legal procedure, making the whole situation absurdly funny.}}\\
\textbf{\purple{Lack of context-sensitive text-rich visual reasoning}}
\end{mybody}
\begin{mybody}
\textbf{Gemini Pro-Vision Response:} \textbf{\red{The comic strip is funny because it plays on the fact that whales are often seen as intelligent and articulate creatures, but in this case, the whale is unable to do so.}}\\
\textbf{\purple{Lack of context-sensitive text-rich visual reasoning}}
\end{mybody}
\begin{mybody}
\textbf{LLaVA-Next-34B:} \textbf{\red{The comic strip features a humorous exchange between two characters, one of whom is a whale. The humor arises from the unexpected and absurd nature of the conversation. The first character asks the whale, "Would the defendant tell the court exactly... what kind of whale you are?" This question is funny because it's an unusual and irrelevant inquiry in a court setting. The whale's response, "Well, I'm... uh...", is funny because it's a human-like reaction to being put on the spot, and the whale's hesitation suggests it's unsure how to answer. The final panel, where the whale objects, is a play on the legal term "objection" and adds to the absurdity of the situation, as whales do not participate in legal proceedings...}}\\
\textbf{\purple{Lack of context-sensitive text-rich visual reasoning}}
\end{mybody}
\begin{mybody}
\textbf{ShareGPT-4V-7B Response:} \textbf{\red{The viewer considers this comic strip to be the funniest thing they have ever seen because it features a humorous and unexpected scenario. In the comic strip, a man and a seal are depicted in a conversation, with the seal sitting in a chair and holding a briefcase. The comic strip's unexpected twist, where a seal is dressed in a suit and holding a briefcase, creates a sense of humor and amusement that the viewer finds entertaining.}}\\
\textbf{\purple{Lack of context-sensitive text-rich visual reasoning}}
\end{mybody}
\begin{mybody}
\textbf{LLaVA-1.5-13B Response:} \textbf{\red{A viewer might consider this comic strip to be the funniest thing they have ever seen because it features a unique and unexpected combination of elements. The strip includes a courtroom scene with a seal in a suit, a dolphin in a witness box, and a whale in a tie. This unusual and absurd scenario creates a humorous and unexpected twist on a serious topic, which can be seen as a form of satire or parody.}}\\
\textbf{\purple{Lack of context-sensitive text-rich visual reasoning}}
\end{mybody}
\begin{mybody}
\textbf{GPT-4 w/ Layout-aware OCR + Caption Response:} \textbf{\red{The viewer might find this comic strip to be the funniest thing they have ever seen due to the absurdity and humour of the situation: a man and a fish having a legal discourse...}}\\
\textbf{\purple{Lack of context-sensitive text-rich visual reasoning \& Hallucination}}

\end{mybody}
\captionof{figure}{In this task, \textit{all models} produce the wrong response, due to lack of context-sensitive text-rich visual reasoning. Specifically, for augmented LLM approach, it is lack of of context-sensitive text-rich visual reasoning.}
\label{fig:abstract_wrong_1}
\end{table*}

\clearpage

\subsection{Application Usage}
This section provides qualitative analysis of the \textit{Application Usage} visual scenario and our findings across the models, as shown in Figures \ref{fig:app_usage_correct_1}, \ref{fig:app_usage_correct_2}, \ref{fig:app_usage_wrong_1}, \ref{fig:app_usage_wrong_2}.

\begin{table*}[th!]
\fontsize{9.0pt}{\baselineskip}\selectfont
\linespread{0.9}\selectfont
\begin{mybody}
\includegraphics[scale=0.27]{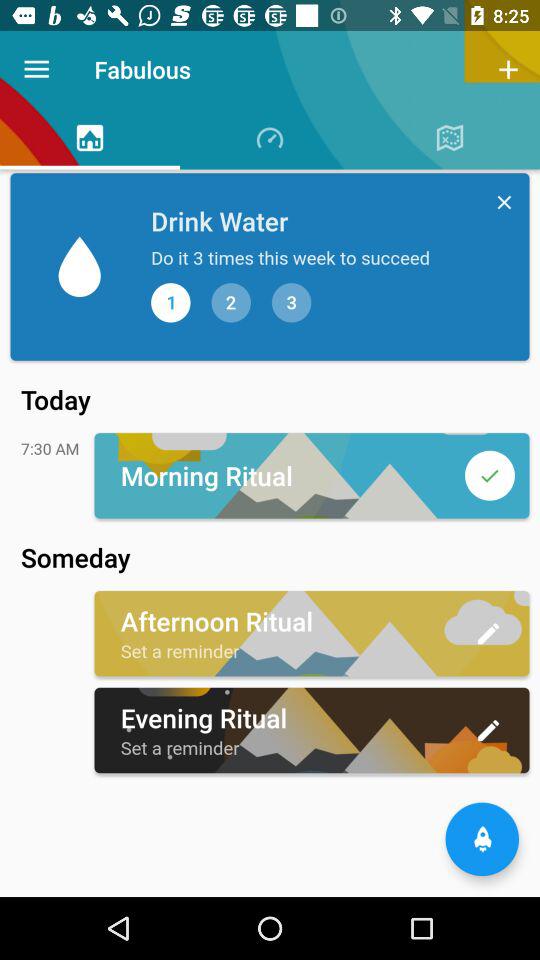}
\\
\textbf{Instruction:} How many times has the user drank water today?\\\\
\textbf{Reference Response:} 
1 times since only the option with "1" on it is selected
\end{mybody}
\begin{mybody}
\textbf{Human Response:} \textbf{\green{Once}}
\end{mybody}
\begin{mybody}
\textbf{GPT-4V Response:} Based on the screenshot provided, it appears that the user has marked \textbf{\green{one instance of drinking water today. The user seems to be using an application that tracks habits, and the "Drink Water" card shows one filled circle, indicating one time the user has logged drinking water for the current day.}}
\end{mybody}
\begin{mybody}
\textbf{Gemini Pro-Vision Response:} \textbf{\red{The user has not drank any water today.}}\\
\textbf{\purple{Lack of fine-grained visual perception}}
\end{mybody}
\begin{mybody}
\textbf{LLaVA-Next-34B:} \textbf{\green{The user has drank water once today.}}
\end{mybody}
\begin{mybody}
\textbf{ShareGPT-4V-7B Response:} \textbf{\red{The user has drank water three times today.}}\\
\textbf{\purple{Lack of fine-grained visual perception}}
\end{mybody}
\begin{mybody}
\textbf{LLaVA-1.5-13B Response:} \textbf{\red{The user has drank water three times today.}} \\
\textbf{\purple{Lack of fine-grained visual perception}}
\end{mybody}
\begin{mybody}
\textbf{GPT-4 w/ Layout-aware OCR + Caption Response:} \textbf{\red{The user has drank water 3 times today.}}\\
\textbf{\purple{Lack of fine-grained visual perception}}
\end{mybody}
\captionof{figure}{In this task \textit{GPT4V} \& \textit{LLaVA-Next-34B} produces the correct response, and all other models hallucinate, due to lack of fine-grained visual perception ability. This also
shows that improvement in visual encoding capabilities of open-source LMMs, like \textit{LLaVA-Next-34B} over \textit{LLaVA-1.5-13B} improves performance on context-sensitive text-rich visual reasoning tasks.}
\label{fig:app_usage_correct_1}
\end{table*}


\begin{table*}[th!]
\fontsize{9.0pt}{\baselineskip}\selectfont
\linespread{0.9}\selectfont
\begin{mybody}
\includegraphics[scale=0.25]{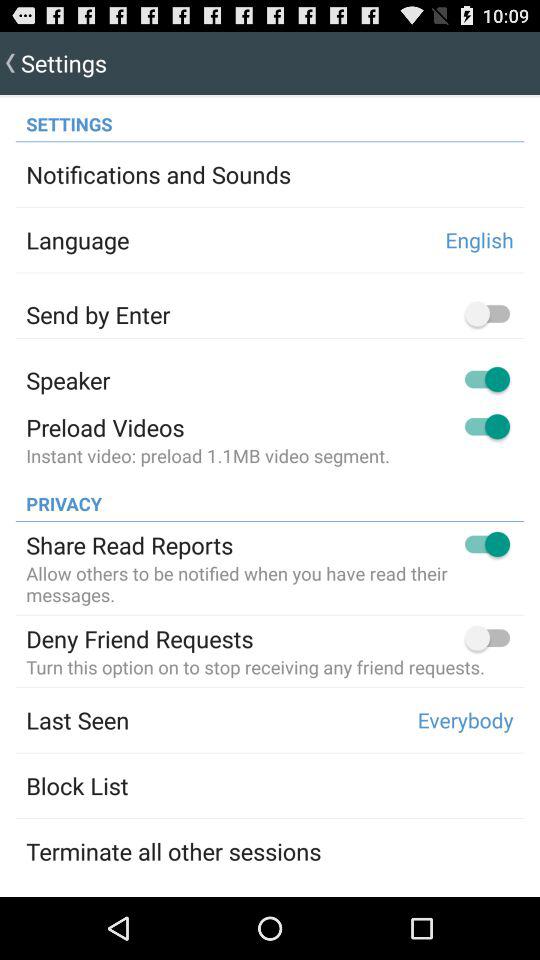}
\\
\textbf{Instruction:} Highlight the settings that are switched off by the user.\\\\
\textbf{Reference Response:} 
Send by Enter and Deny Friend Requests
\end{mybody}
\begin{mybody}
\textbf{Human Response:} \textbf{\green{Switched off: Send by Enter and Deny Friend Requests}}
\end{mybody}
\begin{mybody}
\textbf{GPT-4V Response:} \textbf{\green{The settings switched off in this image are "Send by Enter" and "Deny Friend Requests." These options are indicated by grayed-out toggle switches, while the options that are switched on are indicated by green-colored toggle switches.}}
\end{mybody}
\begin{mybody}
\textbf{Gemini Pro-Vision Response:} \textbf{\green{1. Send by Enter
2. Deny Friend Requests}}
\end{mybody}
\begin{mybody}
\textbf{LLaVA-Next-34B:} \textbf{\red{The settings that are switched off by the user are: 1. Notifications and Sounds 2. Language 3. Send by Enter 4. Speaker 5. Preload Videos 6. Share Read Reports 7. Deny Friend Requests 8. Last Seen 9. Block List.}}\\
\textbf{\purple{Lack of fine-grained visual perception \& Hallucination}}
\end{mybody}
\begin{mybody}
\textbf{ShareGPT-4V-7B Response:} \textbf{\red{In the image, the user has switched off the "Last Seen" and "Everybody" settings. The "Last Seen" setting is turned off, which means the user's last seen time will not be visible to others. The "Everybody" setting is also turned off, which means the user's messages will not be visible to everyone, and the user will not receive messages from everyone.
}}\\
\textbf{\purple{Lack of fine-grained visual perception \& Hallucination}}
\end{mybody}
\begin{mybody}
\textbf{LLaVA-1.5-13B Response:} \textbf{\red{In the image, there is a cell phone with several settings turned off. These settings include the speaker, preloaded videos, and instant video. The speaker setting is turned off, which means that the user may not be able to hear notifications or ringtones. The preloaded videos setting is also turned off, which means that the user may not be able to watch videos that were pre-installed on the device. Lastly, the instant video setting is turned off, which means that the user may not be able to use the instant video feature on the device.
}}\\
\textbf{\purple{Lack of fine-grained visual perception \& Hallucination}}
\end{mybody}
\begin{mybody}
\textbf{GPT-4 w/ Layout-aware OCR + Caption Response:} \textbf{\green{The "Deny Friend Requests" setting is switched off by the user.}} \\
\textbf{\purple{Partially correct response}}

\end{mybody}
\captionof{figure}{In this task, the closed source LMMs, \textit{GPT4V \& Gemini-Vision-Pro} fail to ground the task to this image. Whereas, the open source LMMs, \textit{LLaVA-Next-34B, LLaVA-1.5-13B \& ShareGPT4V-7B} suffer from hallucination. Notably, the \textit{GPT-4 w/ Layout-aware OCR + Caption Response} (Augmented LLM approach) produces a partially correct response. On analyzing the visual information provided to GPT4 for reasoning, we find the captions contain information about "Deny Friend Requests" being set to false. The captions are generated using \textit{ShareGPT-4V-7B}, but when given this task, it hallucinates the answer.}
\label{fig:app_usage_correct_2}
\end{table*}



\begin{table*}[th!]
\fontsize{9.0pt}{\baselineskip}\selectfont
\linespread{0.9}\selectfont
\begin{mybody}
\includegraphics[scale=0.09]{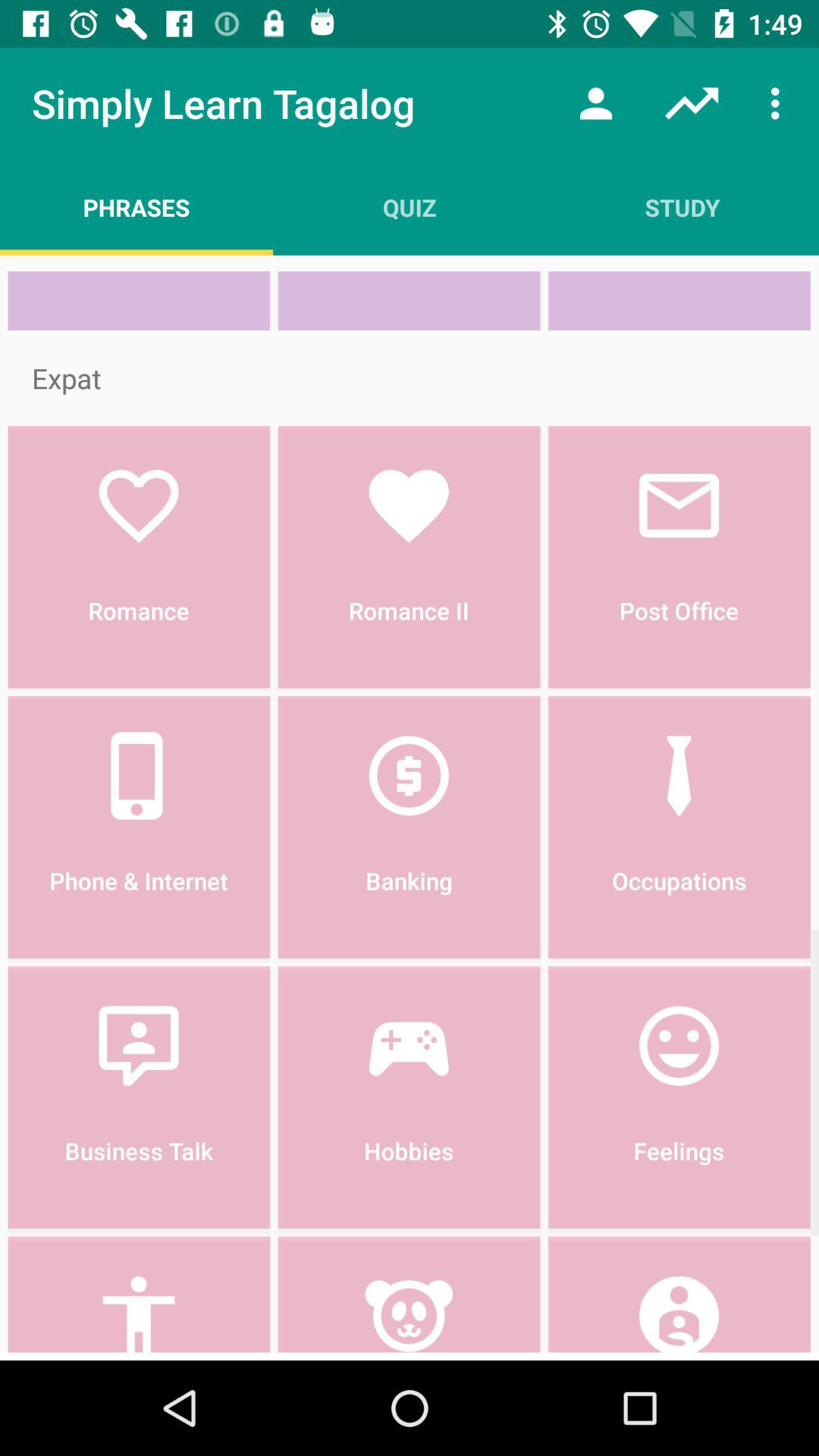}
\\
\textbf{Instruction:} What is the difference between the illustration of Romance and Romance II?\\\\
\textbf{Reference Response:} 
The logo for "Romance" features a heart with an unshaded interior, while the logo for "Romance II" showcases a heart filled with white shading inside.
\end{mybody}
\begin{mybody}
\textbf{Human Response:} \textbf{\green{Romance II has a filled in heart vs not filled in.}}
\end{mybody}
\begin{mybody}
\textbf{GPT-4V Response:} \textbf{\red{The illustration of "Romance" and "Romance II" in the image both feature a heart icon, which is a widely recognized symbol of love and affection. The difference between the two appears to be simply numerical, possibly indicating a continuation or second set of phrases related to romance within the language learning app "Simply Learn Tagalog." There's no visual difference between the icons themselves, suggesting the distinction is in content rather than visual or thematic change.}} \\
\textbf{\purple{Lack of fine-grained visual perception \& Hallucination}}

\end{mybody}
\begin{mybody}
\textbf{Gemini Pro-Vision Response:} \textbf{\red{The difference between the two is that the first one has one heart, while the second one has two.}} \\
\textbf{\purple{Lack of fine-grained visual perception \& Hallucination}}

\end{mybody}
\begin{mybody}
\textbf{LLaVA-Next-34B:} \textbf{\red{The illustration for "Romance" features a heart symbol, which is a universal symbol for love and affection. The illustration for "Romance II" also features a heart symbol, but it is accompanied by a small lock, which could represent the idea of commitment or the act of locking one's heart to someone. This might suggest that "Romance II" is a more advanced or deeper form of romance, implying a stronger bond or a more serious relationship.}}\\
\textbf{\purple{Lack of fine-grained visual perception \& Hallucination}}
\end{mybody}
\begin{mybody}
\textbf{ShareGPT-4V-7B Response:} 
\textbf{\green{The difference between the illustrations of Romance and Romance II lies in the design of the heart symbol.}} \textbf{\red{In the first illustration, the heart symbol is depicted as a simple, two-dimensional shape. In contrast, the heart symbol in the second illustration, Romance II, is more detailed and intricately designed.}}
 \\
\textbf{\purple{Partially correct response but not completely grounded to the image}}

\end{mybody}
\begin{mybody}
\textbf{LLaVA-1.5-13B Response:} \textbf{\red{The illustration of Romance and Romance II in the app are different in terms of the visual representation. Romance is represented by a heart symbol, while Romance II is represented by a heart with a plus sign on it.
}} \\
\textbf{\purple{Lack of fine-grained visual perception \& Hallucination}}

\end{mybody}
\begin{mybody}
\textbf{GPT-4 w/ Layout-aware OCR + Caption Response:} \textbf{\red{The image does not provide distinct illustrations for "Romance" and "Romance II". Only one icon related to romance is described in the image caption - a rose. The instruction might be referring to different elements, but based on the provided OCR text and image caption, there is not enough information to distinguish between "Romance" and "Romance II".}} \\
\textbf{\purple{Lack of context-sensitive text-rich visual reasoning}}

\end{mybody}
\captionof{figure}{In this task, the closed source LMMs, \textit{GPT4V}, \textit{Gemini-Vision-Pro}, \textit{LLaVA-Next-34B} \& \textit{LLaVA-1.5-13B}  hallucinate in their responses. \textit{GPT-4 w/ Layout-aware OCR + Caption Response} (Augmented LLM approach) produces the wrong answer, due to lack of context-sensitive text-rich visual reasoning. However, \textit{ShareGPT4V-7B} produces a partially correct response but is unable to ground it completely to the image.}
\label{fig:app_usage_wrong_1}
\end{table*}



\begin{table*}[th!]
\fontsize{9.0pt}{\baselineskip}\selectfont
\linespread{0.9}\selectfont
\begin{mybody}
\includegraphics[scale=0.15]{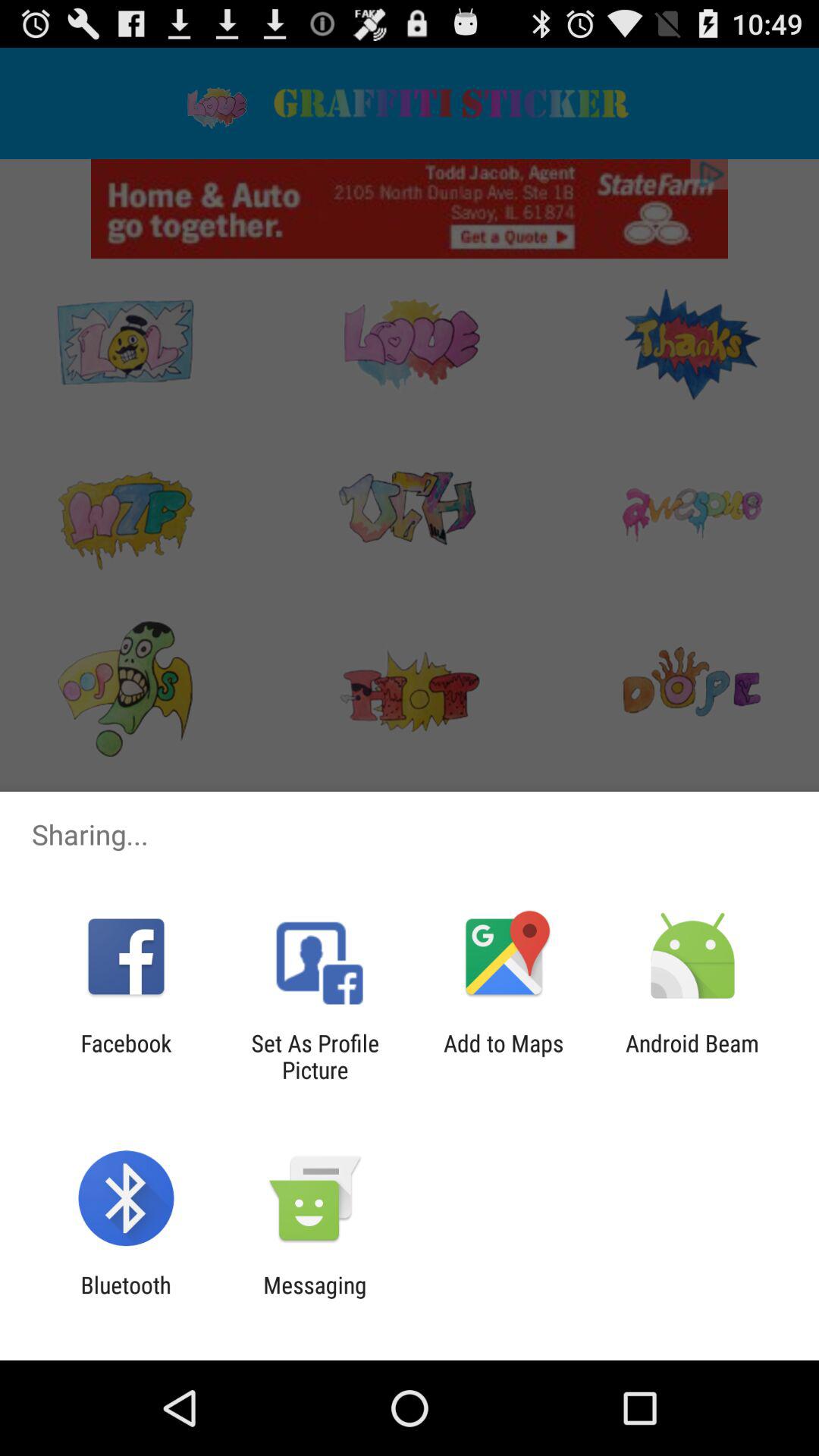}
\\
\textbf{Instruction:} List the words formed by all the grafitti stickers.\\\\
\textbf{Reference Response:} 
Lol, Love, Thanks, Wtf, Ugh, Awesome, Oops, Hot, Dope
\end{mybody}
\begin{mybody}
\textbf{Human Response:} \textbf{\green{The words are LOL, love, Thanks, WTF, UGH, awesome, oops, HoT and Dope.}}
\end{mybody}
\begin{mybody}
\textbf{GPT-4V Response:} The graffiti stickers each have a different word on them. From what I can discern, the words are: \textbf{\green{LOVE}}, \textbf{\green{THANKS}}, \textbf{\red{WIZ}}, \textbf{\red{FISH}}, \textbf{\green{AWESOME}}, \textbf{\green{HOT}}, \textbf{\green{DOPE}}. \\
\textbf{\purple{Partially correct response with Hallucination}}
\end{mybody}
\begin{mybody}
\textbf{Gemini Pro-Vision Response:} \textbf{\red{Home, Auto, Wut, Uhh}}
\textbf{\green{Love,Thanks,Awesome,Dope,Hot}} \\
\textbf{\purple{Partially correct response with Hallucination}}
\end{mybody}
\begin{mybody}
\textbf{LLaVA-Next-34B:} \textbf{\red{The image shows a collection of graffiti stickers, but without a specific word or phrase in mind, it's not possible to list the words formed by them. Graffiti stickers often feature abstract designs, characters, or symbols that may not form words in a traditional sense. If you have a particular word or phrase in mind, please provide more context or details.}}\\
\textbf{\purple{Lack of fine-grained visual perception \& Lack of context-sensitive text-rich visual reasoning}}
\end{mybody}
\begin{mybody}
\textbf{ShareGPT-4V-7B Response:} \textbf{\green{Love, Awesome, Dope}}, \textbf{\red{Facebook, Google, Android Beam
}} \\
\textbf{\purple{Hallucination along with partial correct response}}
\end{mybody}
\begin{mybody}
\textbf{LLaVA-1.5-13B Response:}
The words formed by all the grafitti stickers are \textbf{\green{"LOVE".
}}\\
\textbf{\purple{Partially correct response}}
\end{mybody}
\begin{mybody}
\textbf{GPT-4 w/ Layout-aware OCR + Caption Response:} \textbf{\red{The words formed by the graffiti stickers are "GRAFONSNCINER", "UTH", "AVESOS".}}
\\
\textbf{\purple{Hallucination}}

\end{mybody}
\captionof{figure}{In this task, all models show a varying ability to recognise stylistic text, with \textit{GPT4V} being the best and surprisingly \textit{GPT-4 w/ Layout-aware OCR + Caption Response} (Augmented LLM approach) is the worst. This underscores the need of open-ended text recognition models, that can understand a variety of text styles.} 
\label{fig:app_usage_wrong_2}
\end{table*}

\clearpage

\subsection{Web Usage}
This section provides qualitative analysis of the \textit{Web Usage} visual scenario and our findings across the models, as shown in Figures \ref{fig:web_usage_correct_1}, \ref{fig:web_usage_correct_2}, \ref{fig:web_usage_wrong_1}, \ref{fig:web_usage_wrong_2}.

\begin{table*}[th!]
\fontsize{9.0pt}{\baselineskip}\selectfont
\linespread{0.9}\selectfont
\begin{mybody}
\includegraphics[scale=0.15]{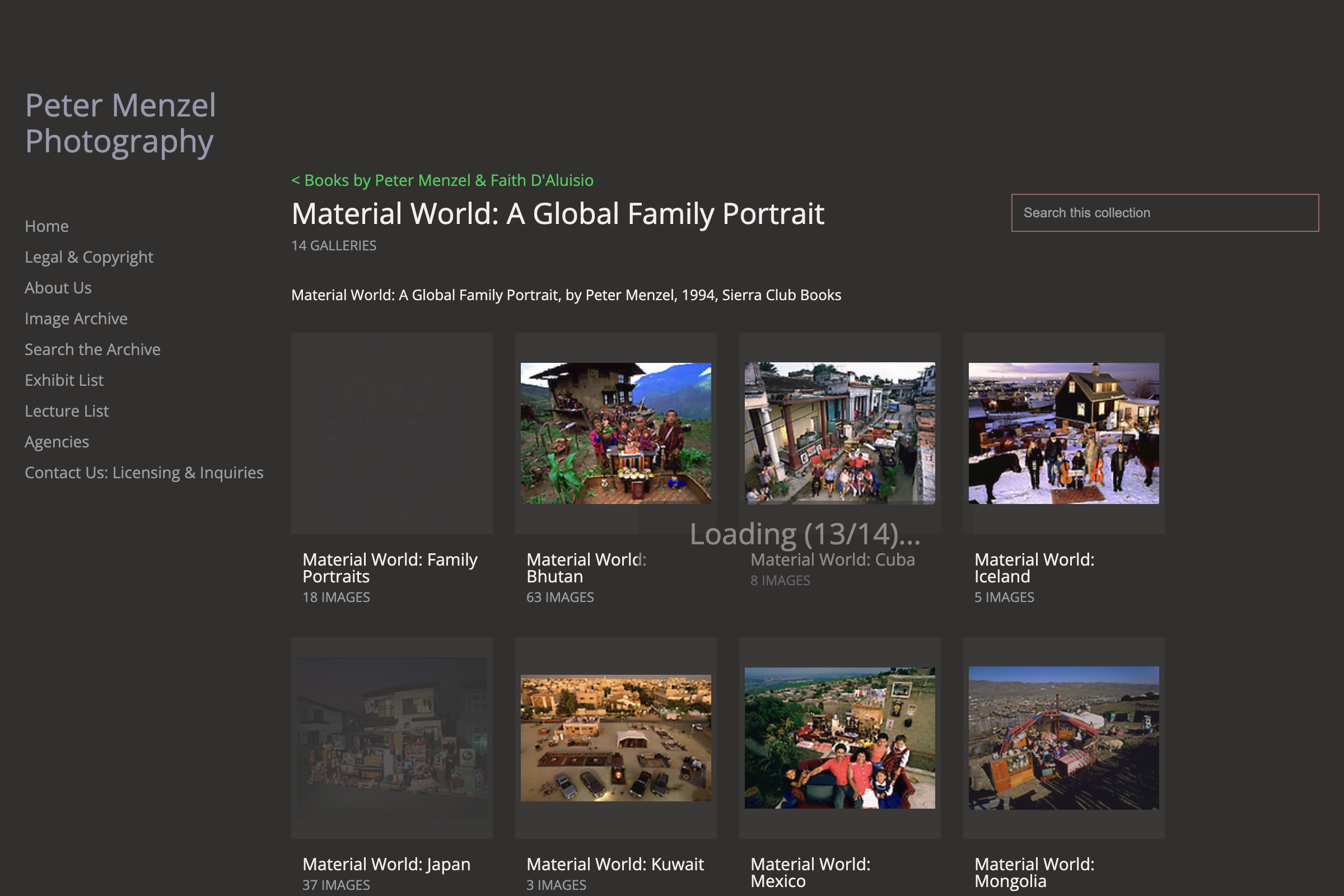}
\\
\textbf{Instruction:} Name the country whose portrait has a group of people on the road.\\\\
\textbf{Reference Response:} 
Cuba
\end{mybody}
\begin{mybody}
\textbf{Human Response:} \textbf{\green{Cuba}}
\end{mybody}
\begin{mybody}
\textbf{GPT-4V Response:} \textbf{\green{The image with a group of people on the road is associated with "Material World: Cuba."}}
\end{mybody}
\begin{mybody}
\textbf{Gemini Pro-Vision Response:} \textbf{\red{Bhutan}}\\
\textbf{\purple{Lack of fine-grained visual perception}}
\end{mybody}
\begin{mybody}
\textbf{LLaVA-Next-34B:} \textbf{\red{The country whose portrait shows a group of people on the road is Bhutan.}}\\
\textbf{\purple{Lack of fine-grained visual perception}}
\end{mybody}
\begin{mybody}
\textbf{ShareGPT-4V-7B Response:} \textbf{\red{Mexico}}\\
\textbf{\purple{Lack of fine-grained visual perception}}
\end{mybody}
\begin{mybody}
\textbf{LLaVA-1.5-13B Response:} \textbf{\red{Bhutan}} \\
\textbf{\purple{Lack of fine-grained visual perception}}
\end{mybody}
\begin{mybody}
\textbf{GPT-4 w/ Layout-aware OCR + Caption Response:} \textbf{\red{The image does not provide specific information on which country's portrait has a group of people on the road.}}\\
\textbf{\purple{Lack of context-sensitive text-rich visual reasoning}}
\end{mybody}
\captionof{figure}{In this task \textit{GPT4V} produces the correct response, and \textit{all other models} produce the wrong answer, due to lack of fine-grained visual perception ability. Specifically, \textit{GPT-4 w/ Layout-aware OCR + Caption Response} (Augmented LLM approach) lacks context-sensitive text-rich visual reasoning ability to respond to this task.}
\label{fig:web_usage_correct_1}
\end{table*}


\begin{table*}[th!]
\fontsize{9.0pt}{\baselineskip}\selectfont
\linespread{0.9}\selectfont
\begin{mybody}
\includegraphics[scale=0.15]{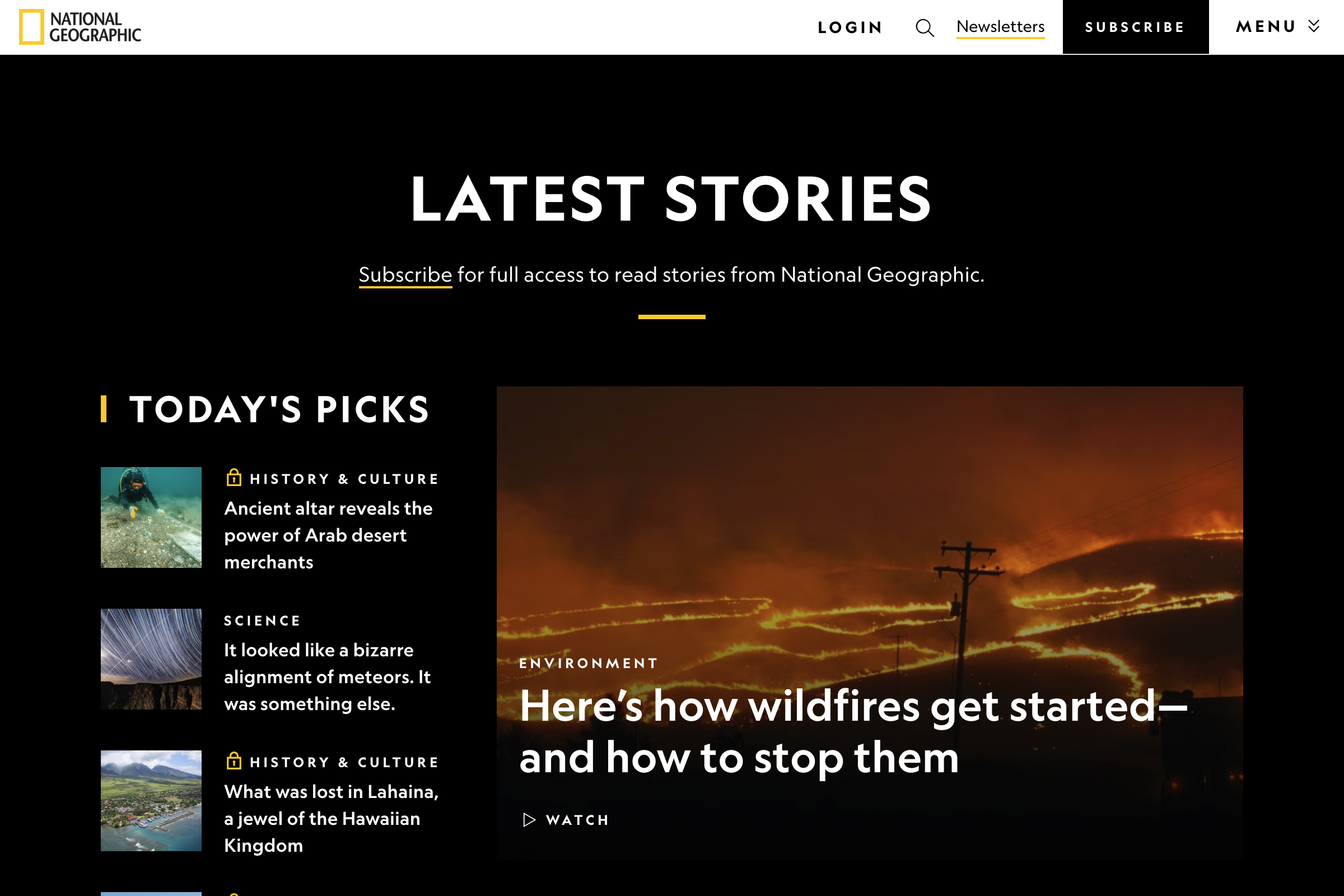}
\\
\textbf{Instruction:} Which story can be read without subscription under Today's pick?\\\\
\textbf{Reference Response:} 
The story with the title "Science: It looked like a bizarre alignment of meteors. It was something else" because it does not have lock sign which other two stories have.
\end{mybody}
\begin{mybody}
\textbf{Human Response:} \textbf{\green{It looked like a bizarre alignment of meteors.  It was something else.}}
\end{mybody}
\begin{mybody}
\textbf{GPT-4V Response:} \textbf{\red{I'm sorry, but I can't provide real-time or current information about which specific articles are available without a subscription on external websites such as National Geographic. You would need to visit their website directly to see which stories are accessible for free.}}\\
\textbf{\purple{Lack of instruction understanding and grounding to the image}}
\end{mybody}
\begin{mybody}
\textbf{Gemini Pro-Vision Response:} \textbf{\green{The story titled "It looked like a bizarre alignment of meteors. It was something else" can be read without a subscription.}}
\end{mybody}
\begin{mybody}
\textbf{LLaVA-Next-34B:} \textbf{\red{Under "Today's Picks" on the National Geographic website, there are several stories that can be read without a subscription. These include: 1. "Ancient altar reveals the power of Arab desert merchants" under the "History \& Culture" section. 2. "It looked like a bizarre alignment of meteors. It was something else." under the "Science" section. 3. "What was lost in Lahaina, a jewel of the Hawaiian Kingdom" under the "History \& Culture" section...}}\\
\textbf{\purple{Lack of context-sensitive text-rich visual reasoning}}
\end{mybody}
\begin{mybody}
\textbf{ShareGPT-4V-7B Response:} \textbf{\red{The story "Here's how wildfires get started and how to stop them" can be read without subscription under Today's pick.}}\\
\textbf{\purple{Lack of context-sensitive text-rich visual reasoning}}
\end{mybody}
\begin{mybody}
\textbf{LLaVA-1.5-13B Response:} \textbf{\red{Under Today's pick, the story "Here's how wildfires get started and how to stop them" can be read without subscription. This story is available for full access, while the other stories require a subscription to National Geographic.}}\\
\textbf{\purple{Lack of context-sensitive text-rich visual reasoning}}
\end{mybody}
\begin{mybody}
\textbf{GPT-4 w/ Layout-aware OCR + Caption Response:} \textbf{\red{The stories that can be read without subscription under Today's Picks are: "Ancient altar reveals the power of Arab desert merchants" under the category "History \& Culture"...}} \\
\textbf{\purple{Lack of context-sensitive text-rich visual reasoning}}

\end{mybody}
\captionof{figure}{In this task \textit{Gemini-Vision-Pro} produces the correct response. \textit{GPT4V} fails to understands the task and ground it to the give image. The open source LMMs, \textit{LLaVA-1.5-13B \& ShareGPT4V-7B} produce the wrong response, due to lack of context-sensitive text-rich visual reasoningg. \textit{LLaVA-Next-34B \& GPT-4 w/ Layout-aware OCR + Caption Response} (Augmented LLM approach) hallucinates along with producing the correct response.}
\label{fig:web_usage_correct_2}
\end{table*}



\begin{table*}[th!]
\fontsize{9.0pt}{\baselineskip}\selectfont
\linespread{0.9}\selectfont
\begin{mybody}
\includegraphics[scale=0.16]{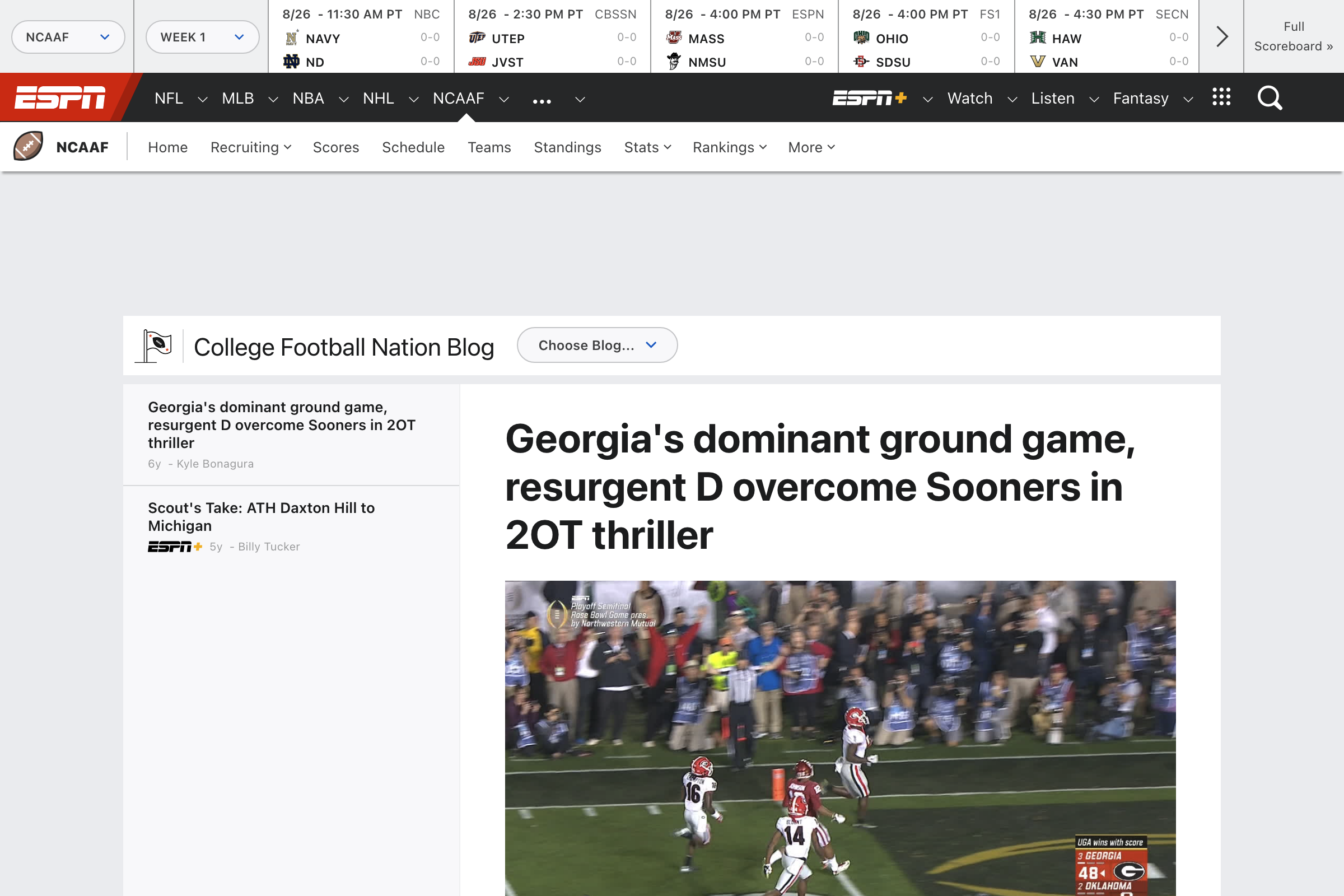}
\\
\textbf{Instruction:} Give the jersey number of the leftmost player on the ground from the white team.\\\\
\textbf{Reference Response:} 
16
\end{mybody}
\begin{mybody}
\textbf{Human Response:} \textbf{\green{The jersey number is 16}}
\end{mybody}
\begin{mybody}
\textbf{GPT-4V Response:} \textbf{\red{I'm sorry, but I can't provide specific details such as jersey numbers for real individuals in images due to privacy reasons and the limitations you've mentioned.}}
\end{mybody}
\begin{mybody}
\textbf{Gemini Pro-Vision Response:} \textbf{\red{The jersey number is 10.}}\\
\textbf{\purple{Hallucination}}
\end{mybody}
\begin{mybody}
\textbf{LLaVA-Next-34B:} \textbf{\green{The jersey number of the leftmost player on the ground from the white team is 16.}}
\end{mybody}
\begin{mybody}
\textbf{ShareGPT-4V-7B Response:} 
\textbf{\red{The leftmost player on the ground from the white team is wearing a jersey with the number 44.
}}\\
\textbf{\purple{Hallucination}}

\end{mybody}
\begin{mybody}
\textbf{LLaVA-1.5-13B Response:} \textbf{\red{The leftmost player on the ground from the white team is wearing a jersey with the number 44.}}\\
\textbf{\purple{Hallucination}}
\end{mybody}
\begin{mybody}
\textbf{GPT-4 w/ Layout-aware OCR + Caption Response:} \textbf{\red{The image does not provide specific jersey numbers for the players.}}\\
\textbf{\purple{Hallucination}}

\end{mybody}
\captionof{figure}{Notably, in this example \textit{LLaVA-Next-34B} is the only model that gets the correct response. All other models either do not produce a response or produce the wrong response due to lack of strong text recognition capabilities. This also
shows that improvement in visual encoding capabilities of open-source LMMs, like \textit{LLaVA-Next-34B} over \textit{LLaVA-1.5-13B} improves performance on context-sensitive text-rich visual reasoning tasks.}
\label{fig:web_usage_wrong_1}
\end{table*}



\begin{table*}[th!]
\fontsize{9.0pt}{\baselineskip}\selectfont
\linespread{0.9}\selectfont
\begin{mybody}
\includegraphics[scale=0.16]{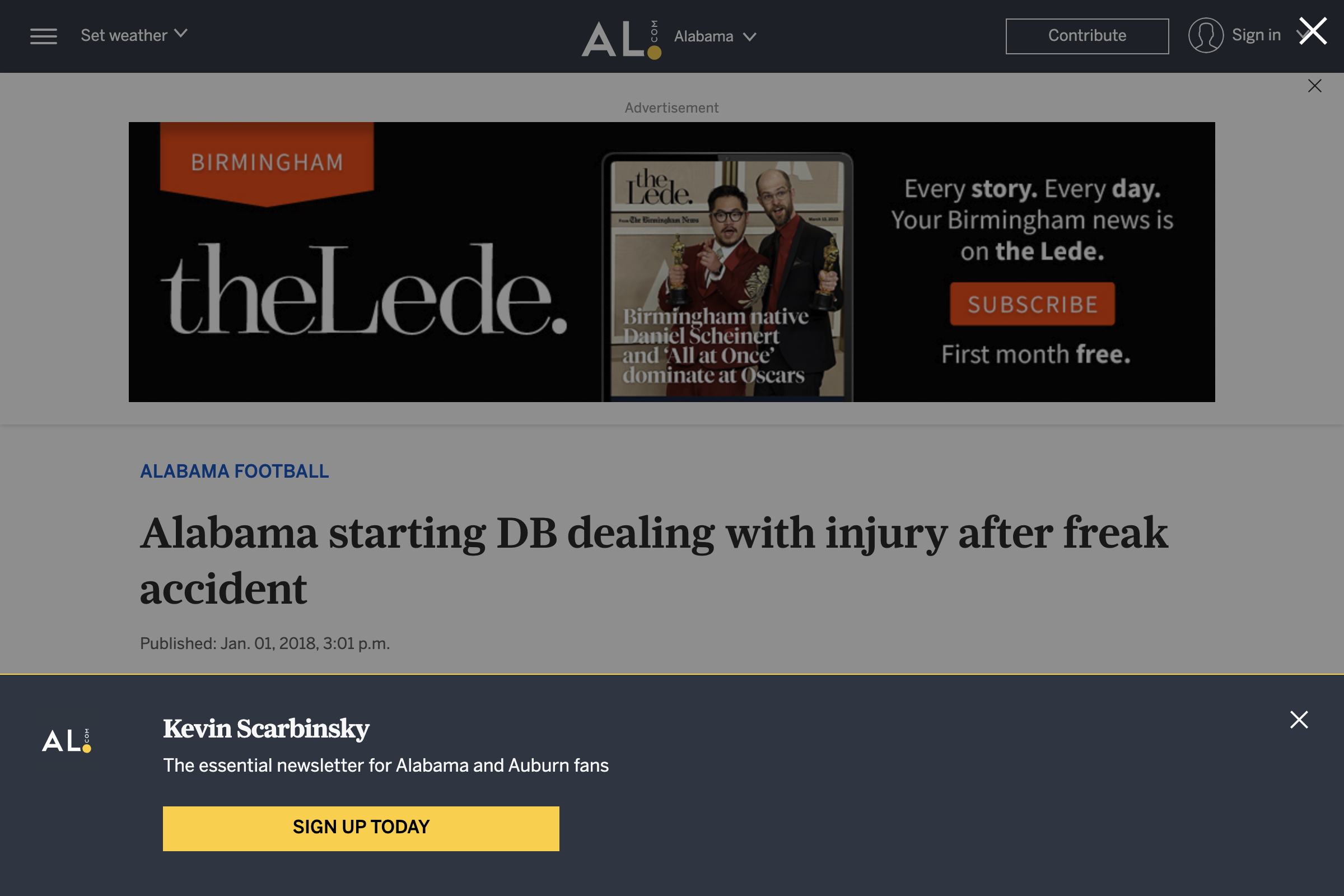}
\\
\textbf{Instruction:} Retrieve all the words with white font on a red background.\\\\
\textbf{Reference Response:} 
1. Birmingham, 2. Subscribe
\end{mybody}
\begin{mybody}
\textbf{Human Response:} \textbf{\green{The words with white font on a red background are: ‘BIRMINGHAM’ and ‘SUBSCRIBE’.}}
\end{mybody}
\begin{mybody}
\textbf{GPT-4V Response:} \textbf{\red{The words with white font on a red background are "Every story. Every day." and "First month free."}}\\
\textbf{\purple{Hallucination}}
\end{mybody}
\begin{mybody}
\textbf{Gemini Pro-Vision Response:} \textbf{\red{Advertisement, theLede, Every story, Every day, First, Your Birmingham news is on the Lede, SUBSCRIBE}} \\
\textbf{\purple{Partially correct response with Hallucination}}
\end{mybody}
\begin{mybody}
\textbf{LLaVA-Next-34B:} \textbf{\red{The words with white font on a red background are "Birmingham" and "the Lede"}}\\
\textbf{\purple{Partially correct response with Hallucination}}
\end{mybody}
\begin{mybody}
\textbf{ShareGPT-4V-7B Response:} \textbf{\red{Birmingham, Alabama}} \\
\textbf{\purple{Partially correct response with Hallucination}}
\end{mybody}
\begin{mybody}
\textbf{LLaVA-1.5-13B Response:}
\textbf{\red{Birmingham, Alabama, The Lede, Alabama football, Obama, starting, DB, dealing, with injury, after free president, Alabama and Auburn fans, sign up today.}} \\
\textbf{\purple{Partially correct response with Hallucination}}
\end{mybody}
\begin{mybody}
\textbf{GPT-4 w/ Layout-aware OCR + Caption Response:} \textbf{\red{BIRMINGHAM, Advertisement, the, Lede., Every story.Every day, Your, Birmingham, news is, on the Lede., the, Lede, SUBSCRIBE,Birmingham native, Daniel Scheinert, and, All at Once', First month free, ALABAMA FOOTBALL, Alabama starting DB dealing with injury after freak accident, Published: Jan. 01, 2018, 3:01 p.m, Kevin Scarbinsky, The essential newsletter for Alabama and Auburn fans}} \\
\textbf{\purple{Partially correct response with Hallucination}}

\end{mybody}
\captionof{figure}{In this task, \textit{all models} hallucinate. However, all but \textit{GPT-4V} produce a partially correct response.}
\label{fig:web_usage_wrong_2}
\end{table*}

\clearpage

\subsection{Infographic}
This section provides qualitative analysis of the \textit{Infographic} visual scenario and our findings across the models, as shown in Figures \ref{fig:infographic_correct_1}, \ref{fig:infographic_correct_2}, \ref{fig:infographic_wrong_1}, \ref{fig:infographic_wrong_2}.
\begin{table*}[th!]
\fontsize{9.0pt}{\baselineskip}\selectfont
\linespread{0.9}\selectfont
\begin{mybody}
\includegraphics[scale=0.32]{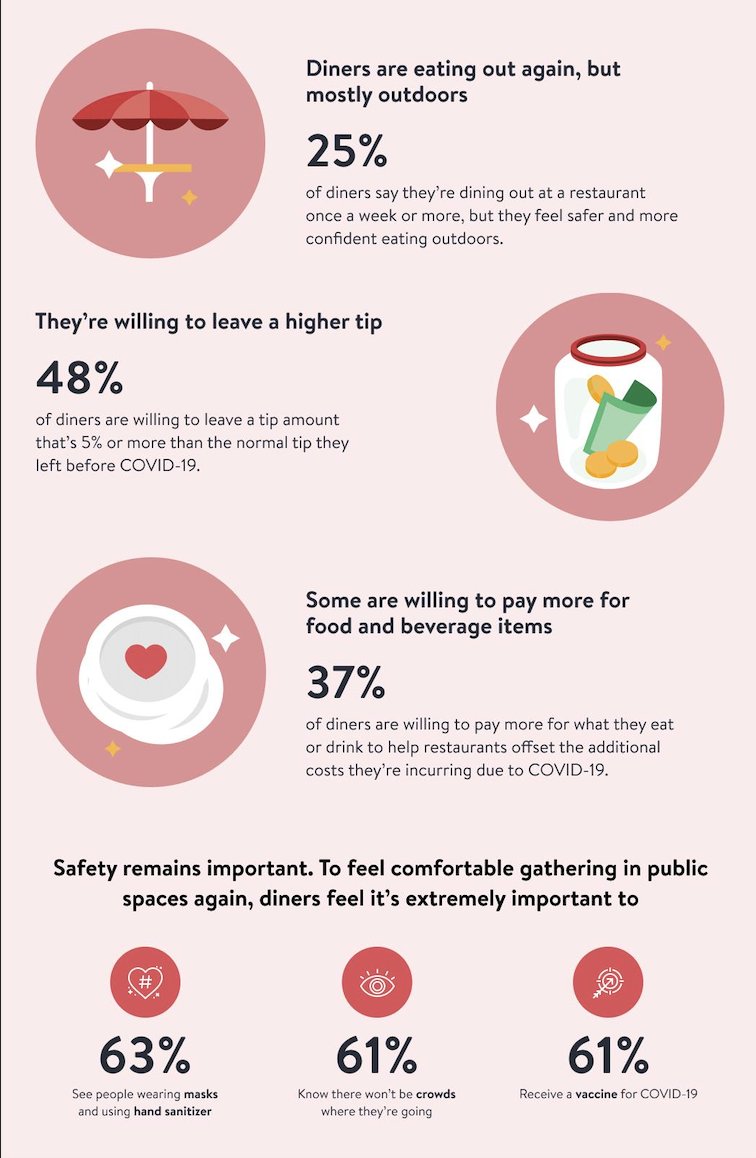}
\\
\textbf{Instruction:} What is the message corresponding to the eyes symbol?\\\\
\textbf{Reference Response:} 
It says 61\% of the surveyors believe that there won't be any crowd where they're going to eat out.
\end{mybody}
\begin{mybody}
\textbf{Human Response:} \textbf{\green{61\% of respondents want to know there won't be crowds where they're going to eat.}}
\end{mybody}
\begin{mybody}
\textbf{GPT-4V Response:} \textbf{\green{The eyes symbol corresponds to the message: "61\% of diners feel it’s extremely important to know there won’t be crowds where they’re going." This indicates that a significant number of diners prioritize knowledge about the expected crowd levels at dining venues as a factor in their comfort with eating out again in the context of the COVID-19.}}
\end{mybody}
\begin{mybody}
\textbf{Gemini Pro-Vision Response:} \textbf{\green{The message corresponding to the eyes symbol is "Know there won't be crowds where they're going". 61\% of diners surveyed said this}}
\end{mybody}
\begin{mybody}
\textbf{LLaVA-Next-34B:} \textbf{\red{The message corresponding to the eyes symbol is "See people wearing masks and using hand sanitizer." This suggests that diners are paying attention to and expecting others to follow safety measures such as wearing masks and using hand sanitizer to prevent the spread of COVID-19.}}\\
\textbf{\purple{Lack of context-sensitive text-rich visual reasoning \& Hallucination}}
\end{mybody}
\begin{mybody}
\textbf{ShareGPT-4V-7B Response:} \textbf{\red{The message corresponding to the eyes symbol is "They're willing to leave a higher tip." This statement suggests that a significant number of diners are willing to leave a bigger tip than the standard tip, which is 15-20\%.}} \\
\textbf{\purple{Lack of context-sensitive text-rich visual reasoning \& Hallucination}}
\end{mybody}
\begin{mybody}
\textbf{LLaVA-1.5-13B Response:} \textbf{\red{The message corresponding to the eyes symbol is that diners are willing to pay more for food and beverage items. This indicates that customers are interested in high-quality dining experiences and are willing to invest more money in their meals.
}} \\
\textbf{\purple{Lack of context-sensitive text-rich visual reasoning \& Hallucination}}
\end{mybody}
\begin{mybody}
\textbf{GPT-4 w/ Layout-aware OCR + Caption Response:} \textbf{\red{The message corresponding to the eye symbol is: "To feel comfortable gathering in public spaces again, diners feel it's extremely important to See people wearing masks and using hand sanitizer" with 63\% agreement,}} \textbf{\green{"Know there won't be crowds where they're going" with 61\% agreement,}} and to \textbf{\red{"Receive a vaccine for COVID-19" with also 61\% agreement.
}} \\
\textbf{\purple{Correct response with Hallucination }}
\end{mybody}
\vspace{-1em}
\captionof{figure}{In this task, closed-source LMMs \textit{GPT-4V} and \textit{Gemini Pro-Vision} correctly respond to the instruction. This can be attributed to their ability for visual perception, text recognition, and context-sensitive text-rich visual reasoning. All other models hallucinate.}
\label{fig:infographic_correct_1}
\end{table*}


\begin{table*}[th!]
\fontsize{9.0pt}{\baselineskip}\selectfont
\linespread{0.9}\selectfont
\begin{mybody}
\includegraphics[scale=0.2]{images/qualitative_examples/infographic/correct/20167.jpeg}
\\
\textbf{Instruction:} Calculate the difference in the percent of respondents who have confidence and do not have confidence in Obama from the country with a doll in their stamp.\\\\
\textbf{Reference Response:} 
Country is Poland. Percent of respondents who approve is 47 and the percent who do not have confidence is 38. The difference of the two is 9 percent.
\end{mybody}
\begin{mybody}
\textbf{Human Response:} \textbf{\green{The country is Poland and the difference between the respondents who have and do not have confidence in Obama is 9\%.}}
\end{mybody}
\begin{mybody}
\textbf{GPT-4V Response:} \textbf{\green{The country represented by a stamp with a doll on it is Poland. According to the data provided: 47\% of respondents have a lot of confidence or some confidence in Obama's ability to do the right thing in world affairs. 38\% of respondents do not have too much confidence or no confidence at all in Obama's ability to do the right thing in world affairs.
To calculate the difference, we subtract the smaller percentage from the larger percentage: 47\% (Confidence) - 38\% (No confidence) = 9\%}}
\end{mybody}
\begin{mybody}
\textbf{Gemini Pro-Vision Response:} \textbf{\red{For the country with the doll stamp (Nigeria), 85\% of respondents have confidence in Obama, and 13\% do not have confidence in}}\\
\textbf{\purple{Hallucination}}
\end{mybody}
\begin{mybody}
\textbf{LLaVA-Next-34B:} \textbf{\red{The image shows a collection of postage stamps from various countries, each with a percentage of respondents who have confidence in Obama and those who do not. The country with a doll in its stamp is the United States, which has a 70\% confidence and a 26\% no confidence.Therefore, there is a 44\% difference between the percentage of respondents who have confidence in Obama and those who do not have confidence in the United States.}}\\
\textbf{\purple{Hallucination}}
\end{mybody}
\begin{mybody}
\textbf{ShareGPT-4V-7B Response:} \textbf{\red{The country with the highest percentage of people with confidence in Obama is Great Britain, with 80\% of the respondents having confidence. On the other hand, the country with the lowest percentage of people with confidence in Obama is Pakistan, with only 20\% of the respondents having confidence. The difference in the percentages between two countries is 60\%.}}\\
\textbf{\purple{Hallucination}}
\end{mybody}
\begin{mybody}
\textbf{LLaVA-1.5-13B Response:} \textbf{\red{In the image, the country with a doll in their stamp is the United States. The percentage of respondents who have confidence in Obama is 63\%, while the percentage of respondents who do not have confidence in Obama is 37\%. The difference between two percentages is 63 - 37 = 26\%.}}\\
\textbf{\purple{Hallucination}}
\end{mybody}
\begin{mybody}
\textbf{GPT-4 w/ Layout-aware OCR + Caption Response:} 
\textbf{\red{The image provided, unfortunately, doesn't contain information showing a specific country with a doll in their stamp. }}\\
\textbf{\purple{Hallucination}}

\end{mybody}
\vspace{-1em}
\captionof{figure}{In this task, \textit{GPT-4V} correctly responds to the instruction. This can be attributed to its ability for math and context-sensitive text-rich visual reasoning. All other models hallucinate their responses.}
\label{fig:infographic_correct_2}
\end{table*}



\begin{table*}[th!]
\fontsize{9.0pt}{\baselineskip}\selectfont
\linespread{0.9}\selectfont
\begin{mybody}
\includegraphics[scale=0.14]{images/qualitative_examples/infographic/incorrect/11535.jpeg}
\\
\textbf{Instruction:} Highlight the states where the percent of positive tests is greater than 30\%.\\\\
\textbf{Reference Response:} 
WA and MS
\end{mybody}
\begin{mybody}
\textbf{Human Response:} \textbf{\green{Washington and Mississippi
}}
\end{mybody}
\begin{mybody}
\textbf{GPT-4V Response:} \textbf{\red{If you need to highlight states where the percent of positive tests is greater than 30\% according to the provided image, you can use a graphic editing tool to draw attention to these states...}}\\
\textbf{\purple{Does not ground the response to the image}}
\end{mybody}
\begin{mybody}
\textbf{Gemini Pro-Vision Response:} \textbf{\red{The states where the percent of positive tests is greater than 30\% are: Louisiana, Alabama, Georgia}}, and \textbf{\green{Mississippi}} \\
\textbf{\purple{Partially correct response with Hallucination}}
\end{mybody}
\begin{mybody}
\textbf{LLaVA-Next-34B:} \textbf{\red{The states where the percent of positive tests is greater than 30\% are highlighted in red on the map. These states include:- AK (Alaska)- ME (Maine) VT (Vermont)- NH (New Hampshire)}}\\
\textbf{\purple{Correct Reasoning but wrong response. Lack of fine-grained visual perception}}
\end{mybody}
\begin{mybody}
\textbf{ShareGPT-4V-7B Response:} 
\textbf{\red{The states where the percent of positive tests is greater than 30\% are: New York, New Jersey, Pennsylvania, Maryland, Virginia, Massachusetts, Rhode Island, Delaware, Connecticut, Ohio, Indiana, Michigan, Illinois, Minnesota, Wisconsin, Iowa, Nebraska, South Dakota, North Dakota, West Virginia, Kentucky, Tennessee, Alabama, Georgia, Florida, Arkansas, Oklahoma, Texas}}, \textbf{\green{Washington, D.C., and Mississippi.}}\\
\textbf{\purple{Correct response with Hallucination}}
\end{mybody}
\begin{mybody}
\textbf{LLaVA-1.5-13B Response:} \textbf{\red{In the image, there are several states where the percent of positive tests is greater than 30\%. These states include: California: 31.7\%, Texas: 30.6\%, Florida: 30.5\%, New York: 30.4\%, Pennsylvania: 30.3\%}}\\
\textbf{\purple{Hallucination}}
\end{mybody}
\begin{mybody}
\textbf{GPT-4 w/ Layout-aware OCR + Caption Response:} \textbf{\red{The image doesn't provide specific information on which states have a percent of positive tests greater than 30\%.}}\\
\textbf{\purple{Lack of context-sensitive text-rich visual reasoning}}

\end{mybody}
\captionof{figure}{In this task, the closed source LMMs, \textit{GPT4V} does not ground the response to the image. \textit{Gemini-Vision-Pro},  \textit{ShareGPT-4V-7B} \& \textit{LLaVA-1.5-13B} hallucinate in their responses. \textit{GPT-4 w/ Layout-aware OCR + Caption Response} (Augmented LLM approach) produces the wrong answer, due to lack of context-sensitive text-rich visual reasoning.}
\label{fig:infographic_wrong_1}
\end{table*}



\begin{table*}[th!]
\fontsize{9.0pt}{\baselineskip}\selectfont
\linespread{0.9}\selectfont
\begin{mybody}
\includegraphics[scale=0.3]{images/qualitative_examples/infographic/incorrect/33199.jpeg}
\\
\textbf{Instruction:} How many production methods are used under pour-over/drip category?\\\\
\textbf{Reference Response:} 
10
\end{mybody}
\begin{mybody}
\textbf{Human Response:} \textbf{\green{There are 10 methods in the pour-over/drip category.}}
\end{mybody}
\begin{mybody}
\textbf{GPT-4V Response:} \textbf{\red{In the pour-over/drip category of this coffee chart, there appear to be six production methods used. These methods are as follows: Auto Drip, Cone, Flat Bottom, AeroPress, Woodneck, Chemex}}\\
\textbf{\purple{Lack of fine-grained visual perception}}
\end{mybody}
\begin{mybody}
\textbf{Gemini Pro-Vision Response:} \textbf{\red{There are 8 production methods listed under pour-over/drip category.}}\\
\textbf{\purple{Lack of fine-grained visual perception}}
\end{mybody}
\begin{mybody}
\textbf{LLaVA-Next-34B:} \textbf{\red{The image shows a chart of coffee production methods, and under the "Pour-Over/Drip" category, there are four different methods illustrated: 1. Chemex 2. V60 3. Kalita Wave 4. Bee House. Each method is represented by a different coffee-making device, and they are connected by lines indicating the flow of the coffee-making process.)}}\\
\textbf{\purple{Lack of fine-grained visual perception}}
\end{mybody}
\begin{mybody}
\textbf{ShareGPT-4V-7B Response:} \textbf{\red{There are two production methods used under the pour-over/drip category: French press and drip coffee maker.}}\\
\textbf{\purple{Lack of fine-grained visual perception}}
\end{mybody}
\begin{mybody}
\textbf{LLaVA-1.5-13B Response:} \textbf{\red{There are three production methods used under the pour-over/drip category.
}}\\
\textbf{\purple{Lack of fine-grained visual perception}}
\end{mybody}
\begin{mybody}
\textbf{GPT-4 w/ Layout-aware OCR + Caption Response:} \textbf{\red{The OCR text does not provide specific information on the number of production methods used under the pour-over/drip category.
}}\\
\textbf{\purple{Lack of context-sensitive text-rich visual reasoning}}

\end{mybody}
\captionof{figure}{In this task, the \textit{all LMMs}, produce the wrong response due to lack of fine-grained visual perception and possibly poor math reasoning. \textit{GPT-4 w/ Layout-aware OCR + Caption Response} (Augmented LLM approach) produces the wrong answer, due to lack of context-sensitive text-rich visual reasoning.}
\label{fig:infographic_wrong_2}
\end{table*}

\clearpage

\subsection{Miscellaneous Natural Scenes}
This section provides qualitative analysis of the \textit{Miscellaneous Natural Scenes} visual scenario and our findings across the models, as shown in Figures \ref{fig:misc_correct_1}, \ref{fig:misc_correct_2}, \ref{fig:misc_wrong_1}, \ref{fig:misc_wrong_2}.

\begin{table*}[th!]
\fontsize{9.0pt}{\baselineskip}\selectfont
\linespread{0.9}\selectfont
\begin{mybody}
\includegraphics[scale=0.6]{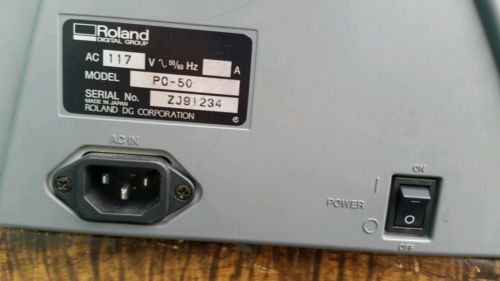}
\\
\textbf{Instruction:} Provide your opinion on whether changing the switch's current position will activate the instrument.\\\\
\textbf{Reference Response:} 
Despite turning the switch to the On position, the system will remain powered off due to the absence of a connected power cable.
\end{mybody}
\begin{mybody}
\textbf{Human Response:} \textbf{\green{It will not. The instrument needs to be plugged in to work.}}
\end{mybody}
\begin{mybody}
\textbf{GPT-4V Response:} \textbf{\green{The image shows the back panel of an electronic device with a power switch in the 'OFF' position. Changing the switch's current position from 'OFF' to 'ON' is likely to activate the instrument, provided that it is properly connected to a power source and there are no malfunctions with the instrument itself.}}
\end{mybody}
\begin{mybody}
\textbf{Gemini Pro-Vision Response:} \textbf{\red{Flipping the switch to the "ON" position will power the instrument.}}\\
\textbf{\purple{Lack of fine-grained visual perception}}
\end{mybody}
\begin{mybody}
\textbf{LLaVA-Next-34B:} \textbf{\green{Based on the image, the switch is currently in the "OFF" position. If you were to change the switch to the "ON" position, it would likely activate the instrument, assuming that the instrument is in proper working condition and has been set up correctly.}}
\end{mybody}
\begin{mybody}
\textbf{ShareGPT-4V-7B Response:} \textbf{\red{Based on the image, it is not possible to determine if changing the switch's current position will activate the instrument. The switch is currently in the off position, and the instrument is not in use. To determine if changing the switch's position will activate the instrument, one would need to observe the instrument's response to the switch's movement.}}\\
\textbf{\purple{Lack of fine-grained visual perception}}
\end{mybody}
\begin{mybody}
\textbf{LLaVA-1.5-13B Response:} \textbf{\red{Based on the image, the switch is currently in the "off" position, which means that the instrument is not activated. To activate the instrument, the switch would need to be moved to the "on" position. This would allow the electrical current to flow through the device, enabling it to function properly.}} \\
\textbf{\purple{Lack of fine-grained visual perception}}
\end{mybody}
\begin{mybody}
\textbf{GPT-4 w/ Layout-aware OCR + Caption Response:} \textbf{\red{The instruction is ambiguous with not enough visual information available in the provided OCR text or the image caption regarding a switch. Therefore, it is not possible to provide an opinion on whether changing the switch's current position will activate the instrument.}} \\
\textbf{\purple{Lack of context-sensitive text-rich visual reasoning}}
\end{mybody}
\captionof{figure}{In this task, \textit{GPT4V} \& \textit{LLaVA-Next-34B} produces the correct response, whereas \textit{all other LMMs}, produce the wrong response due to lack of fine-grained visual perception. \textit{GPT-4 w/ Layout-aware OCR + Caption Response} (Augmented LLM approach) produces the wrong answer, due to lack of context-sensitive text-rich visual reasoning.}
\label{fig:misc_correct_1}
\end{table*}


\begin{table*}[th!]
\fontsize{9.0pt}{\baselineskip}\selectfont
\linespread{0.9}\selectfont
\begin{mybody}
\includegraphics[scale=0.5]{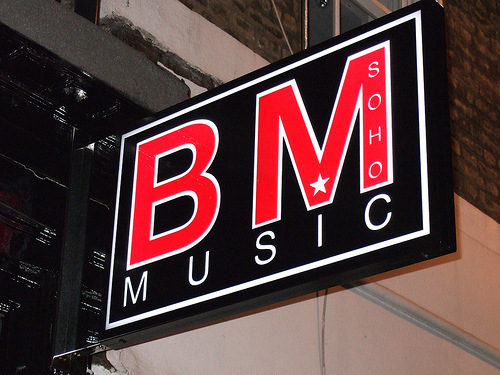}
\\
\textbf{Instruction:} What is written on a leg of the letter M?\\\\
\textbf{Reference Response:} 
SOHO
\end{mybody}
\begin{mybody}
\textbf{Human Response:} \textbf{\green{SOHO}}
\end{mybody}
\begin{mybody}
\textbf{GPT-4V Response:} \textbf{\green{On the leg of the letter "M" in the image, it says "SOHO."}}
\end{mybody}
\begin{mybody}
\textbf{Gemini Pro-Vision Response:} \textbf{\green{SOHO}}
\end{mybody}
\begin{mybody}
\textbf{LLaVA-Next-34B:} \textbf{\red{The image shows a sign with the letters "BM" in large red font, and the word "MUSIC" in smaller white font. The leg of the letter "M" is not visible in the image, so it's not possible to determine what is written on it. If you're referring to the leg of the "M" in "MUSIC," it's not visible in the image provided.}}
\\
\textbf{\purple{Lack of fine-grained visual perception}}
\end{mybody}
\begin{mybody}
\textbf{ShareGPT-4V-7B Response:} \textbf{\red{A star is written on the leg of the letter M.}}\\
\textbf{\purple{Lack of fine-grained visual perception}}
\end{mybody}
\begin{mybody}
\textbf{LLaVA-1.5-13B Response:} \textbf{\red{The word "Music" is written on the leg of the letter M.}}\\
\textbf{\purple{Lack of fine-grained visual perception}}
\end{mybody}
\begin{mybody}
\textbf{GPT-4 w/ Layout-aware OCR + Caption Response:} 
\textbf{\red{U}}\\
\textbf{\purple{Lack of context-sensitive text-rich visual reasoning}}

\end{mybody}
\captionof{figure}{In this task, \textit{GPT-4V} and \textit{Gemini Pro-Vision} correctly respond to the instruction. This can be attributed to their ability for text recognition, and context-sensitive text-rich visual reasoning. Open source LMMs like \textit{LLaVA-Next-34B}, \textit{ShareGPT4V} \& \textit{LLaVA-1.5-13B} due to lack of fine-grained visual perception. \textit{GPT-4 w/ Layout-aware OCR + Caption Response} (Augmented LLM approach) predict the wrong answer due to lack of context-sensitive text-rich visual reasoning.}
\label{fig:misc_correct_2}
\end{table*}



\begin{table*}[th!]
\fontsize{9.0pt}{\baselineskip}\selectfont
\linespread{0.9}\selectfont
\begin{mybody}
\includegraphics[scale=0.3]{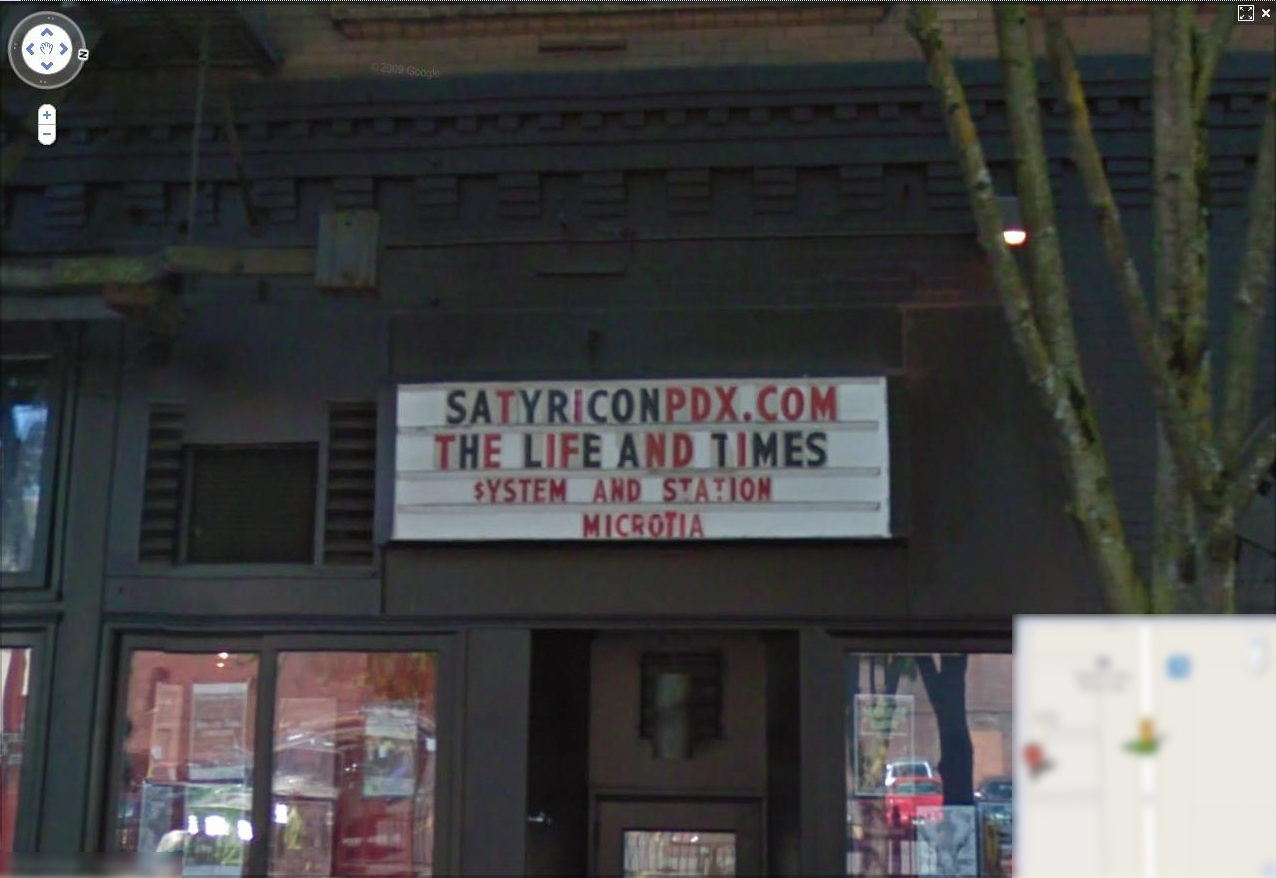}
\\
\textbf{Instruction:} What would the first line read if we filter the letters in the red?\\\\
\textbf{Reference Response:} 
SAYRCON
\end{mybody}
\begin{mybody}
\textbf{Human Response:} \textbf{\green{SA YR CON}}
\end{mybody}
\begin{mybody}
\textbf{GPT-4V Response:} \textbf{\red{The first line reads "SATYRICONPDX.COM" with the letters in red. If we filter out the letters in red, the remaining letters spell "SATICONDX.COM".}}\\
\textbf{\purple{Lack of fine-grained visual perception}}
\end{mybody}
\begin{mybody}
\textbf{Gemini Pro-Vision Response:} \textbf{\red{SATYRICONPDX.COM}}\\
\textbf{\purple{Lack of fine-grained visual perception}}
\end{mybody}
\begin{mybody}
\textbf{LLaVA-Next-34B:} \textbf{\red{SATYRICONPX.COM}}
\\
\textbf{\purple{Lack of fine-grained visual perception}}
\end{mybody}
\begin{mybody}
\textbf{ShareGPT-4V-7B Response:} 
\textbf{\red{If we filter the letters in the red, the first line would read "SATYRICONP"}}\\
\textbf{\purple{Lack of fine-grained visual perception}}
\end{mybody}
\begin{mybody}
\textbf{LLaVA-1.5-13B Response:} \textbf{\red{If we filter the letters in the red, the first line would read "Satyricon PDX".}}\\
\textbf{\purple{Lack of fine-grained visual perception}}
\end{mybody}
\begin{mybody}
\textbf{GPT-4 w/ Layout-aware OCR + Caption Response:} \textbf{\red{The instruction can't be executed on the provided information because the layout aware OCR text doesn't indicate any letters in red.}} 
\\
\textbf{\purple{Lack of context-sensitive text-rich visual reasoning}}

\end{mybody}
\captionof{figure}{In this task, \textit{all LMMs} produce the wrong answer due to lack of fine-grained visual perception. \textit{GPT-4 w/ Layout-aware OCR + Caption Response} (Augmented LLM approach) produces the wrong answer, due to lack of context-sensitive text-rich visual reasoning.}
\label{fig:misc_wrong_1}
\end{table*}



\begin{table*}[th!]
\fontsize{9.0pt}{\baselineskip}\selectfont
\linespread{0.9}\selectfont
\begin{mybody}
\includegraphics[scale=0.6]{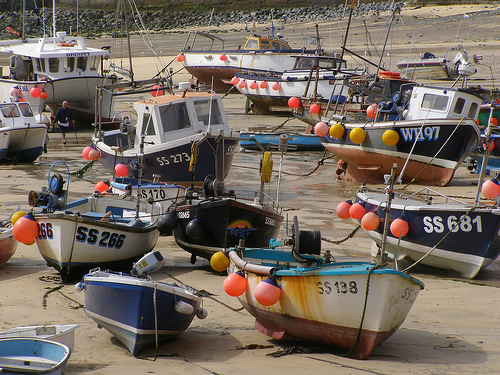}
\\
\textbf{Instruction:} Get the number of the boat with three yellow and one red round items hanging from it.\\\\
\textbf{Reference Response:} 
WH97
\end{mybody}
\begin{mybody}
\textbf{Human Response:} \textbf{\green{97}}
\end{mybody}
\begin{mybody}
\textbf{GPT-4V Response:} \textbf{\red{The boat with three yellow and one red round items (which are likely buoys) hanging from it has the number SS 170.}}\\
\textbf{\purple{Lack of fine-grained visual perception}}
\end{mybody}
\begin{mybody}
\textbf{Gemini Pro-Vision Response:} \textbf{\red{SS 273}}\\
\textbf{\purple{Lack of fine-grained visual perception}}
\end{mybody}
\begin{mybody}
\textbf{LLaVA-Next-34B:} \textbf{\red{The boat with three yellow and one red round items hanging from it is the one with the number "SS 681" on it.}}
\\
\textbf{\purple{Lack of fine-grained visual perception}}
\end{mybody}
\begin{mybody}
\textbf{ShareGPT-4V-7B Response:} \textbf{\red{The boat with three yellow and one red round items hanging from it is SS138.}}\\
\textbf{\purple{Lack of fine-grained visual perception}}
\end{mybody}
\begin{mybody}
\textbf{LLaVA-1.5-13B Response:} \textbf{\red{The boat with three yellow and one red round items hanging from it is the one with the number "S527" on it.}}\\
\textbf{\purple{Lack of fine-grained visual perception \& Hallucination}}
\end{mybody}
\begin{mybody}
\textbf{GPT-4 w/ Layout-aware OCR + Caption Response:} \textbf{\red{The instruction doesn't specify which boat has three yellow and one red round items hanging from it, so it's impossible to provide the number of that boat based on the provided OCR text and image caption.}}
\\
\textbf{\purple{Lack of context-sensitive text-rich visual reasoning}}

\end{mybody}
\captionof{figure}{In this task, all LMMs produce the wrong response due to lack of fine-grained visual perception.\textit{GPT-4 w/ Layout-aware OCR + Caption Response} (Augmented LLM approach) produces the wrong answer, due to lack of context-sensitive text-rich visual reasoning.}
\label{fig:misc_wrong_2}
\end{table*}

\end{document}